\documentclass[lettersize,journal]{IEEEtran}
\usepackage{amsmath,amsfonts}
\usepackage{algorithmic}
\usepackage{algorithm}
\usepackage{array}
\usepackage[caption=false,font=normalsize,labelfont=sf,textfont=sf]{subfig}
\usepackage{textcomp}
\usepackage{stfloats}
\usepackage{url}
\usepackage{verbatim}
\usepackage{graphicx}
\usepackage{cite}
\usepackage{amsfonts}
\usepackage{url}
\usepackage{booktabs}
\usepackage{multirow}
\usepackage{array}
\usepackage{tabularx}
\usepackage{soul}
\usepackage{color}
\begin{document}
	
	\title{Multi-objective Optimal Roadside Units Deployment in Urban Vehicular Networks}
	
	% Enhanced Adaptive NSGA-III for Optimal RSU Deployment and Data Offloading Management in Urban Vehicular Networks: A Constrained Multi-objective Optimization Approach
	
	\author{Weian Guo~\IEEEmembership{Member,~IEEE,}, Zecheng Kang, Dongyang Li, Lun Zhang, and Li Li~\IEEEmembership{Member,~IEEE}
		% <-this % stops a space
		% <-this % stops a space
		\thanks{This work is supported by the National Natural Science Foundation of China under Grant Number 62273263, 72171172 and 71771176; Shanghai Municipal Science and Technology Major Project (2022-5-YB-09); Natural Science Foundation of Shanghai under Grant Number 23ZR1465400. (Corresponding author: Li Li).}
		\thanks{Weian Guo, Zecheng Kang, and Dongyang Li are with Sino-German College of Applied Sciences, Tongji University, Shanghai, China (email:\{guoweian, kangzecheng, lidongyang0412\}@163.com). Lun Zhang is with School of Transportation, Tongji University, Shanghai, China (email: lun$\_$zhang@tongji.edu.cn). Li Li is with the Department of Electronics and Information Engineering, Tongji University, Shanghai, 201804, China (email: lili@tongji.edu.cn).}
	}
	
	
	
	\maketitle
	
	\begin{abstract}
		The significance of transportation efficiency, safety, and related services is increasing in urban vehicular networks. Within such networks, roadside units (RSUs) serve as intermediates in facilitating communication. Therefore, the deployment of RSUs is of utmost importance in ensuring the quality of communication services. However, the optimization objectives, such as time delay and deployment cost, are commonly developed from diverse perspectives. As a result, it is possible that conflicts may arise among the objectives. Furthermore, in urban environments, the presence of various obstacles, such as buildings, gardens, lakes, and other infrastructure, poses challenges for the deployment of RSUs. Hence, the deployment encounters significant difficulties due to the existence of multiple objectives, constraints imposed by obstacles, and the necessity to explore a large-scale optimization space. To address this issue, two versions of multi-objective optimization algorithms are proposed in this paper. By utilizing a multi-population strategy and an adaptive exploration technique, the methods efficiently explore a large-scale decision-variable space. In order to mitigate the issue of an overcrowded deployment of RSUs, a calibrating mechanism is adopted to adjust RSU density during the optimization procedures. The proposed methods also take care of data offloading between vehicles and RSUs by setting up an iterative best response sequence game (IBRSG). By comparing the proposed algorithms with several state-of-the-art algorithms, the results demonstrate that our strategies perform better in both high-density and low-density urban scenarios. The results also indicate that the proposed solutions substantially improve the efficiency of vehicular networks.
	\end{abstract}
	
	
	\section{Introduction}
	\IEEEPARstart{W}{ith} the accelerated expansion of urbanization, traffic networks have grown significantly. To provide a safe and efficient traffic environment, Vehicle-to-Infrastructure (V2I) framework in the Intelligent Transportation System (ITS) gains prominence. Since roadside units (RSUs) provide data transfer services for vehicles in such a framework, it is essential to deploy RSUs at optimal locations to assure quality of service (QoS) in complex urban environments \cite{kim2016}.
	
    To provide optimal RSU deployment in urban areas, the challenges can be summed up in two aspects. First, many factors with potential conflicting perspectives, such as quality of service and economics of deployment, should be taken into account simultaneously in real-world implementations \cite{yu2021}. Second, the urban environment is characterized by a large number of vehicles, various obstacles and structures, and a variety of road types. Hence, it is vital to contemplate the appropriate placement of RSUs to alleviate signal interference resulting from intricate topographies  \cite{fogue2018}. Therefore, an optimal RSU deployment should provide effective and cost-efficient communication services to guarantee the reliability, and stability of connectivity \cite{xue2017}.
	
	In existing studies, a part of the focus is on optimizing several single objectives using mathematical programming methods, heuristics \cite{laha2021,li2018} or data-driven approaches \cite{liu2020}, which may not be competent to simultaneously balance the trade-offs between objectives from different perspectives. In another part of the research, multi-objective optimization modeling is used, but it disregards the complexity of actual implementations in urban environments, such as physical landscapes and excessive RSU density \cite{wang2021}. In order to address these challenges, this study incorporates techniques for addressing constraints into a multi-objective optimization algorithm. This approach aims to consider the impact of actual urban terrain on the RSU deployment. Furthermore, during deployment, the data offloading is also modeled to take into account the stability and reliability of the connection between vehicles and RSUs \cite{Raja2020}. 
	
    To investigate optimal RSU deployments, we consider both communication quality and development cost together in the problem modeling. Since the landscape is gridded into hundreds of map pieces, the problem belongs to large-scale optimization issues. Meanwhile, the road terrain and buildings also bring constraints. To deal with these problems, this paper proposes an improved NSGA-III algorithm that uses a multiple-population strategy and an adaptive balancing strategy to balance exploration and exploitation in a large-scale optimization search space. Furthermore, it incorporates an offspring calibration mechanism to prevent excessive RSU density. The contributions of this paper can be summarized as follows:
	
	\begin{itemize}
		\item  First, we propose a multi-objective deployment model for RSUs in urban scenarios. To evaluate the QoS in communications, we have set time delays as objectives. In the meantime, the construction budget is also an objective. For the constraints, we consider the elements in urban environments, including road maps, building landscapes, RSU density, and so forth. In vehicle-RSU connections, a strategy named iterative best response with sequential game (IBRSG) is proposed for data offloading.
		
		\item Second, we improved the NSGA-III algorithm by adding a multi-population strategy and an adaptive balancing strategy for exploring and exploitation in the search space. This helps solve the proposed multi-objective optimization problem with constraints. To further optimize the density of RSUs, we also present a distance-inhibition method for calibrating solutions during the optimization process.
	\end{itemize}
	
	The remaining sections of this paper are structured in the following manner: Section \ref{sec:literature} presents a comprehensive overview of the existing literature pertaining to the RSUs deployment. In Section \ref{sec:problem}, a multi-objective problem with constraints is modeled, specifically targeting urban scenarios. In order to handle the problems, an enhanced multi-objective optimization algorithm is elaborated upon in Section \ref{sec:algorithm}. The experiments are conducted in Section \ref{sec:experiments}, where comparisons and related analyses of the results are also performed. Finally, the conclusions and future work are presented in Section \ref{sec:conlusions}.
	
	\section{Related Work}
	\label{sec:literature}
	The RSUs deployment in urban environments involves modeling and algorithm design. The related works are introduced as follows. 
	
	\subsection{Optimization of RSUs Deployment}	
	The growing demand for vehicle network connectivity has resulted in the widespread deployment of RSUs. The deployment scenarios have been categorized as static deployment  \cite{guerna2019} and dynamic deployment \cite{cai2020,cai20202} accordingly. For deployment modeling, it is common to encounter two types of objectives. The first one concerns communication profit, while the second addresses expenses. The profit-related factors encompass RSU network coverage \cite{cai20202}, response time delay \cite{ghorai2018,anbalagan2021}, service profit \cite{ni2018}, and so on. For the expenses issue, construction and maintenance costs for RSUs tend to be included \cite{guerna2021}.

	First, researchers combine several optimization objectives into one target and use single-objective optimization algorithms to solve problem. The methods include mathematical programming, heuristic algorithms, data-driven methodologies, etc. The authors  in \cite{qi2020} weight two factors, say time delay and cellular bandwidth cost, as one objective to model RSU deployment to build a nonlinear integer programming problem, then use a two-step centralized heuristic approach for solving the problem. In \cite{guerna2019}, researchers consider RSU coverage and the number of RSUs to build a single-objective optimization model and propose an improved genetic algorithm to solve the problem. In \cite{liang2021}, an enhanced genetic technique is improved to address a single-objective optimization model by prefixing weights for network coverage, communication interference, and deployment costs, respectively. In current research of deployment modeling, the data-driven method is popular to examine real-world traffic flows to better reflect communication needs. In \cite{ghost2023}, the authors train a traffic forecasting model using Google Maps data, providing accurate data for RSU deployment. The study in \cite{yu2021} emphasizes the importance of RSU deployment in VANETs and proposes a mechanism to prioritize investments based on real traffic features. The authors of \cite{ben2021} suggest a traffic data-based RSU deployment strategy to reduce costs and increase communication coverage using traffic data from Maghrebian cities. Additional research on real-world data is available in \cite{anbalagan2021,ni2018,liu2023}. These algorithms anticipate RSU demand and QoS using historical data, traffic flow data, and vehicle trajectories to determine appropriate RSU deployment locations. However, in the above studies, the single-objective method cannot find a pareto set of solutions since each objective's weight is prefixed. Thus, the solutions rarely provide a full view of RSU deployment in practice. In addition, mathematical programming also takes a long time to solve optimization problems when confronted with large-scale variables.
	
	Second, several studies have employed multi-objective optimization techniques for the purpose of deploying RSUs. However, it is worth noting that the research conducted in this particular area is very limited. In \cite{yuzhang2023} and \cite{zhangz2021}, the authors consider three objectives, namely RSU deployment cost, coverage area, and expected data transmission delay respectively, to propose a multi-objective optimization model, then employ NSGA-II related algorithms to address the problem. An improved multi-objective particle swarm optimization technique optimizes RSU deployment cost and communication coverage simultaneously \cite{zhangl2023}. Olia et al. optimize communication efficiency while minimizing RSU count by using NSGA-II \cite{olia2017}. Wang et al. propose a novel multi-objective differential evolutionary method that minimizes RSU count and maximizes communication coverage \cite{wang2021}. In the above studies, it has been observed that multi-objective optimization techniques have the capability to generate Pareto-optimal solutions. However, these methods often encounter difficulties when dealing with intricate constraints, such as those found in urban environments. Additionally, the optimization efficiency may be compromised, particularly when encountering a large-scale decision-variable space.
	
	\subsection{Some Related Issues}
	\subsubsection{Constraints Handling}
	Constraints are an unavoidable aspect of modeling the deployment of RSUs in urban conditions. Hence, the researchers in \cite{ghorai2018} and \cite{laha2021} incorporate budgetary limitations, geographical constraints, and road construction into their approaches for addressing the deployment of RSUs, which brings challenges in problem modeling and algorithm design. In order to overcome these constraints, \cite{ghorai2018} and \cite{laha2021} have put forth enhancements, namely the constrained Delaunay triangulation (CDT) approach and the enhanced greedy algorithm, respectively. By effectively considering and dealing with the constraints, the optimization strategies for the RSUs deployment will become more viable and realistic.
	 
	\subsubsection{Data Offloading Problem}
	Many academics have developed techniques and strategies for data offloading in vehicular networks, which is influential on RSUs' efficiency. In \cite{li2020,sun2020,huang2020}, the authors focus on optimizing specific metrics such as maximizing V2I system utility, minimizing offloading latency \cite{qian2021,zhou2019,liu2019}, or reducing system budgets \cite{cheng2019,yang2019,li20202}. However, these approaches often lean on static or predetermined models, which might not efficiently handle dynamic traffic flows. To address this issue, heuristic approaches combine multiple algorithms to provide flexible solutions, such as hybrid intelligent optimization algorithms \cite{sun2020} or load balancing schemes \cite{zhang20192,lin2022s}. However, it also brings challenges with large computational demands and low efficiency in complex scenarios. Besides,   based on historical and real-time data, supervised learning and reinforcement learning approaches are also favored in optimizing data offloading strategies. Liu et al. present two reinforcement learning algorithms to optimize data offloading and resource allocation, taking into account vehicle traffic, and dynamic requests under different communication environments \cite{liu20192}. These algorithms are also computationally intensive, and their performances are dependent on the training procedures as well. To investigate the interactions among vehicles and RSUs, there also exist game theory methods to assign RSUs resources for vehicle data offloading \cite{zhang2017,wu2020}. By formulating the RSU deployment as a game, the authors can analyze the strategic behavior of RSUs and derive optimal deployment strategies that maximize their individual benefits while considering the overall network performance.

	\section{Problem Formulation}
	\label{sec:problem}
	In this section, we mainly take into account two aspects of the RSUs deployment. First, delay time is involved, which includes minimizing the total delay and maximizing the maximum delay in latency-sensitive areas. Second, to control the construction cost, minimizing the number of RSUs is also set as an optimization objective. In the modeling, the constraints consist of the areas covered by various kinds of obstacles, which are essential factors in real-world urban scenarios.
	
%	\subsection{Definitions and Notations}
%	The main definitions and notations are presented as follows. 
%	\begin{itemize}
%		\item $V$ - the index set of vehicles
%		\item $T$ - the index set of time periods
%		\item $K$ - the index set of all positions
%		\item $R$ - the index set of RSU locations
%		\item $O$ - the index set of obstacles (e.g., buildings)
%		\item $A$ - the set of latency-sensitive areas
%		\item $d_{ij}^t$ - the delay experienced by vehicle $i$ at time period $t$ if it is connected to RSU $j$
%%		\item $d_{ij}^{t,trans}$ - the transmission delay in $d_{ij}^t$
%%		\item $d_{ij}^{t,queue}$ - the queuing delay in $d_{ij}^t$
%		\item $x_k$ - a binary variable to depict the location index, equal to 1 if an RSU is deployed at location $k$, and 0 otherwise
%%		\item $y_{ij}^t$ - a binary variable, equal to 1 if vehicle $i$ is connected to RSU $j$ at time period $t$, and 0 otherwise
%%		\item $D_{max}$ - the maximum acceptable delay in delay-sensitive areas
%%		\item $L_{fs}$ - the free-space path loss
%%		\item $L_{shad}$ - the shadowing loss
%%		\item $Tr_{ij}^t$ - the transmission rate of vehicle $i$ at time period $t$ if it is connected to RSU $j$
%%		\item $L_p$ - the packet size
%%		\item $B$ - the bandwidth
%%		\item $P_t$ - the transmission power
%%		\item $N_0$ - the noise power spectrum density
%	\end{itemize}
	
	\subsection{Objective Functions}
	\label{sec:objFunc}
	In this paper, we assume there are two ways of transferring data. A vehicle prefers a connection to a RSU, while cellular networks are connected if the vehicle is out of RSUs' coverage.
	
	The first goal is to decrease the overall latency delay that all vehicles experience at all time-periods because this helps improve the effectiveness of network communication as a whole. By minimizing the total latency time, it ensures that the network meets the requirements of as many vehicles as possible. The objective is given in \eqref{eqn:total_delay}.
	\begin{equation}
		\label{eqn:total_delay}
		\min \sum_{i=1}^V \sum_{t=1}^T d_{i}^t 
	\end{equation}
	where $d$ is the total latency time, $V$ is the number of vehicles, $T$ is the number of time-periods, $i$ and $t$ are the corresponding indexes of vehicle and time-period, respectively. For $d_i^t$, it consists of several components and the details can be found in Appendix \ref{sec:app:explain_time_delay}. 
	
    In \eqref{eqn:total_delay}, the latency time is evaluated for the whole target region. However, there are several geographical areas that are latency-sensitive, such as accident-prone areas, crossroads surrounding buildings, the central business district, and other similar locales, where the latency time should be strictly limited. Therefore, the second objective aims to minimize the maximum delay-time in latency-sensitive areas. By this means, we can ensure that the vehicular network satisfies the stringent requirements of latency-sensitive responses. The objective is given in \eqref{eqn:max_delay}.
	\begin{equation}
		\label{eqn:max_delay}
		\min \max_{i \bot A} \sum_{t=1}^T d_{i}^t
	\end{equation}
	where $d_{i}^t$ can be obtained as the same as in \eqref{eqn:total_delay}, $i$ is the index of vehicle, $A$ is the set of latency-sensitive areas, $i \bot A$ means the vehicle $i$ is located at the area $A$, and $T$ is all time periods. Based on \eqref{eqn:max_delay}, it limits the maximum time-delay for the latency-sensitive areas, which further strengthens the communication efficiency in such areas.
	
	Lastly, the third objective focuses on minimizing the quantity of deployed RSUs to control the investment costs of construction and further reduce maintenance expenses. The objective is described in \eqref{eqn:rsu_number}.
	\begin{equation}
		\label{eqn:rsu_number}
		\min \sum_{k=1}^K x_k
	\end{equation}
	where $K$ is the number of all potential positions, $x_k = 1$ means the location with index $k$ is positioned by a RSU, while $x_k = 0$ otherwise.	
	
	\subsection{Descriptions of Constraints}
	In urban areas, the constraints for RSU deployment are mainly from two aspects. The first kind of constraint is from the urban environment, which is related to obstacles such as buildings, gardens, rivers, and many others, where RSUs cannot be deployed within these obstructions. Therefore, RSUs must not be deployed in such locations. For the obstacles set $O=\{o_1,o_2,..,o_K\}$, $o_i=1$ means there exists an obstacle at location $i$, while $o_i=0$ otherwise, where $i \in \{1,...,K\}$ and $K$ is the number of all potential positions. Meanwhile, the locations of RSUs are defined as a set $R=\{r_1,r_2,...,r_K\}$ ,where $r_i=1$ means there exists a RSU at location $i$, while $r_i=0$ otherwise. Then, the constraint can be expressed by an inequality shown in \eqref{eqn:constraints_obstacles_1}.
	\begin{equation}
		\label{eqn:constraints_obstacles_1}
		r_i \land o_i \neq 1, \quad i\in\{1,...,K\}
	\end{equation} 
	In \eqref{eqn:constraints_obstacles_1}, $\land$ is the logic symbol ``and". It means that $r_j$ and $o_j$ cannot simultaneously be the value of $1$. If a RSU is deployed inside an obstacle, the violation is calculated by the distance from the RSU to the nearest edge of the obstacle.
%	By changing \eqref{eqn:constraints_obstacles_1} to an equality constraint and summarizing all positions, we obtain \eqref{eqn:constraints_obstacles}.
%	\begin{equation}
%		\label{eqn:constraints_obstacles}
%		\sum_{j=1}^{k} {\lnot r_j \lor \lnot o_j} = 0
%	\end{equation}
%	where $\lnot$ and $\lor$ are the logic symbol ``not" and ``or" respectively.  	
	Second, signal interference may occur when RSUs are deployed too closely, deteriorating communication quality and network performance \cite{mao2021}. Therefore, the distance limitation is implemented to prevent communication interference between RSUs. By presetting a minimum distance $D_{min}$ between RSUs, it helps reduce interference risk and improve network stability. Therefore, the constraint is expressed in \eqref{eqn:constraints_dis}.
	\begin{equation}
		\label{eqn:constraints_dis}
		\mathbf{dis}(r_i,r_j)  \geq D_{min}, \quad \forall i,j\in\{1,...,K\} , j \neq i
	\end{equation}
	where $r_i=1$ and $r_j=1$ are any two positions deploying RSUs, $\mathbf{dis}()$ is a function to calculate the distance between two positions, $D_{min}$ is a preset minimum distance.  For \eqref{eqn:constraints_dis}, it is used to control the density of RSUs. However, during the optimization process, this constraint on the RSU density may prevent algorithms from yielding feasible solutions. Therefore, in Section \ref{sec:algorithm}, two versions of algorithms are proposed. The one employs this RSUs-density constraint in calibrating candidate solutions during the whole optimization process, while the other one does not.	
	
	\subsection{Problem Statement}
	Based on the established optimization objectives and constraints, the RSUs deployment in urban environment can be stated as a multi-objective optimization problem with constraints, which is formulated in \eqref{eqn:modeling}.	
	\begin{align}
		\label{eqn:modeling}
		\begin{split}
			&\quad\quad \quad \min \quad \left\{ \begin{array}{l} f_1 = \sum\limits_{i=1}\limits^V \sum\limits_{t=1}\limits^T d_{i}^t \\ [2mm]
				f_2 = \max \limits_{i \bot A} \sum\limits_{t=1}\limits^N d_{i}^t \\[3mm]
				 f_3 = \sum\limits_{k =1}^K x_k    \end{array} \right. \\
			& \text{subject to:} \\
			& \text{(1)} \quad r_i \land o_i \neq 1, \quad i\in\{1,...,K\} \\
			& \text{(2)} \quad \mathbf{dis}(r_i,r_j)  \geq D_{min}, \quad \forall i,j\in\{1,...,K\} , j \neq i
		\end{split}
	\end{align}	
	Based on the \eqref{eqn:modeling}, we have built a multi-objective optimization problem with constraints. To address this problem, the algorithms are designed in Section \ref{sec:algorithm}.

	\section{Adaptive Multi-population NSGA-III}
	\label{sec:algorithm}
	% \subsection{EEBNSGA-III}
    According to the modeling in \eqref{eqn:modeling}, the RSUs deployment is a multi-objective optimization with constraints. In addition, in the urban scenario, the map is gridded into hundreds of pieces, which makes this problem exhibit large-scale optimization properties. To simultaneously take into account the multiple objectives, an improved NSGA-III is considered in this paper. In the improvements, it is crucial to design a balancing strategy for exploitation and exploration because of the large scale of the decision-variable space. To be specific, the innovations are generally made in two aspects. First, we employ a multi-population strategy to enhance the exploration capability of the proposed algorithm. Meanwhile, adaptive parameter tuning for both mutation and crossover is proposed to dynamically balance exploitation and exploration. Second, to handle the constraints, an $\epsilon$-level comparison method cited from \cite{fan2019} is used to compare solutions even if they violate the constraints. In addition, to make RSUs retain a certain distance when generating offspring, a calibrating strategy is also proposed in this paper. All the details are illustrated in the following subsections.
	
	\subsection{Rules for Comparing Solutions}
	For multi-objective optimization algorithms, the rules for comparing solutions are crucial to determine the selection mechanism. In our design, we compare solutions in three cases. First, for solutions $\mathbb{A}$ and $\mathbb{B}$, if $\mathbb{A}$ satisfies the constraints, but $\mathbb{B}$ does not, then $\mathbb{A}$ is better than $\mathbb{B}$. Second, if $\mathbb{A}$ and $\mathbb{B}$ both satisfy the constraints, then pareto-dominance rule is preferred in comparing the two solutions \cite{Guo2022}. Third, if neither $\mathbb{A}$ nor $\mathbb{B}$ satisfies the constraints, we employ an improved $\epsilon$-level rule  to compare the two solutions \cite{fan2019}. The details about the $\epsilon$-level rule are explained as follows. For two solutions, $\mathbb{A}$ and $\mathbb{B}$, their overall constraint violations are defined as $\phi^{A}$ and $\phi^{B}$ respectively. Then, for any $\epsilon \geq 0$, the epsilon level comparison $\preceq_{\epsilon}$ is defined in \eqref{eqn:epsilon_comparison}.
	\begin{equation}
		\label{eqn:epsilon_comparison}
		(\mathbb{A},\phi^{A})  \preceq_{\epsilon} (\mathbb{B},\phi^{B}) \leftrightarrow  \left \{
		\begin{split}
			&\mathbb{A} \preceq \mathbb{B},\ \textbf{if}\ \phi^{A}, \phi^{B} \leq \epsilon\\
			&\mathbb{A} \preceq \mathbb{B},\ \textbf{if}\ \phi^{A} = \phi^{B}\\
			&\phi^{A} < \phi^{B},\ \textbf{otherwise}
		\end{split}
		\right.
	\end{equation}
	where $\mathbb{A} \preceq \mathbb{B}$ means that $\mathbb{A}$ dominates $\mathbb{B}$. The symbol ``$\preceq_{\epsilon}$" means the comparison is conducted with a $\epsilon$-level relaxation of constraints. The detail explanations on $\epsilon$-level rule are presented in Appendix \ref{sec:app:epsilon_rules}.  
	
	\subsection{Strategies in Balancing Exploitation and Exploration}
	\label{subsec:optimal_individual_migration}
	In the urban environment, the regions are partitioned into hundreds of grids for RSUs deployment, which brings a multitude of variables and therefore forms a large-scale optimization issue. To address such problems, it is important to emphasize the algorithm's capability to balance exploration and exploitation. In this paper, we employ a multi-population strategy so that the whole population is divided into $N$ sub-populations. In each sub-population, the algorithm runs independently. Among sub-populations, solutions migrate after each generation. In this way, superior solutions in one sub-population will replace inferior solutions in other sub-populations. This migration strategy helps improve the diversity of solutions in each population and, meanwhile, accelerates convergence by sharing superior solutions among sub-populations. The pseudo-codes of the migration operator are given in Algorithm \ref{alg:migration}.
	\begin{algorithm}[!htbp]
		\caption{Migration Operator among Sub-populations}
		\label{alg:migration}
		\begin{algorithmic}[1]
			\REQUIRE
			\begin{itemize}
				\item [ ]
				\item Total amount of sub-populations $m$
				\item Sub-populations $SP_1$, $SP_2$, ... ,$SP_m$
				\item Amount of emigrants $\mathbf{N_{EM}}$ in each sub-population
			\end{itemize}
			\ENSURE 
			\begin{itemize}
				\item [ ]
				\item Updated $SP_1$, $SP_2$, ... ,$SP_m$
			\end{itemize}
			\FOR{$i=1$ to $m$}
			\STATE Select the best $\mathbf{N_{EM}}$ individuals in $SP_i$
			\STATE Store the $\mathbf{N_{EM}}$ individuals in $D_i$
			\ENDFOR        
			\FOR{$i=1$ to $m$}
			\STATE $\mathbf{EM_i} = \mathbf{NULL}$
			\STATE Merge all $D_j$ ($j \in \{1,...,m\} \& j \neq i$) into $\mathbf{EM_i}$
			\STATE In $SP_i$, select the worst $(m-1)\times\mathbf{N_{EM}}$ individuals and mark them as $\mathbf{IM_i}$
			\STATE In $SP_i$, instead $\mathbf{IM_i}$ by $\mathbf{EM_i}$
			\ENDFOR
			\RETURN Updated $SP_1$, $SP_2$, ... ,$SP_m$
		\end{algorithmic}
	\end{algorithm}	
	
	Besides the multi-population strategy, in the proposed algorithm, during iterations, an adaptive strategy in both crossover and mutation parameter tuning is applied. According to each sub-population's performance in the current generation, mutation and crossover probabilities are adjusted dynamically so as to balance exploration and exploitation. If the current best fitness in a sub-population has not been improved for a predefined number of iterations, the mutation probability increases while the crossover probability decreases. In this way, exploration is promoted. Conversely, if the current best fitness improves after the same predefined number of iterations, the mutation probability decreases and the crossover probability increases. Then exploitation is emphasized. The pseudo-codes are provided in Algorithm \ref{alg:rate_change}. It is noteworthy that due to the independent execution of the algorithm in each sub-population, there may be variations in the crossover rate and mutation rate across distinct sub-populations.
		\begin{algorithm} [!htbp]
		\caption{Adaptive Crossover Rate and Mutation Rate}
		\label{alg:rate_change}
		\begin{algorithmic}[1]
			\REQUIRE
			\begin{itemize}
				\item [ ]
				\item A sub-population $SP$
				\item The iteration index $g$, where $g>1$
				\item Initial crossover rate $C_r(g-1)$ and mutation rate $M_r(g-1)$
				\item Value limitations of crossover rate $C_{max}$, $C_{min}$, and limitations of mutation rate $M_{max}$ and $M_{min}$
				\item Range-ability for changing crossover rate $\Delta_c >0$ and for mutation rate $\Delta_m >0$
			\end{itemize}
			\ENSURE  
			\begin{itemize}
				\item [ ]
				\item Updated $C_r(g)$ and $M_r(g)$
			\end{itemize}
			\STATE Evaluate the current sub-population $P$
			\STATE Mark the current best fitness as $F(g)$
			\IF{$F(g)$ is better than $F(g-1)$}
			\STATE $C_{r}(g) = C_{r}(g-1) + \Delta_c$
			\STATE $M_{r}(g) = M_{r}(g-1) - \Delta_m$
			\ELSE
			\STATE $C_{r}(g) = C_{r}(g-1) - \Delta_c$
			\STATE $M_{r}(g) = M_{r}(g-1) + \Delta_m$
			\ENDIF
			\STATE Regulate the values of $C_{r}(g)$ and $M_{r}(g)$ by limitations
			\RETURN $C_r(g)$, $M_r(g)$
		\end{algorithmic}
	\end{algorithm}
		
	\subsection{Offspring Calibrating Strategy}
	\label{subsec:calibrating}
    In this problem, the locations of RSUs cannot be too close in order to prevent interference in data transmission. Therefore, an inhibition distance is preset as $D_{min}$. The offspring naturally generated by multi-objective optimization algorithms may not satisfy the constraints, and it should be calibrated after generating offspring. The steps are described as follows.
	\begin{enumerate}
		\item Calculate the distance between all RSU pairs.
		\item Identify the RSU pairs that violate the inhibition distance constraint $D_{min}$ according to \eqref{eqn:constraints_dis}.
		\item For each violating RSU pair, rank the RSUs according to the traffic volume in the range of RSU's coverage.
		\item Retain the RSU with higher traffic volume and remove the deployment of the other RSU.
	\end{enumerate}
    This calibrating strategy guarantees that the retaining RSUs satisfy the inhibition distance constraint and therefore avoid communication interference. Meanwhile, the RSU retention mechanism in this strategy also helps maintain large-scale coverage for traffic volume.	
	
	\subsection{Data Offloading Strategy}
	In transportation networks, vehicles offload data to RSUs and request services from RSUs \cite{saleem2021}. When vehicles have more than one RSU to be connected, the data offloading strategy is much more influential in evaluating network performance, e.g., the calculation of latency time \cite{ghorai2018}. In this section, we propose a novel strategy named iterative best response sequential game (IBRSG). For $V$ vehicles and $R$ RSUs, the strategy for the $i^{th}$ vehicle is marked as $s_i$. The value of $s_i$ is an index value of a certain RSU. Then the establishment of data offloading can be described as follows.
    	\begin{enumerate}
    			\item Establish a random strategy $S_{old}=\{s_{o1}, s_{o2},...,s_{oV}\}$, where $s_{oi} \in \{1,...,R\}$, $i\in \{1,...,V\}$
    			\item Set up a $\mathbf{NULL}$ strategy $S_{new}=\{s_{n1}, s_{n2},..., s_{nV}\}$, where $s_{ni}=\mathbf{NULL}$, $i\in \{1,...,V\}$
    			\item Initialize an index value $\mathbf{index} =1$
    			\label{dataoffloading:out_loop}
    			\item For all vehicles, their index is $i \in \{1,...,V\}$, update their connection strategy based on the following rule: For $i > \mathbf{index} $, the $i^{th}$ vehicle uses the strategy $s_{oi}$; For $i < \mathbf{index}$, the $i^{th}$ vehicle uses the strategy $s_{ni}$; For $i$ equals to $\mathbf{index}$, the $i^{th}$ vehicle is connected to a RSU, which can minimize the total latency time for all $V$ vehicles. Store the $i^{th}$ vehicle's strategy to $s_{ni}$. 
    			\label{dataoffloading:inner_loop}
    			\item If $\mathbf{index} < V$, then $\mathbf{index}=\mathbf{index}+1$ and goto ``Step \ref{dataoffloading:inner_loop}"; Otherwise, update $S_{new} = \{s_{n1}, s_{n2},..., s_{nV}\}$   
    			\item If $||S_{old} - S_{new}||$ $\geq$ $\mathbf{error}$, instead $S_{old}$ by $S_{new}$, then goto ``Step \ref{dataoffloading:out_loop}"; Otherwise, stop the iterations and output $S_{new} $
    			\label{dataoffloading:error_setting}
    		\end{enumerate}
       In Step \ref{dataoffloading:error_setting}, $\mathbf{error}$ is a predefined threshold to evaluate the difference between two data offloading strategies. The whole loop runs until there is no big change between two iterative strategies. $S_{new}$ is the final output strategy for data offloading. The use of a sequential strategy based on iterative best responses ensures the convergence and robustness of the algorithm.  Since different ways to implement data offloading will affect the transmission efficiency and the load balance of RSUs, we also compare the proposed strategy with several other data offloading strategies in Section \ref{sec:experiments}. 

	\subsection{Proposed Algorithm}
    According to the above designs, we propose adaptive multi-population NSGA-III, which has two versions. For the first version, it is with offspring calibrating strategy (marked by AM-NSGA-III-c), while the other version (named as AM-NSGA-III) is not. The pseudo-codes of AM-NSGA-III-c are shown in Algorithm \ref{alg:AM-NSGA-III-c}. For the version of AM-NSGA-III, its pseudo-codes just delete the part of the offspring calibrating strategy that is marked in Line \ref{alg:line} of Algorithm \ref{alg:AM-NSGA-III-c}. As mentioned in Section \ref{subsec:calibrating}, the offspring calibrating strategy is used to mitigate the RSU density by reducing the number of RSUs during algorithm executions. For the version without the offspring calibrating strategy, the number of RSUs may increase, potentially resulting in violations due to the overcrowded RSUs deployment. The violations persist consistently and are passed down from iteration to iteration, ultimately undermining the algorithm's ability to achieve feasible solutions. In addition, the number of RSUs obtained by AM-NSGA-III may be greater than that obtained by AM-NSGA-III-c. In the subsequent experiments, we will investigate the influence of offspring calibration on the algorithm's performance in detail.
    
    \begin{algorithm}
    	\caption{AM-NSGA-III-c}
    	\label{alg:AM-NSGA-III-c}
    	\begin{algorithmic}[1]
    		\REQUIRE
    		\begin{itemize}
    			\item []
    			\item Objective functions $f_1$, $f_2$ and $f_3$
    			\item Obstacle locations $O_c$, Distance constraint $D_{min}$
    			\item Population size $N$
    			\item Amount of sub-populations $m$
    			\item Initial crossover rate $Cr$, initial mutation rate $Mr$, initial violation tolerance $\epsilon$
    			\item Maximum number of iterations $\mathbf{G}$
    		\end{itemize}
    		\ENSURE
    		\begin{itemize}
    			\item []
    			\item A set of pareto solutions $\mathbf{Pareto}$
    		\end{itemize}
    		\STATE \textbf{Initialization:}
    		\begin{enumerate}
    			\item Randomly initialize a population $P$ with size of $N$.
    			\item Partition $P$ into $m$ sub-populations $SP_{i}$, where $i\in\{1,...,m\}$, the size of each sub-population is $\frac{N}{m}$
    			\item In each sub-population $SP_{i}$, the initial crossover rate $c_{ri}=C_r$, mutation rate $m_{ri}=M_r$, violation tolerance $\epsilon_i=\epsilon$
    			%				\item Evaluate the fitness of $P$ by three objectives $f_1$, $f_2$ and $f_3$, with the constraints $O_c$.				
    			%				\item In $Sub\_P_{i}$, amount of Emigrants $EM$
    			%				\item In $Sub\_P_{i}$, amount of candidate individuals to be instead $IM=(k-1)\times EM$ 
    			%				\item Maximum number for iterations $\mathbf{MAX\_G}$
    		\end{enumerate}
    		\STATE \textbf{Main Loop:}
    		\FOR{$g$ = $1$ to $\mathbf{G}$}
    		\FOR{$i$ = $1$ to $m$}
    		\STATE Evaluate the fitness of $SP_i$ based on the objectives $f_1$, $f_2$, $f_3$, where the IBRSG data offloading strategy is employed in the evaluations.
    		%\STATE For the $i^{th}$ individual, its fitness is marked as $fit_{i}(t)$
    		%\STATE The current best fitness is marked as $fit_{best}(t)$    		
    		\STATE Run NSGA-III for each sub-population $SP_{i}$ to get offspring
    		\STATE Calibrate offspring according to $D_{min}$, where the rules are given in Section \ref{subsec:calibrating}.
    		\label{alg:line}
    		\STATE Rank and select solutions to update $SP_i$ based on \eqref{eqn:epsilon_comparison}
    		%\STATE In $Sub\_P_{i}$, collect $EM$ best individuals into $\mathbf{R_i}$ 
    		\STATE Update $c_{ri}$ and $m_{ri}$ according to Algorithm \ref{alg:rate_change}
    		\STATE Update $\epsilon_i$ according to \eqref{eqn:epsilon}
    		\ENDFOR
    		\STATE Migrate solutions among sub-populations according to Algorithm \ref{alg:migration}.			
    		\ENDFOR
    		\STATE Collect all $SP_{i}$ to update $P$
    		\STATE Select non-dominated solutions from $P$ to compose the pareto solutions $\mathbf{Pareto}$
    		\RETURN $\mathbf{Pareto}$
    	\end{algorithmic}
    \end{algorithm}	

	\section{Experiments and Analysis}
	\label{sec:experiments}
    In this section, experiments are conducted to investigate the proposed algorithms' performance. The public datasets of vehicle transportation in Chengdu City, China, are obtained from the ``DiDi Gaia Open Data Program".  The data includes time stamps, coordinates, original locations, and destination locations. According to the data, we construct the traffic volumes by calling the Gaode Map API to perform route planning so that urban traffic scenarios can be simulated. In the urban environment, we select areas with high-density and low-density scenarios, respectively, which are explained in Appendix \ref{sec:app:scenarios}. 
	
	\subsection{Parameter setting}
	In this subsection, two categories of parameters are introduced. The first kind of parameters are about scenarios, while the second kind of parameters are set for algorithms. The parameters on scenarios are presented in Appendix \ref{sec:app:scenarios}. For the parameters in algorithms, we summarize them as follows. The population size ($N$) is set as $360$ and the number of sub-populations ($m$) is set as $3$. For \eqref{eqn:epsilon}, the initial value of parameters are set as follows: $\theta= 0.05\times N=18$, $\alpha= 0.95 $ and $\tau=0.1$.  The initial crossover rate is set as $50\%$, while the mutation rate is set as $0.05$. The crossover rate ranges from $20\%$ to $100\%$, and the mutation rate varies between $0$ and $0.1$. The range-ability for crossover rate and mutation rate are set as $\Delta_c=0.1$ and $\Delta_m=0.01$, respectively. For the migration operator, each sub-population will select the top $10\%$ best solutions to emigrate, while the worst $20\%$ ($10\% \times (m-1)$) solutions will be instead by immigrants. The maximum number $G$ for algorithm's iteration is limited to $50$. 	
	
	\subsection{Evaluations of Algorithms Performance}
\begin{figure}[!htbp]
		\centering
		\includegraphics[width=2.5in]{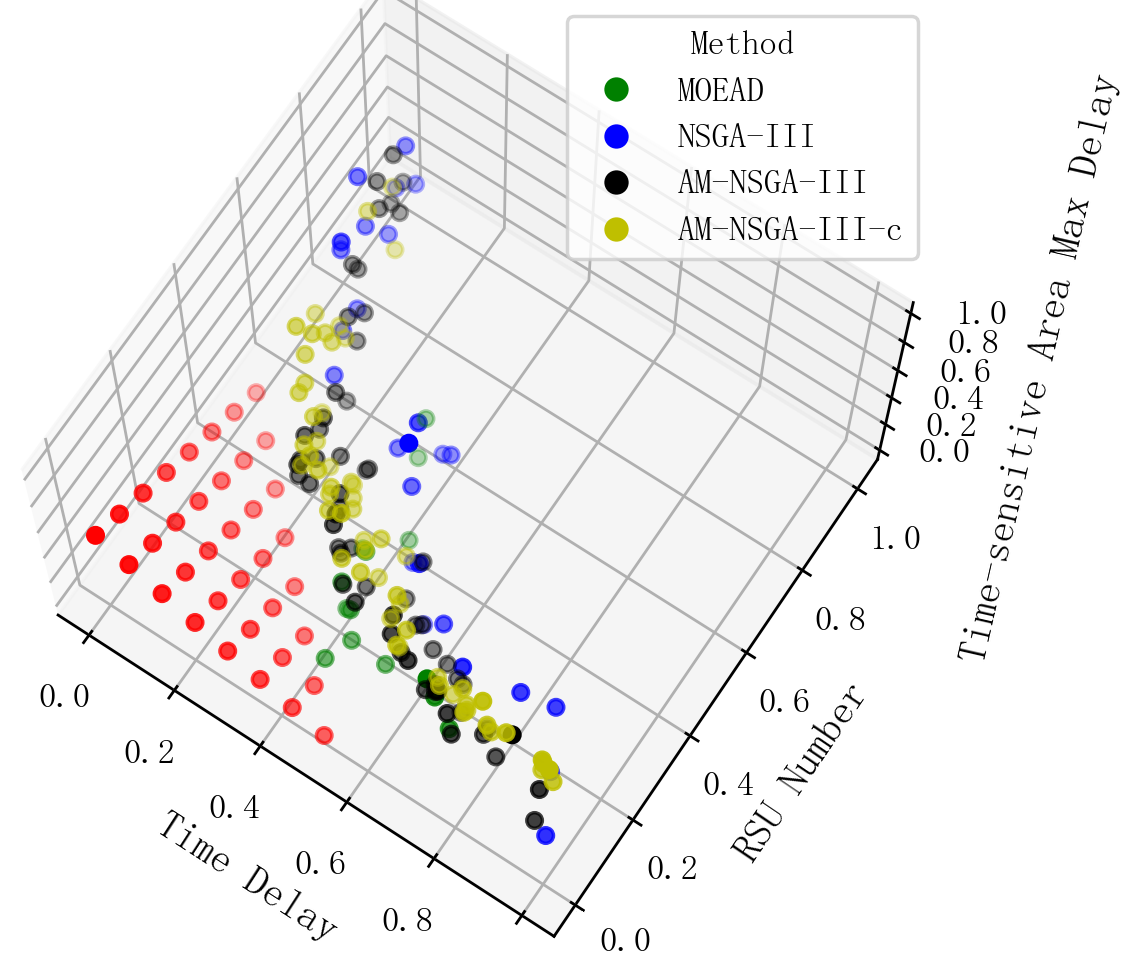}
		\caption{Pareto fronts obtained by four algorithms}
		\label{fig:pareto_2}
	\end{figure}

	In this subsection, we conduct experiments of AM-NSGA-III and AM-NSGA-III-c respectively. Their performances are used to compare with MOEA/D \cite{zhang2007moea} and NSGA-III \cite{deb2013evolutionary}. The pareto-solutions obtained by each algorithm is given in Fig. \ref{fig:pareto_2}, where the red points are the reference point. By this figure, it means the four algorithms can be implemented to the RSUs deployment problem to obtain pareto solutions. 
		
	To obtain a straightforward comparison, we present Table \ref{tab:performance_comparison} with performance metrics including the number of Pareto-optimal solutions (NPS), the number of feasible solutions (NFS), the Inverted Generational Distance (IGD) indicator, the Hypervolume (HV) indicator and the S-Metric indicator.
	\begin{table*}[!htbp]
		\caption{Performance Comparison of Algorithms}
		\label{tab:performance_comparison}
		\centering
		\begin{tabularx}{\textwidth}{XXXXX}
			\toprule
			\textbf{Metric} & \textbf{MOEA/D} & \textbf{NSGA-III} & \textbf{AM-NSGA-III} & \textbf{AM-NSGA-III-c} \\
			\midrule
			NPS & 14 & 26 &  \textbf{59} & 58 \\
			NFS & 0 & 4 & 15 &  \textbf{22} \\
			IGD & 0.39171 & 0.36199 & 0.25565 &  \textbf{0.24437} \\
			HV & 2.10352 & 3.03098 & 11.60189 &  \textbf{12.07229} \\
			S-Metric &  \textbf{0.0242} & 0.0623 & 0.0302 & 0.0296 \\
			% \multicolumn{2}{l}{Long Text Here} & Data 10 & Data 11 \\
			\bottomrule		
		\end{tabularx}
	\end{table*}
	
	\subsubsection{Performances in High Density Scenarios}
    From Table \ref{tab:performance_comparison}, it can be observed that both AM-NSGA-III and AM-NSGA-III-c achieve a high score for NPS and NFS, which significantly outperforms MOEA/D and NSGA-III. Moreover, according to NFS, AM-NSGA-III-c dramatically performs the best. In addition, it is worth noting that MOEA/D does not generate any feasible solutions.
    
    Upon analysis of the IGD indicator and careful examination of the pareto front depicted in Figure \ref{fig:pareto_2}, it becomes apparent that the pareto front produced by the AM-NSGA-III-c algorithm exhibits the highest proximity to the reference points, which serves as a clear indication of the superiority of AM-NSGA-III-c's performance. In addition, the HV indicator provides further evidence for this result, indicating that AM-NSGA-III-c has superior performance compared to other algorithms. The S-Metric indicator reveals that MOEA/D achieves the most favorable outcome, while the AM-NSGA-III-c algorithm performs the second best. However, considering the absence of feasible solutions obtained by MOEA/D, it can be concluded that the AM-NSGA-III-c demonstrates superior overall performance for high-density scenarios in the comparisons.
    
    The reasons behind this performance can be attributed to the improved exploration and exploitation capabilities of the AM-NSGA-III-c algorithm, which enable it to effectively navigate the search space and discover more feasible and high-quality solutions that satisfy the given constraints. Additionally, the adaptive mechanism for calibrating crossover and mutation rates helps balance exploration and exploitation, further contributing to performance. 
    %Moreover, the introduction of the calibrating offspring strategy with the RSU inhibition distance constraint plays a crucial role in the algorithm's success. By using this strategy, the iterative evolution process ensures that the offspring meet the RSU inhibition distance constraint. This keeps the solutions from being impossible to find. This constraint-guided approach helps the algorithm focus on the regions of the search space with feasible solutions, thereby increasing the efficiency of the exploration and exploitation processes.
    Moreover, by incorporating the calibrating offspring strategy, AM-NSGA-III-c can effectively balance the trade-off between satisfying constraints and optimizing multiple objectives. As a result, the algorithm generates high-quality and more feasible solutions, which are crucial for the real-world implementation of RSU deployments. 
     
    All the feasible solutions are presented in Appendix \ref{sec:app:experiments_high}. According to Table \ref{tab:performance_comparison}, since MOEA/D fails to obtain any feasible solution, we only list the results obtained by NSGA-III, AM-NSGA-III, and AM-NSGA-III-c, which are shown in Table \ref{tab:performance_objectives}. The three objectives in this table are explained in \eqref{eqn:modeling}. It is noted that  \textbf{Objective 1} is the total latency time,  \textbf{Objective 2} is the maximal delay in the latency-sensitive areas, and \textbf{Objective 3} is the amount of RSUs.  According to Table \ref{tab:performance_objectives}, we know that besides MOEA/D, the algorithm NSGA-III obtained the fewest number of feasible solutions. 
	
    To evaluate the three algorithms, we use the dominance rule to compare all the feasible solutions and obtain Table \ref{tab:pareto_all_high}. According to Table \ref{tab:pareto_all_high}, it can be observed that AM-NSGA-III-c has the largest number of pareto front solutions, followed by AM-NSGA-III, while all feasible solutions from NSGA-III were eliminated, which demonstrates that the two versions of the proposed algorithms outperform both MOEA/D and NSGA-III. In addition, in Table \ref{tab:pareto_all_high}, compared with AM-NSGA-III-c, solutions obtained by AM-NSGA-III usually have a higher number of RSUs. The reason is that, during the algorithm's iteration, AM-NSGA-III does not employ the offspring calibrating strategy. Therefore, in the AM-NSGA-III algorithm's run, violations about minimal distance between RSUs usually exist in generating offspring. However, the same violations will cause RSU deletion in the AM-NSGA-III-c algorithm, and therefore the number of RSUs in AM-NSGA-III-c will be less. On the other hand, the latency time in AM-NSGA-III is generally less than that in AM-NSGA-III-c. Based on the analysis, we know that if decision makers prefer the construction cost, AM-NSGA-III-c is advisable. On the contrary, if decision makers would like to reduce latency time, AM-NSGA-III is appropriate.
	
	To visualize the situations of RSU deployment, we show four feasible solutions of each algorithm to draw the diagrams, which are given in Fig.\ref{fig_sim_nsgaiii_1}, Fig. \ref{fig_sim_eebnsgaiii_1} and Fig. \ref{fig_sim_ieebnsgaiii_1}. In the figures, we mark the latency-sensitive areas with cyan diamonds. The dark blue points are the locations for RSU deployment. In such areas, the maximal latency time should be minimized to pursue fast responses. However, for NSGA-III, it does not result in an overcrowded RSU deployment but also fails to deploy RSUs in narrow traffic hotspot areas. For example, in Fig. \ref{fig_sim_nsgaiii_1}, many RSUs are located very closely for the four deployment strategies. On the other hand, for the four diagrams in Fig. \ref{fig_sim_nsgaiii_1}, especially in the bottom right corner, there are many narrow roads with traffic hotspots, but very few RSUs are deployed at the areas. For AM-NSGA-III, shown in Fig. \ref{fig_sim_eebnsgaiii_1}, although some RSUs are still deployed closely, the narrow areas with traffic hotspots have several RSUs deployed. For AM-NSGA-III-c, shown in Fig. \ref{fig_sim_ieebnsgaiii_1}, the deployed RSUs are relatively uniformly distributed. Meanwhile, the deployment strategy also gives consideration to narrow areas. To sum up, compared with other algorithms, in a high-density environment, AM-NSGA-III-c is competent to obtain enough feasible optimal deployment strategies aiming at the objectives in \eqref{eqn:modeling}.

	\subsubsection{Performances in Low Density Scenarios}
    In urban environment, both high-density and low-density areas exist. In this subsection, we investigate the RSUs deployment in low-density scenarios. In the performance comparisons, we also employ MOEA/D, NSGA-III, AM-NSGA-III, and AM-NSGA-III-c to investigate the RSUs deployment. The metrics include NPS, NFS, IGD, HV and S-Metric, while the results are provided in Table \ref{tab:performance_comparison_low_density}.
	
	\begin{table*}[t]
		\centering
		\caption{Performance Comparison of Algorithms in low density}
		\label{tab:performance_comparison_low_density}
		\begin{tabularx}{\textwidth}{XXXXX}
			\toprule
			\textbf{Metric} & \textbf{MOEA/D} & \textbf{NSGA-III} & \textbf{AM-NSGA-III} & \textbf{AM-NSGA-III-c} \\
			\midrule
			NPS & 22 & \textbf{24} & 23 & 19 \\
			NFS & 6 & 16 & \textbf{18}  &17\\
			IGD & 0.413803 & 0.41570 & 0.36539 &  \textbf{0.19456} \\
			HV & 3.58877 & 4.09742 & 2.65386 &  \textbf{2.2750} \\
			S-Metric &0.06197 &0.08181 &\textbf{0.03986} & 0.05879 \\
			% \multicolumn{2}{l}{Long Text Here} & Data 10 & Data 11 \\
			\bottomrule
		\end{tabularx}
	\end{table*}
	
	Based on the data shown in Table \ref{tab:performance_comparison_low_density}, the quantities of both NPS and NFS achieved by MOEA/D exhibit an increment when compared to its performance in high-density scenarios. Nevertheless, the most significant performance on NFS is still achieved by the AM-NSGA-III and AM-NSGA-III-c algorithms. The distribution of pareto solutions obtained by the proposed algorithms is satisfactory in terms of the metrics of IGD, HV, and S-Metric. This result further supports the competence of the proposed algorithms in handling low-density scenarios.
		
    All feasible solutions are presented in Appendix \ref{sec:app:experiments_low}. The solutions of the algorithms to the three optimization objectives are listed in Table \ref{tab:performance_low_density}. In this table, we find that the number of feasible solutions in NSGA-III increases compared with the case in high-density scenarios, which means that the low-density area is easier for algorithms to handle. On the other hand, in the solutions of AM-NSGA-III-c, the number of RSUs is less than that in other algorithms. The reason is the same as when dealing with high-density scenarios. Correspondingly, the latency time in AM-NSGA-III-c is still larger than that in other algorithms. To analyze all algorithms' performance, we use the dominance rule to compare the feasible solutions and obtain a pareto set summarized in Table \ref{tab:new_pareto2}. In Table \ref{tab:new_pareto2}, we will find that all feasible solutions obtained by MOEA/D are dominated by other algorithms, which means the quality of MOEA/D's solutions is not competitive. In addition, AM-NSGA-III-c (with 10 solutions) still has more solutions than AM-NSGA-III (with 8 solutions) and NSGA-III( with 3 solutions). This indicates that, in low-density scenarios, AM-NSGA-III-c can still generate more feasible solutions than the other algorithms. The number of feasible solutions in NSGA-III is the lowest, which means that NSGA-III's ability to pursue enough feasible solutions is still incompetent. For AM-NSGA-III, the number of feasible solutions is evenly matched with the number in AM-NSGA-III-c, which means that the constraints in low-density obstacle areas are relatively loose. The number of RSUs in AM-NSGA-III ranges from 19 to 29. The corresponding latency time will neither be too large nor too small.  Although the number of RSUs in AM-NSGA-III-c is relatively small, the burden of RSUs will be heavy and therefore cause a high delay in communications. Based on the analysis, the solutions proposed by AM-NSGA-III will be appropriate choices. However, if the decision makers give priority to construction cost in their considerations, AM-NSGA-III-c is preferred. Meanwhile, in time-sensitive areas, the solutions of AM-NSGA-III-c also achieve a relatively low delay. The reason for this is that AM-NSGA-III-c does not deploy RSUs too densely, which avoids the decrement in utilization rate of RSUs and waste of communication resources caused by overcrowded RSU deployment. With a small number of RSUs, AM-NSGA-III-c is competent to cover a large area and provide communication services for more vehicles. 	
	
    To provide visualization comparisons, we also select four results obtained by each algorithm to draw diagrams, which are given in Fig. \ref{fig_sim_nsgaiii_2}, Fig. \ref{fig_sim_eebnsgaiii_2} and Fig. \ref{fig_sim_ieebnsgaiii_2} for NSGA-III, AM-NSGA-III, and NSGA-III-c, respectively. In the figures, there are two latency sensitive areas marked by cyan diamonds, while the dark-blue points are the RSU deployment locations. Based on the deployment locations, it is obvious that the solutions provided by all algorithms can generally cover the traffic flow. In Fig. \ref{fig_sim_nsgaiii_2} and Fig. \ref{fig_sim_eebnsgaiii_2}, the number of RSUs is relatively higher than that in Fig. \ref{fig_sim_ieebnsgaiii_2} and therefore obtains a relatively low time delay according to Table \ref{tab:performance_low_density}. Nevertheless, in Fig. \ref{fig_sim_nsgaiii_2} and Fig. \ref{fig_sim_eebnsgaiii_2}, several RSUs are deployed too closely. In Fig. \ref{fig_sim_ieebnsgaiii_2}, the distance between RSUs is uniformly located in general, and the number of RSUs is relatively low, although each RSU will afford relatively heavy burdens in communications. 
	
	\subsubsection{Analysis of Algorithms with Increased Delay Sensitive Areas}	
    We conducted a comparative study of the algorithms MOEA/D, NSGA-III, AM-NSGA-III, and AM-NSGA-III-c in the same high-density scenario, where the number of delay-sensitive areas increased to 6 and 10, respectively. For the case with 6 latency-sensitive regions, the results are given in Table \ref{tab:performance_NSGA-III-c-6}. The table reveals that neither MOEA/D nor NSGA-III can yield feasible solutions, indicating that an increment in the number of latency-sensitive areas poses a greater challenge for algorithmic resolution. Our proposed algorithms continue to demonstrate proficiency in achieving feasible solutions. Correspondingly, the pareto solutions of AM-NSGA-III and AM-NSGA-III-c can be found in Table \ref{tab:performance_NSGA-III-c-6}, which showcases the performance of the proposed algorithms. In addition, we include four diagrams illustrating the spatial arrangements in Appendix \ref{sec:app:increase_latency}. We also select four figures for each algorithm to demonstrate the spatial deployment, shown in Fig. \ref{fig_nsga_6} and Fig. \ref{fig_am_nsga_6} for AM-NSGA-III and AM-NSGA-III-c, respectively.
%	\begin{table*}[t]
%		\centering
%		\caption{Performance Comparison of Algorithms}
%		\label{tab:performance_comparison_2}
%		\begin{tabularx}{\textwidth}{XXXXX}
%			\toprule
%			\textbf{Metric} & \textbf{MOEA/D} & \textbf{NSGA-III} & \textbf{AEEBNSGA-III} & \textbf{IAEEBNSGA-III} \\
%			\midrule
%			NPS & 18 & 25 &\textbf{30} & 28 \\
%			NFS & 0 & 0 &4 & \textbf{7}\\
%			IGD &0.55675 &0.44755 & 0.39800 &  \textbf{0.24572} \\
%			HV &3.55019 & \textbf{2.44880} & 3.32717 & 2.89005 \\
%			S-Metric & 0.06117 &\textbf{0.04938}  & 0.06051 & 0.05156 \\
%			% \multicolumn{2}{l}{Long Text Here} & Data 10 & Data 11 \\
%			\bottomrule
%		\end{tabularx}
%	\end{table*}	
	\begin{table}[t]
		\centering
		\caption{The performance of feasible solutions of AM-NSGA-III and AM-NSGA-III-c on three objectives.}
		\label{tab:performance_NSGA-III-c-6}
		\begin{tabular}{cccc}
			\toprule
			\textbf{Algorithm} &\textbf{Objective 1} & \textbf{Objective 2} & \textbf{Objective 3} \\
			\midrule
			\multirow{4}{*}{AM-NSGA-III}			
			&225631.26                    & 0.845629        & 35  \\
			&232364.41                   & 0.93568           & 30 \\
			&284269.73                   & 0.98531            & 27 \\
			&346548.37                   & 0.94433            & 20 \\	
			\hline
			\multirow{7}{*}{AM-NSGA-III-c}
			&200765.36                    & 0.79352        & 34  \\
			&226954.83                  & 0.73525           & 29 \\
			&254693.25                   & 0.80863          & 25  \\
			&286536.79                   & 0.80239          & 22  \\
			&302658.57                  & 0.80161          & 20  \\
			&306598.74                    & 0.83094        & 19  \\
			&334569.42                    & 0.85365        & 17 \\		
			% \multicolumn{2}{l}{Long Text Here} & Data 10 & Data 11 \\
			\bottomrule
		\end{tabular}
	\end{table}
	
    For the case with 10 latency-sensitive regions, among the algorithms, only AM-NSGA-III-c obtains feasible solutions shown in Table \ref{tab:performance_AM-NSGA-III-c-10}, while MOEA/D, NSGA-III, and AM-NSGA-III failed to find any feasible solutions. The results clearly indicate the superiority of AM-NSGA-III-c in handling multi lantency-sensitive areas. In Appendix \ref{sec:app:increase_latency}, the spatial deployment of AM-NSGA-III-c for this case are shown in Fig. \ref{fig_am_nsga_10}. Since there only obtains 5 feasible solutions, we show all of them in the figure. 
%    A comparison of the optimization objectives under the same number of RSUs in  reveals the following. First, the total delay tends to decrease slightly as the number of delay-sensitive areas increases. This phenomenon is likely due to the intensified focus on minimizing delay within these sensitive areas, leading to a better global delay minimization strategy. Conversely, the maximum delay within the delay-sensitive areas tends to increase with the number of such areas. The same quantity of RSUs struggles to fully cover more delay-sensitive areas, leading to potential gaps in coverage and thereby resulting in higher maximum delays, according to Fig.\ref{fig_am_nsga_10}.		
	\begin{table}[!htbp]
		\centering
		\caption{The performance of feasible solutions of AM-NSGA-III-c on three objectives with 10 delay sensitive areas.} 
		\label{tab:performance_AM-NSGA-III-c-10}
		\begin{tabular}{cccc}
			\toprule
			\textbf{Algorithm} & \textbf{Objective 1} & \textbf{Objective 2} & \textbf{Objective 3} \\
			\midrule
			\multirow{5}{*}{AM-NSGA-III-c}
			&347348.6  & 1.06849 & 13\\
			&283862.3 & 0.92926  & 19\\
			&280072.1  & 0.82802 & 22\\
			&271401.7  & 0.87875 & 24\\
			&227543.7  & 0.80335 & 27\\
			% \multicolumn{2}{l}{Long Text Here} & Data 10 & Data 11 \\
			\bottomrule
		\end{tabular}
	\end{table}
	According to the above analysis, when evaluating algorithms' performance across different numbers of latency sensitive areas, a trend was observed: the number of feasible solutions decreases as the number of  latency sensitive areas increases. This is due to the increased complexity and restrictions brought about by the greater number of  latency sensitive areas.    
	
	\subsection{Evaluations on Data Offloading Strategy Performance}
    To verify the effectiveness of the proposed data offloading strategy, we compare IBRSG with five other different data offloading strategies in terms of total delay for all vehicles and the load of all RSUs. The experimental scenario and related parameter settings are based on the case of the high-density urban environment. The illustrations for the offloading strategies are presented as follows. 
	\begin{enumerate}
		\item IBRSG data offloading strategy: The proposed strategy in this paper.
		\item Location-based data offloading strategy (marked by ``Mindis"): Offloading vehicle tasks to the nearest RSU \cite{lucas2013}.
		\item Signal strength-based data offloading strategy (marked by ``MinPL"): Offloading vehicle tasks to the RSU with the strongest current signal strength \cite{yiz2017}.
		\item Random data offloading strategy (``marked by Random"): Offloading vehicle tasks to a random RSU \cite{guoc2022}.
		\item Genetic algorithm-based data offloading strategy (``marked by GA"): Allocating vehicle offloading data to an RSU according to a genetic algorithm \cite{jhak2022}.
		\item Multi-metric criteria-based data offloading strategy (``marked by MCDM"): Offloading vehicle tasks to the most optimal RSU based on a weighted combination of multiple metrics \cite{ghorai2018}.
	\end{enumerate}
	
	The metrics to compare the six data offloading strategies include total latency-time, network load balance, and convergence speed, which are explained in Appendix \ref{sec:app:Data_offloading}. The comparison results are given as follows.  	

	\subsubsection{Comparison of Total Latency}
\begin{figure}[!htbp]
    	\centering
    	\includegraphics[width=2.5in]{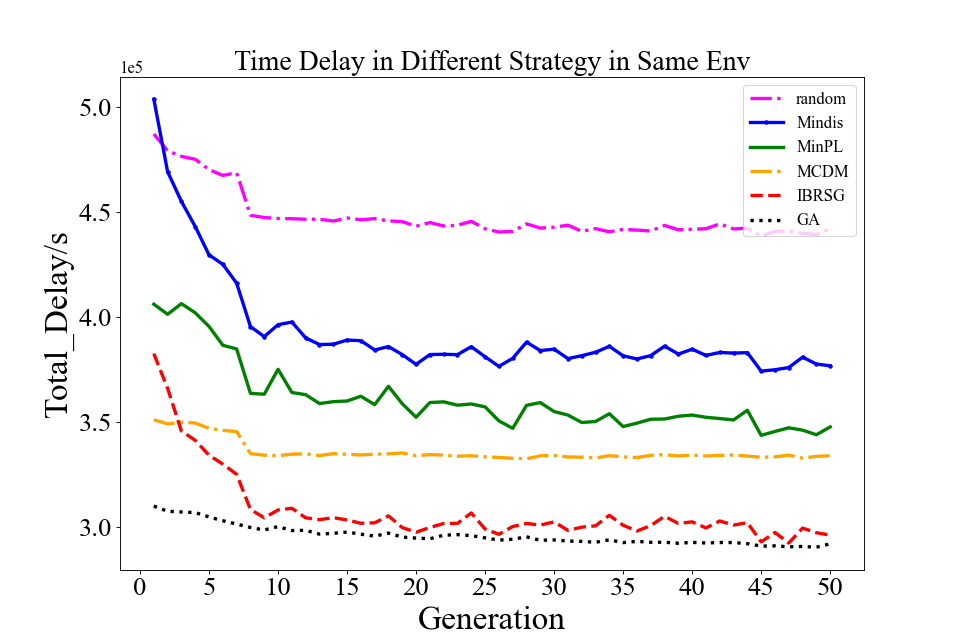}
    	\caption{Time delay in different strategy in same environment.}
    	\label{2_delay}
    \end{figure}

	The latency comparison of data offloading options over iterations is illustrated in Fig. \ref{2_delay}. The IBRSG improves with more iterations, closing the gap with GA and nearing GA. The GA approach initially had an advantage, but the IBRSG eventually caught up. The consistently higher performance of the IBRSG technique suggests it could be a dependable and efficient vehicular network data offloading strategy.
    	
	\subsubsection{Comparison of Load Balance}
\begin{figure}[!htbp]
		\centering
		\includegraphics[width=2.5in]{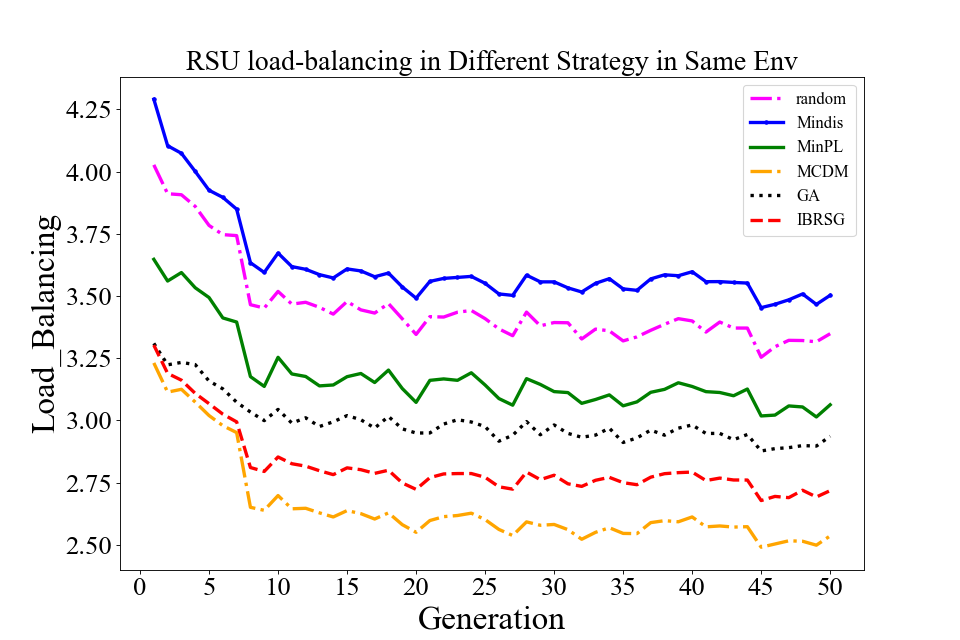}
		\caption{Load balancing in different strategy in same environment.}
		\label{2_load}
	\end{figure}

	Fig.\ref{2_load} reflects a load balance comparison of data offloading options. The MCDM method regularly balances load better than the others,  which optimizes load balance best, followed by the IBRSG strategy. The IBRSG technique outperforms the other four. 

	\subsubsection{Comparison of Time Consumption}
\begin{figure}[!htbp]
	\centering
	\includegraphics[width=2.5in]{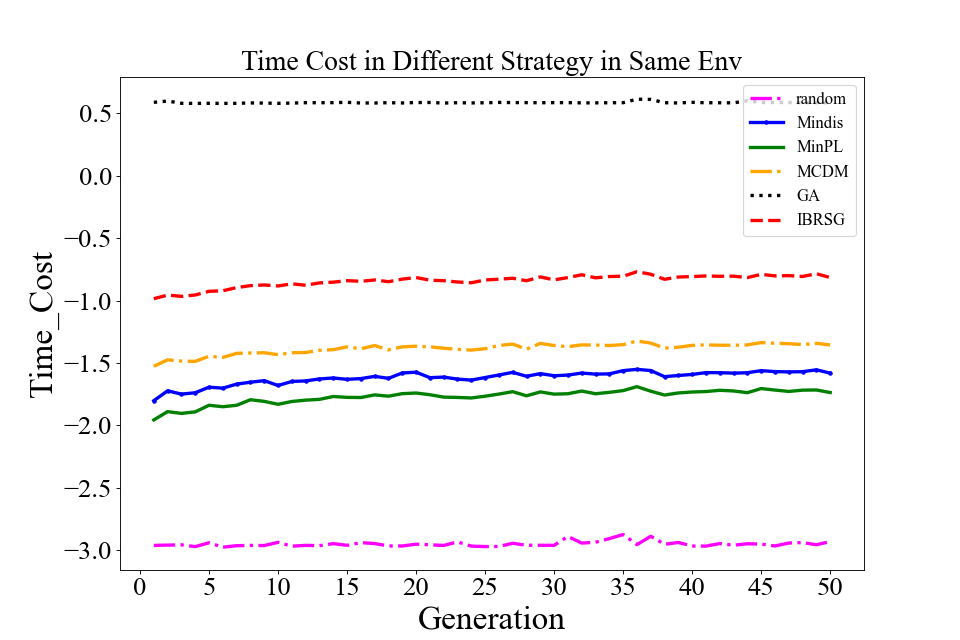}
	\caption{Time cost in different strategy in same environment.}
	\label{2_time}
    \end{figure}

    The comparison of time consumption is shown in Fig. \ref{2_time}, where the values are marked by logarithmic forms, $\log_{10}Time\_consumption$, for an intuitive comparison (this is the reason for the negative values in the figure). Although the random technique has the lowest calculation overhead, its performances on both total latency and load balance deteriorate. The IBRSG strategy computes the second fast.
	
	\subsubsection{Summary of the Data Offloading Strategies}
    In summary, the IBRSG strategy offers a well-balanced performance across key metrics. It is consistently better at minimizing delays and balancing loads than most other strategies, which shows that it has the potential to be a reliable and efficient data offloading strategy. This shows that the IBRSG strategy is a better choice for real-world use in vehicular networks.
		
	\section{Conclusions and Future Work}
	\label{sec:conlusions}	
    This research examines the difficulties associated with the multi-objective optimal deployment of roadside units in urban vehicular networks, with the aims of enhancing communication efficiency and reducing the deployment cost of RSUs. This study proposes two versions of multi-objective optimization algorithms that build on the improvements made to NSGA-III. These algorithms are meant to deal with the problems of conflicting optimization objectives, obstacles in urban environments, and exploring a large-scale optimization space. The algorithms proposed in this study incorporate a multi-population method, an adaptive exploration technique, and an offspring calibration mechanism within the NSGA-III framework. These components are utilized to modify the distribution and density of RSUs. A unique data offloading mechanism is also proposed by designing an iterative best response-based sequential game (IBRSG) approach. The experimental performances of the proposed algorithms are compared and analyzed in both high-density and low-density urban scenarios. A comparison of results is made with state-of-the-art algorithms, and the results demonstrate the superiority and feasibility of the proposed algorithms. In future work, we will explore the integration of other emerging technologies, such as edge computing and machine learning, to further enhance the performance of vehicular networks in urban environments.
    
  \section*{Acknowledgments}
  Thanks for Mr. Wanli CAI for his helps in providing experiment environments to run and debug the algorithms. This work also get guidance and comments from Professor Lei WANG in Tongji University, Shanghai, China.
  
   \bibliographystyle{unsrt} 
   \bibliography{rfere}
   \vspace{-50pt}
	\begin{IEEEbiography}[{\includegraphics[width=1in,height=1.25in,clip,keepaspectratio]{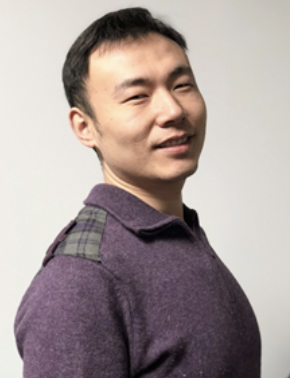}}]{Weian Guo}
received the M.Eng. degree in navigation, guidance, and control from Northeastern University, Shenyang, China, in 2009, and the doctor of engineering degree from Tongji University, Shanghai, China, in 2014. From 2011 to 2013, he was sponsored by China Scholarship Council to carry on his research at the Social Robotics Laboratory, National University of Singapore. He is currently an associate Professor with the Sino-German College of Applied Science, Tongji University. His interests include computational intelligence, optimization, artificial intelligence, and control theory.
\end{IEEEbiography}
	\vspace{-50pt}
	\begin{IEEEbiography}[{\includegraphics[width=1in,height=1.25in,clip,keepaspectratio]{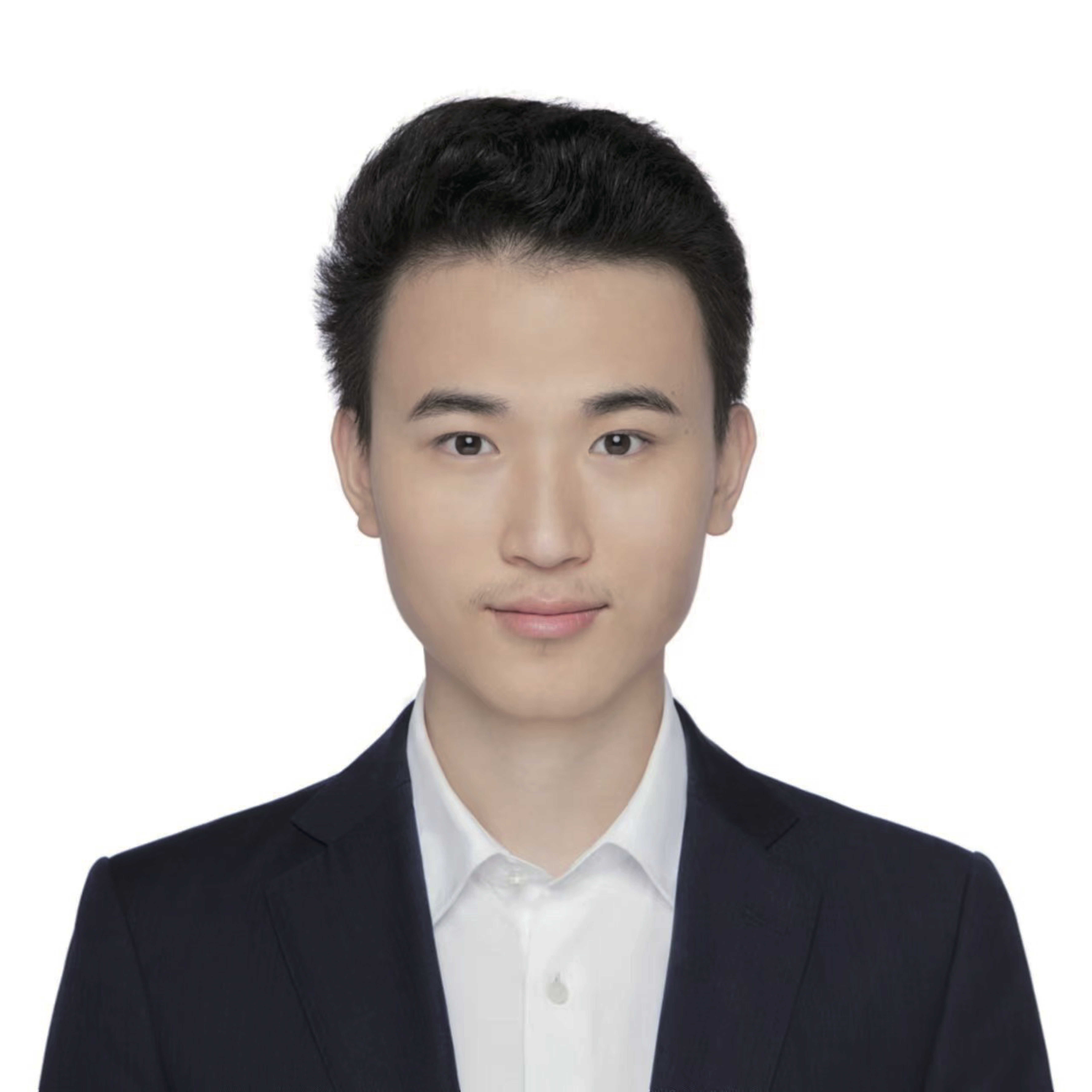}}]{Zecheng Kang}
received the B.S. degree in automotive engineering from Tongji University, Shanghai, China, in 2021, and he is currently pursuing a Master's degree in intelligent science and technology at Tongji University. His research interests are focused on heuristic algorithms, deep learning, and their applications in intelligent transportation system.
\end{IEEEbiography}
	
\vspace{-50pt}
\begin{IEEEbiography}[{\includegraphics[width=1in,height=1.25in,clip,keepaspectratio]{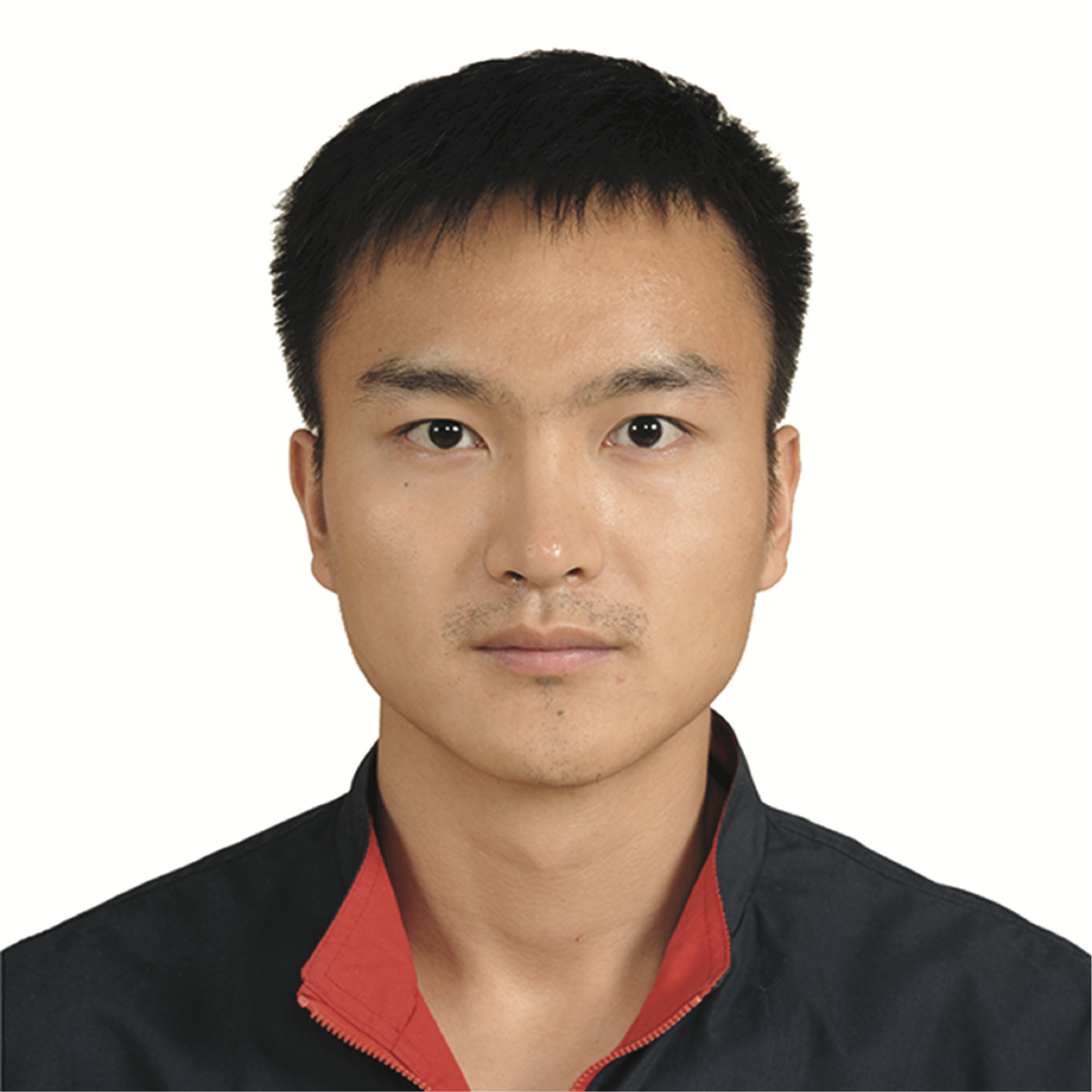}}]{Dongyang Li}
received the M.S. and Ph.D degrees in school of electronics and information engineering, Tongji University, Shanghai, China, in 2017 and 2022, respectively. From 2019 to 2021, he was sponsored by China Scholarship Council to carry on his research at Georgia Institute of Technology. He is now an engineer with the Sino-German College of Applied Science, Tongji University; His research interest includes computational intelligence, deep learning and their applications.
\end{IEEEbiography}
\vspace{-300pt}
\begin{IEEEbiography}[{\includegraphics[width=1in,height=1.25in,clip,keepaspectratio]{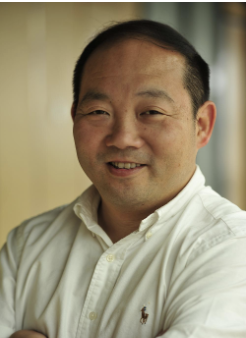}}]{Lun Zhang}
received the B.S. and Ph.D degrees in computer communications, transportation information engineering and control, Tongji University, Shanghai, China, in 1992 and 2005, respectively. He is currently a professor with School of Transportation, Tongji University. His interests include intelligent transportation, computational intelligence, and deep learning.
\end{IEEEbiography}

\vspace{-300pt}
\begin{IEEEbiography}[{\includegraphics[width=1in,height=1.25in,clip,keepaspectratio]{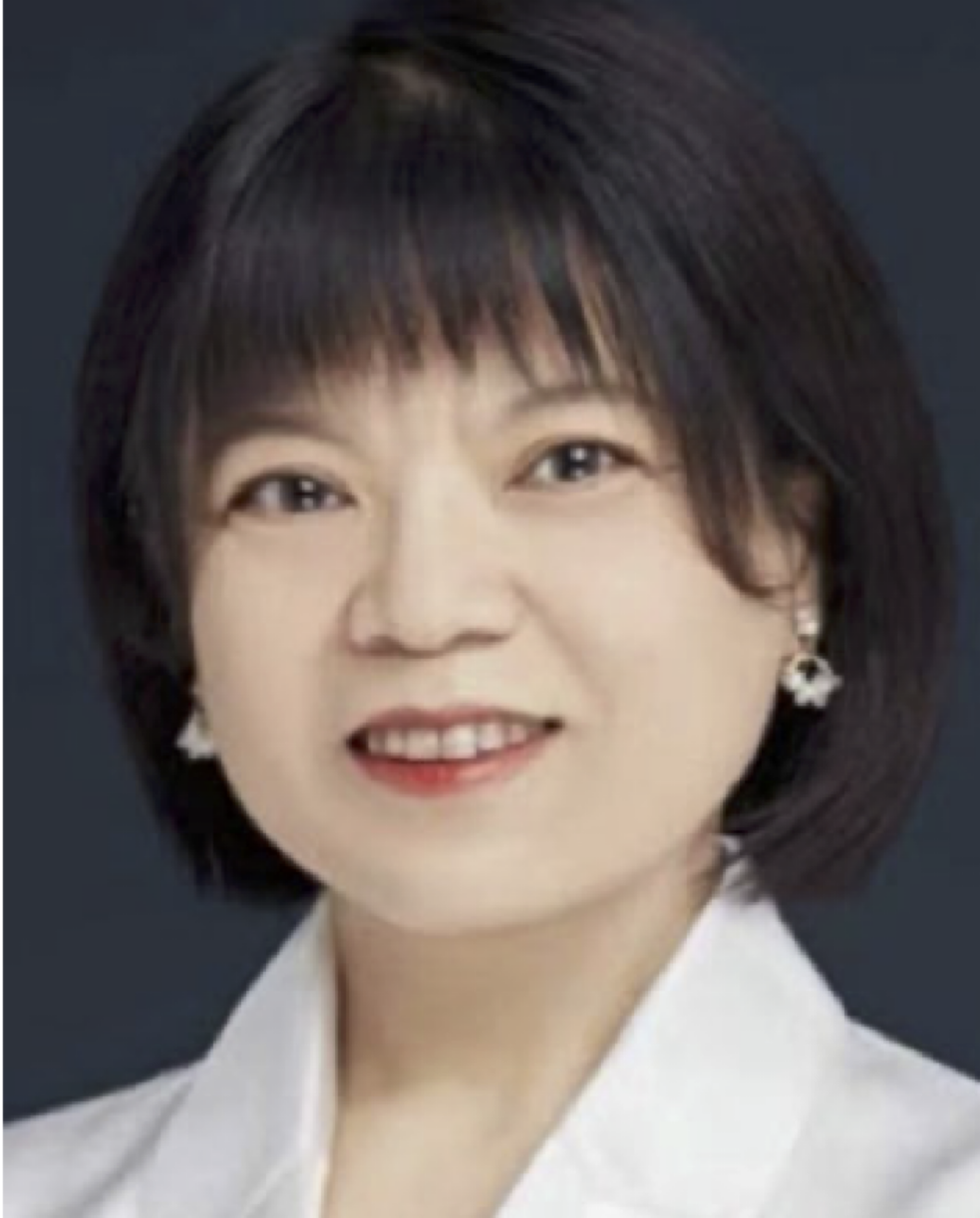}}]{Li Li}
received the B.S. and M.S. degrees in electrical automation from Shengyang Agriculture University, Shengyang, China, in 1996 and 1999, respectively, and the Ph.D. degree in mechatronics engineering from the Shenyang Institute of Automation, Chinese Academy of Science, Shenyang, in 2003.

She joined Tongji University, Shanghai, China, in 2003, where she is currently a professor of control science and engineering. She has over 50 publications, including 4 books, over 30 journal papers, and 2 book chapters. Her current research interests include production planning and scheduling, computational intelligence, data-driven modeling and optimization, semiconductor manufacturing, and energy systems
\end{IEEEbiography}

\clearpage
\appendix
\renewcommand\thefigure{A.\arabic{figure}}
\renewcommand\thetable{A.\Roman{table}}
\renewcommand\thealgorithm{A.\Roman{algorithm}}
\setcounter{figure}{0}
\setcounter{table}{0}
\setcounter{algorithm}{0}
%\section*{The Appendix for ``A Large-Scale Multi-Objective Particle Swarm Optimizer With Explicit Balance of Convergence and Diversity"}
This is the appendix for the manuscript of ``Multi-objective Optimal Roadside Units Deployment in Urban Vehicular Networks".
\subsection{Explanations on time delay}
\label{sec:app:explain_time_delay}
Eqs. \eqref{eqn:delay} to \eqref{eqn:que_delay} are complementary to Sect. \ref{sec:objFunc} for computing the commutation delay.

In general, the communication delay consists of transmission delay, propagation delay, and queuing delay \cite{xu2020}. However, in this paper, by gridding an urban map, the communication range is typically less than a few hundred meters, which does not exceed the RSU's communication capability. Therefore, the propagation delay is negligible. Correspondingly, the total delay in this paper is expressed in \eqref{eqn:delay}.
	\begin{equation}
		\label{eqn:delay}
		d_{ij}^t = d_{ij}^{t,trans} + d_{ij}^{t,queue}
	\end{equation}
	where $d_{ij}^{t,trans}$  is the transmission delay, while $d_{ij}^{t,queue}$ is the queuing delay. It is worth mentioning that from $d_{ij}^t$ to $d_i^t$, the setting of $j$ is based on the data offloading strategy, which should be designed in the communication protocol \cite{saleem2021}. In vehicular networks, vehicles offload data to RSUs and request services from RSUs. When vehicles have a chance to connect to more than one RSU, the design of data offloading strategy is much more influential in evaluating network performance, e.g. the calculation of latency time \cite{ghorai2018}. The details in designing data offloading strategy will be presented in Section \ref{sec:algorithm}.
	
	In \eqref{eqn:delay}, $d_{ij}^{t,trans}$ expressed in \eqref{eqn:trans_delay} is affected by packet size $L_p$ and transmission rate $Tr_{ij}^t$.
	\begin{equation}
		\label{eqn:trans_delay}
		d_{ij}^{t,trans} = \frac{L_p}{Tr_{ij}^t}
	\end{equation}
	where $L_p$ is the size of data packet, $Tr_{ij}^t$ is the transmission rate. In this paper, we set $L_p$ as a constant, which means the data is divided into the same length for transmission \cite{Chen2018}.  The way to calculate $R_{ij}^t$ is obtained from \cite{Tang2019}, which is expressed in \eqref{eqn:trans_rate}.
	% which is related to the path loss due to free-space path loss $L_{fs}$ and shadowing loss from obstacles $L_{shad}$\cite{akhtar2014}. 
	%Consequently, the transmission rate, $R_{ij}^t$, can be expressed as:
	\begin{equation}
		\label{eqn:trans_rate}
		Tr_{ij}^t = B\log_2\left(1 + \frac{P_t}{\left(L_{fs}+L_{shad}+N_0\right)B}\right)
	\end{equation}
	where $B$ is the bandwidth, $P_t$ is the transmission power, $L_{fs}$ is the free-space path loss, $L_{shad}$ is the loss caused by shadows, $N_0$ is the noise power spectrum density. In \eqref{eqn:delay}, for the queuing delay $d_{ij}^{t,queue}$ expressed in \eqref{eqn:que_delay}, it is obtained under a M/M/1 queuing model \cite{ravi2020}. 
	\begin{equation}
		\label{eqn:que_delay}
		d_{ij}^{t,queue} = \frac{1}{\mu_j - \lambda_j^t}
	\end{equation}
	where $i$ and $j$ are the indexes of vehicle and RSU respectively, $t$ is the index of time-period, $\mu_j$ is the service rate of RSU $j$, $\lambda_j^t$ is the arrival rate of packets at RSU $j$ at time period $t$.
	
	\subsection{$\epsilon$-level rules}
	\label{sec:app:epsilon_rules}
	The $\epsilon$-level rules is cited from \cite{fan2019}, which is used for comparing two solutions with violations. The value of $\epsilon$ varies with iterations, which can be obtained by \eqref{eqn:epsilon}. 
	\begin{equation}
		\label{eqn:epsilon}
		\epsilon(g) =  \left \{
		\begin{split}
			&\mathbf{rule1:} \phi_{\sum\theta},\ \textbf{if} \ g=0\\
			&\mathbf{rule2:} (1-\tau)\epsilon(g-1),\ \textbf{if} \ \rho_g<\alpha \ \textbf{and} \ g<G\\
			&\mathbf{rule3:} (1+\tau)\phi_{max},\ \textbf{if} \ \rho_g\geq \alpha \ \textbf{and} \ g<G\\
			&\mathbf{rule4:} 0,\ \textbf{if} \ g\geq G
		\end{split}
		\right.
	\end{equation} 
	where $g$ is the index of iterations, $G$ is the maximum limitation of iterations, $\phi_{\sum\theta}$ is the violation sum of top-$\theta$ ranking  individuals in the initial population ($g=0$), $\phi_{max}$ is the maximum overall constraint violation found by far, $\tau \in [0,1]$ plays the role to reduce the constraints relaxation and to control the scale factor multiplied by maximum overall constraint violation. $\rho_g$ is the ratio of feasible solutions at the $g^{th}$ iteration. $\alpha \in [0,1]$ is to control the searching preference between the feasible and infeasible regions.
	
	Based on \eqref{eqn:epsilon}, it is feasible to compare two solutions with constraints violations. Especially when $g < G$,  the value of $\alpha$ can be used to control the tolerance to violations. If $\alpha$ is a large value, $\mathbf{rule2}$ tends to happen. In this case, $\epsilon(g)$ will be smaller than $\epsilon(g-1)$, which means the violation tolerance goes strict. On the contrary, it is probable to select $\mathbf{rule3}$ if $\alpha$ is a small value. In this case, violation tolerance will be promoted. Therefore, it is crucial to design a suitable value of $\alpha$ in the comparisons. 
	
	\subsection{Scenarios and Parameters}
	\label{sec:app:scenarios}
\begin{figure}[!htpb]
		\centering
		\subfloat[\label{fig:high_env}]{
			\includegraphics[width=0.23\textwidth]{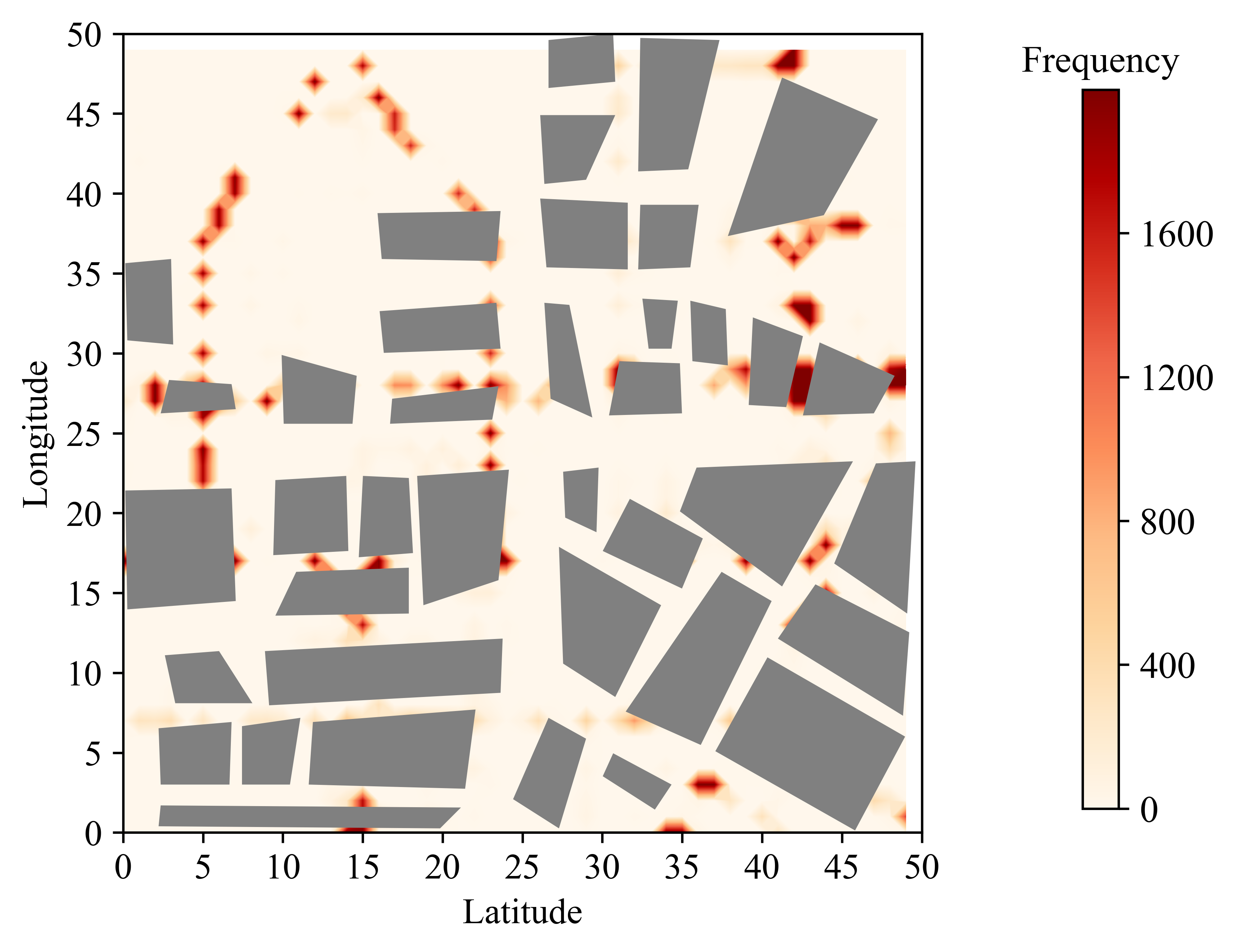}}
		\subfloat[\label{fig:low_env}]{
			\includegraphics[width=0.23\textwidth]{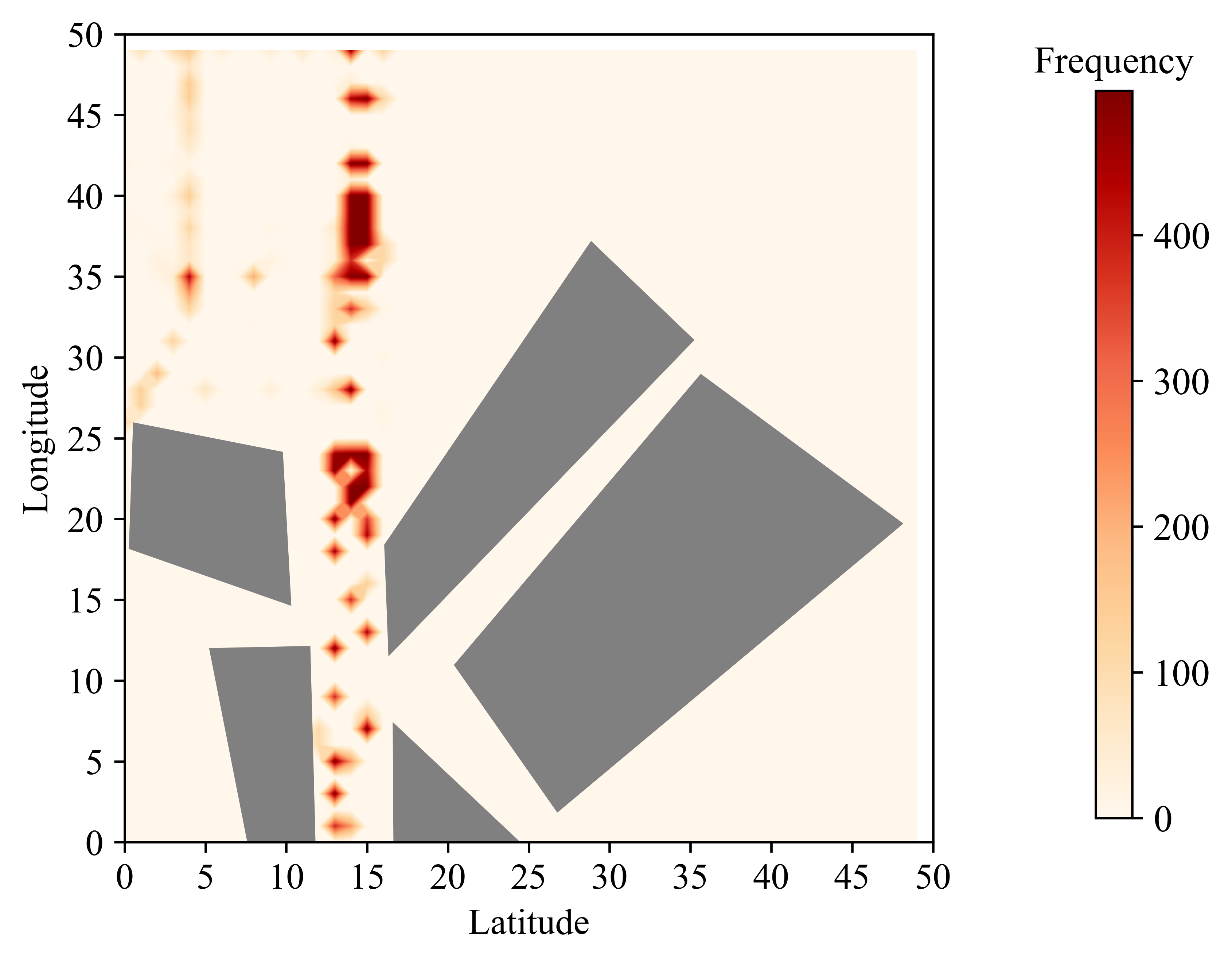}}
		\caption{(a) Scenarios for high-density area. (b) Scenario for low-density area. }
		\label{fig:scenarios_high_low} 
	\end{figure}

    The high-density and low-density scenarios are given in Fig. \ref{fig:scenarios_high_low}. In the high-density scenario, the utilization of Tian-Fu Square is observed, where RSUs cater to a total of 56,704 vehicles. Conversely, in the low-density scenario, the service provision near Du Fu's thatched cottage accommodates 36,405 vehicles. In the figures, the gray part means the obstacles, and the red part denotes the traffic heat marked by vehicles' frequency. The traffic volume is high when the red color is dark, while the light red color means the traffic volume is low. For each scenario, it covers one square kilometer and is uniformly divided into $50\times50$ map pieces, thereby creating 2500 decision variables in the experiments.
	
	For the parameters in urban scenarios, we present Table \ref{tab:parameter_setup} to calculate \eqref{eqn:trans_delay}, \eqref{eqn:trans_rate}, and \eqref{eqn:que_delay}. 
		\begin{table}[!htbp]
		\caption{Simulation Parameter Set-up}
		\label{tab:parameter_setup}
		\centering
		\begin{tabular}{ll}
			\toprule
			\textbf{Parameter} & \textbf{Value} \\
			\midrule
			Packet size($L_p$) & 1Mb \\
			Bandwidth ($B$) & 10 MHz \\
			Transmitting power of RSU ($P_t$) & 23 dBm \\
			Noise power spectrum density ($N_0$) & -174 dBm \\
			Path loss model $L_{fs}$ & Free-space path loss model \\
			Shadowing model $L_{shad}$ & Log-normal shadowing model \\
			RSU service rate ($\mu$) & 20 \\
			Arrive rate ($\lambda$) &Number of arrived vehicles\\
			\bottomrule
		\end{tabular}
	\end{table}	
	In this table, the free-space path loss mode $L_{fs}$ and the log-normal shadowing model $L_{shaw}$ are introduced by \eqref{eqn:L_fs} and \eqref{eqn:L_shaw} respectively, which are cited from \cite{stoya2009} and \cite{ghorai2018}.  $L_{fs}$ is used to access  signal attenuation when no obstacle exists between vehicles and RSUs, while $L_{shaw}$ is used to evaluate the signal attenuation when obstacles exist between vehicles and RSUs. In addition, in the table, for the arrive rate $\lambda$, it is obtained according to the real traffic data.		
	\begin{equation}
		\label{eqn:L_fs}
		L_{\text{fs}} = 20 \log_{10}(dis) + 20 \log_{10}(f_{signal}) - 147.55
	\end{equation}
	where $dis$ is the distance between vehicle and RSU by meters, which is calculated by the real traffic data. $f_{signal}$ is the frequency of the signal, which is set as 5.9GHz in this paper.	
	\begin{equation}
		\label{eqn:L_shaw}
		L_{\text{shad}} = 10 \cdot \sigma \cdot \mathcal{N}(0,1)
	\end{equation}
	where the $\sigma$ is the standard deviation of the shadowing attenuation, which is set as 4dB in this paper, $\mathcal{N}(0,1)$ represents a normal distribution. For other parameters in scenarios, we summarize them as follows. In the communication network described in this paper, vehicles transfer or receive data every 30 seconds. The radius of a latency-sensitive area is 20 meters. For the constraints of RSU density, the minimal distance $D_{min}$ between RSUs is set as 30 meters. In addition, if a vehicle connects to a cellular network, the time delay is set as a constant of 2 seconds\footnote{Since the packet size is 8Mb and the transferring speed of the cellular network in an urban environment is about 4Mbps, the latency of transmitting data through the cellular network is set to a constant value of 2 seconds.}.  
	
	\subsection{Experimental Results for the High-density Scenario}
	\label{sec:app:experiments_high}
\begin{figure*}[!htbp]
    	\centering
    	\subfloat[]{\includegraphics[width=1.7in]{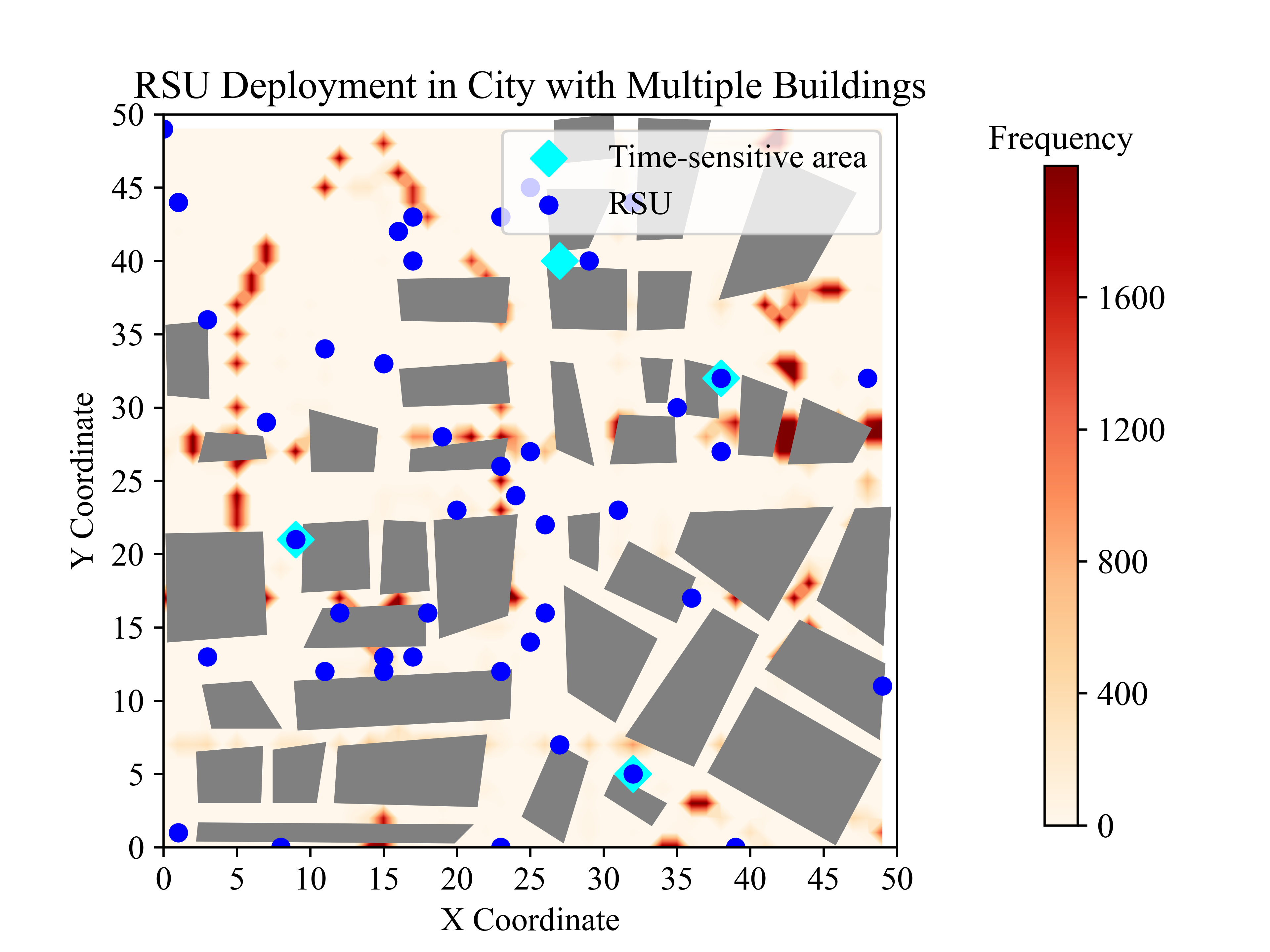}%
    		\label{fig_sim_nsgaiii_1_fig_first_case}}
    	\hfil
    	\subfloat[]{\includegraphics[width=1.7in]{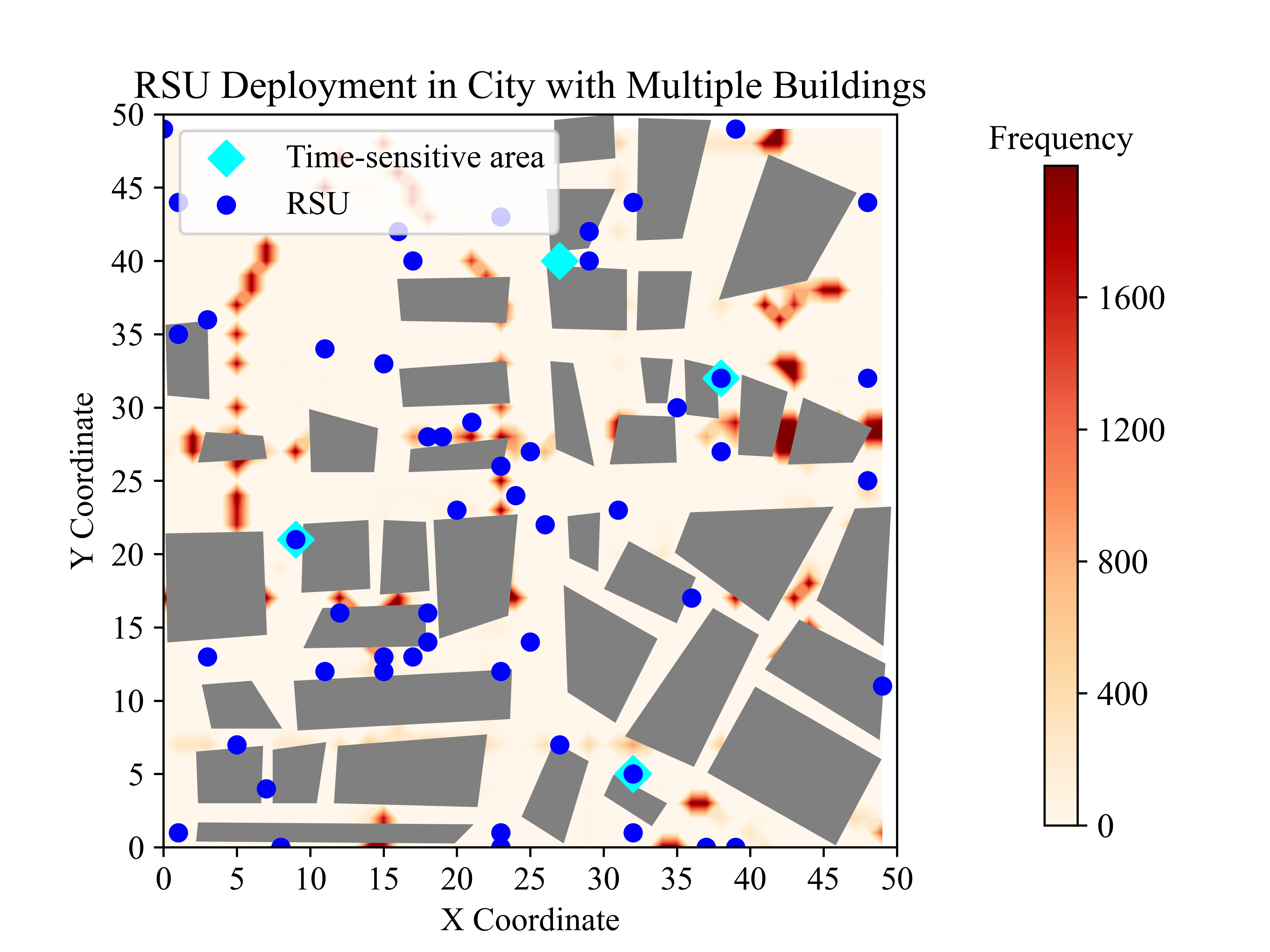}%
    		\label{fig_sim_nsgaiii_1_fig_second_case}}
    	\hfil
    	\subfloat[]{\includegraphics[width=1.7in]{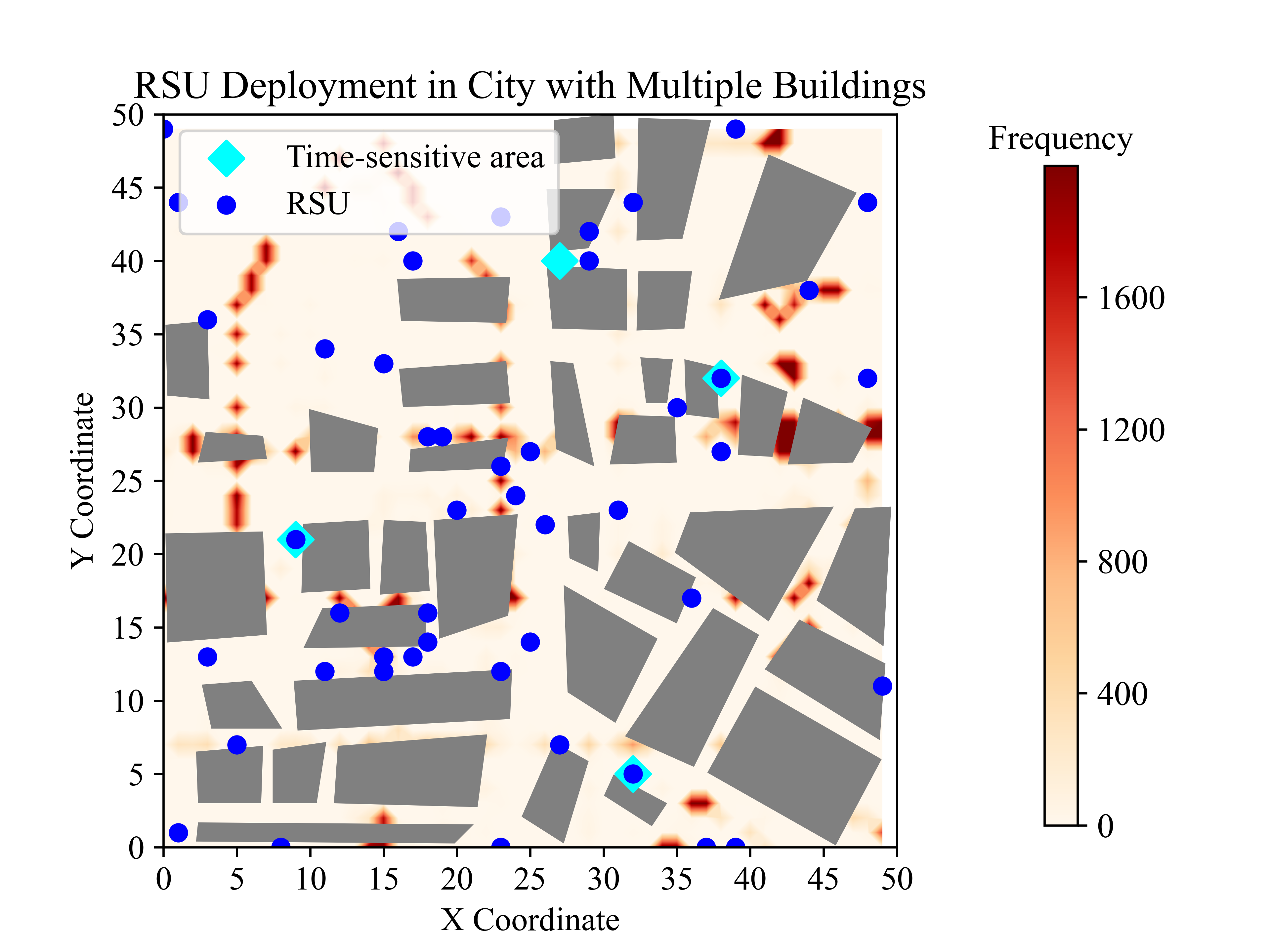}%
    		\label{fig_sim_nsgaiii_1_fig_third_case}}
    	\hfil
    	\subfloat[]{\includegraphics[width=1.7in]{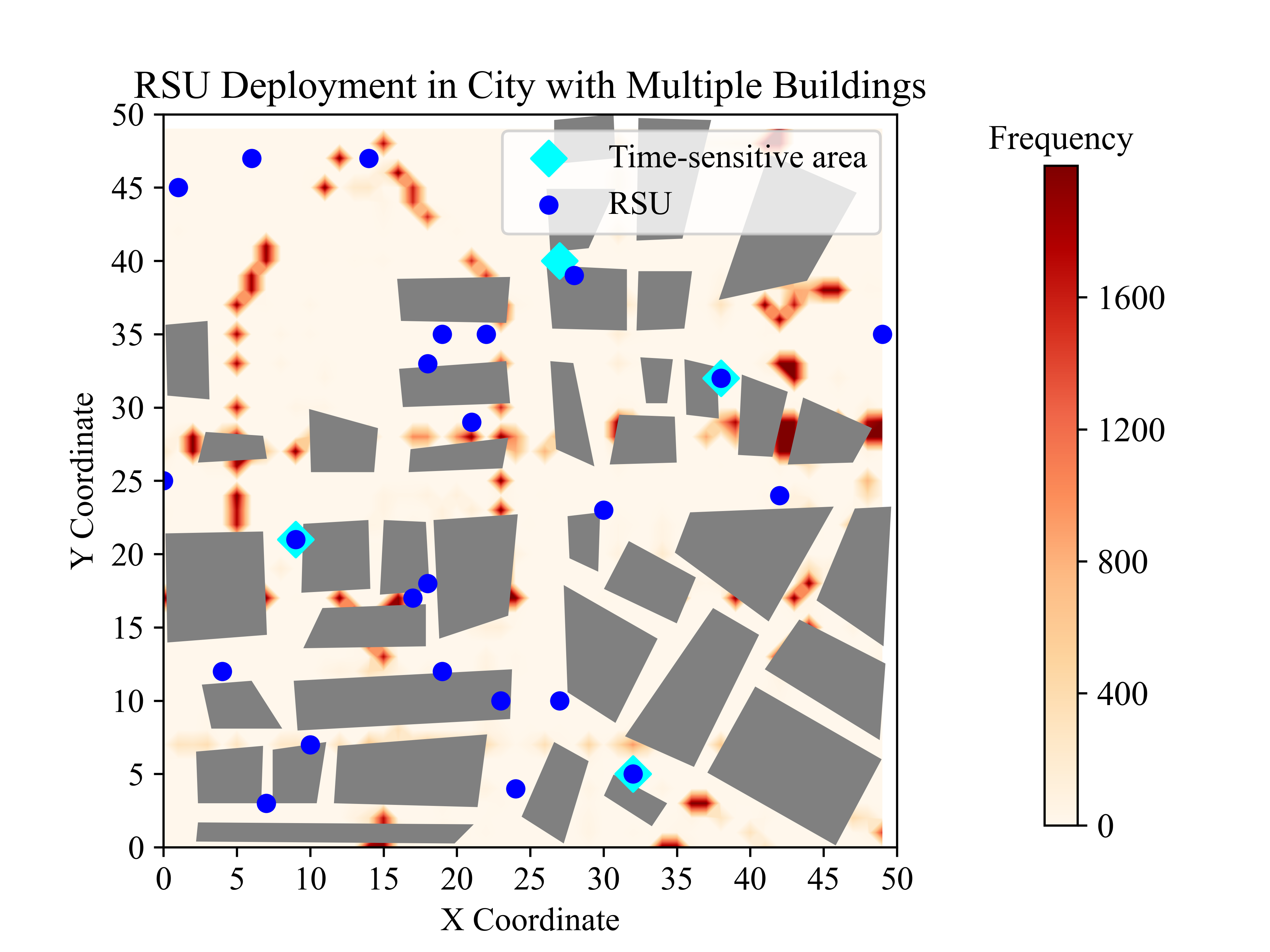}%
    		\label{fig_sim_nsgaiii_1_fig_fourth_case}}
    	\caption{RSUs deployment by NSGA-III in high-density scenario.}
    	\label{fig_sim_nsgaiii_1}
    \end{figure*}

\begin{figure*}[!htbp]
    	\centering
    	\subfloat[]{\includegraphics[width=1.7in]{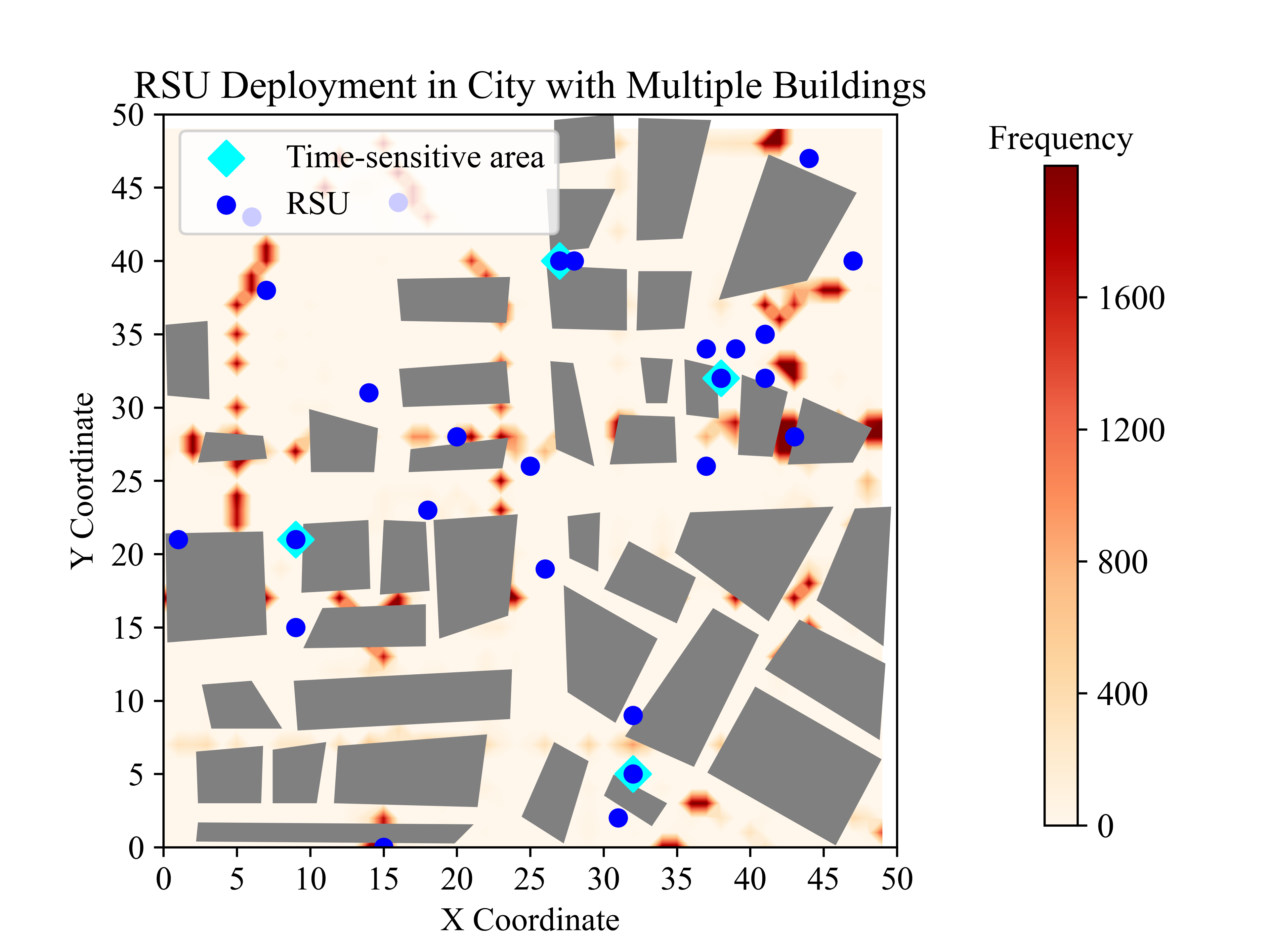}%
    		\label{fig_sim_eebnsgaiii_1_fig_first_case}}
    	\hfil
    	\subfloat[]{\includegraphics[width=1.7in]{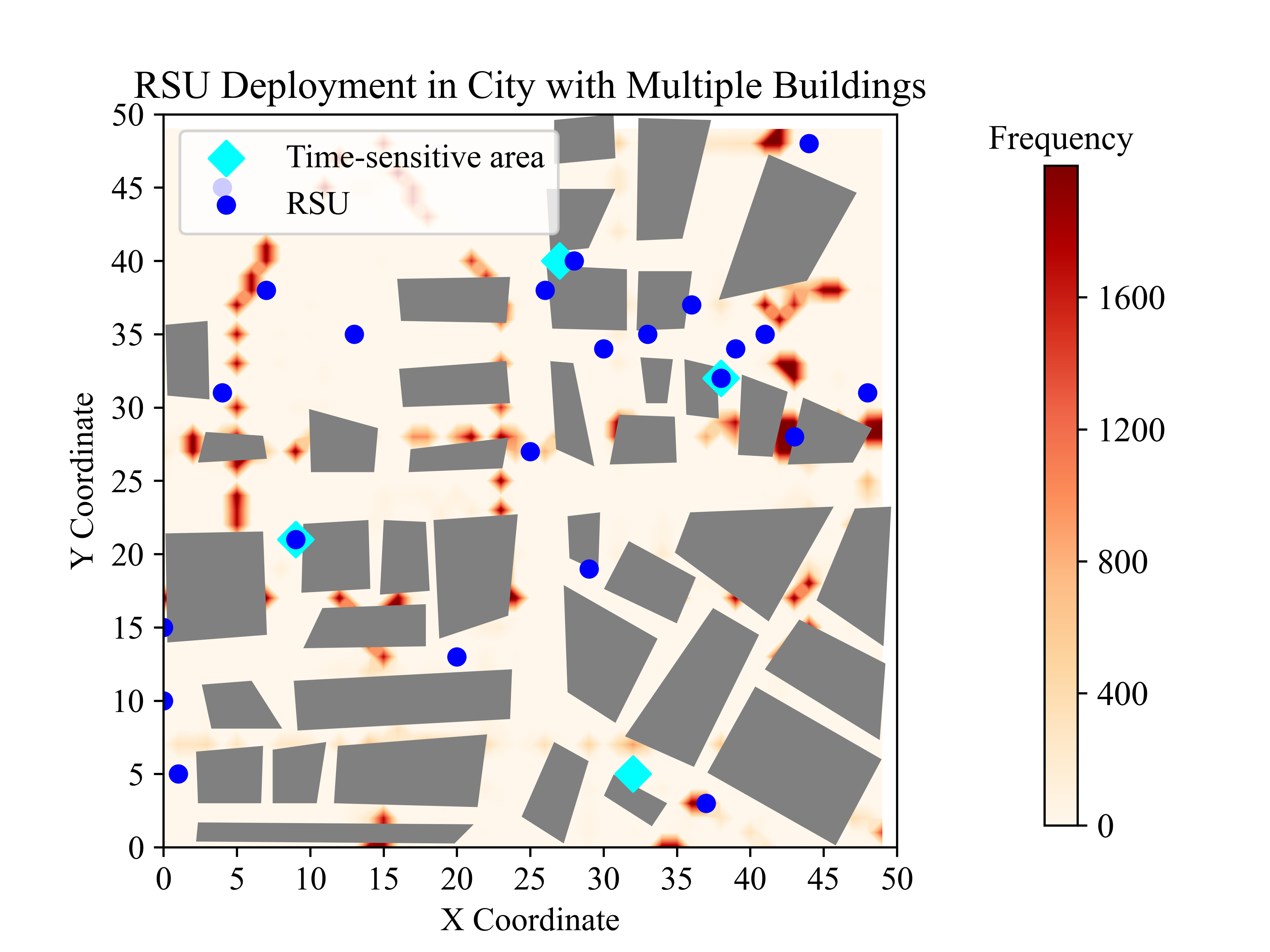}%
    		\label{fig_sim_eebnsgaiii_1_fig_second_case}}
    	\hfil
    	\subfloat[]{\includegraphics[width=1.7in]{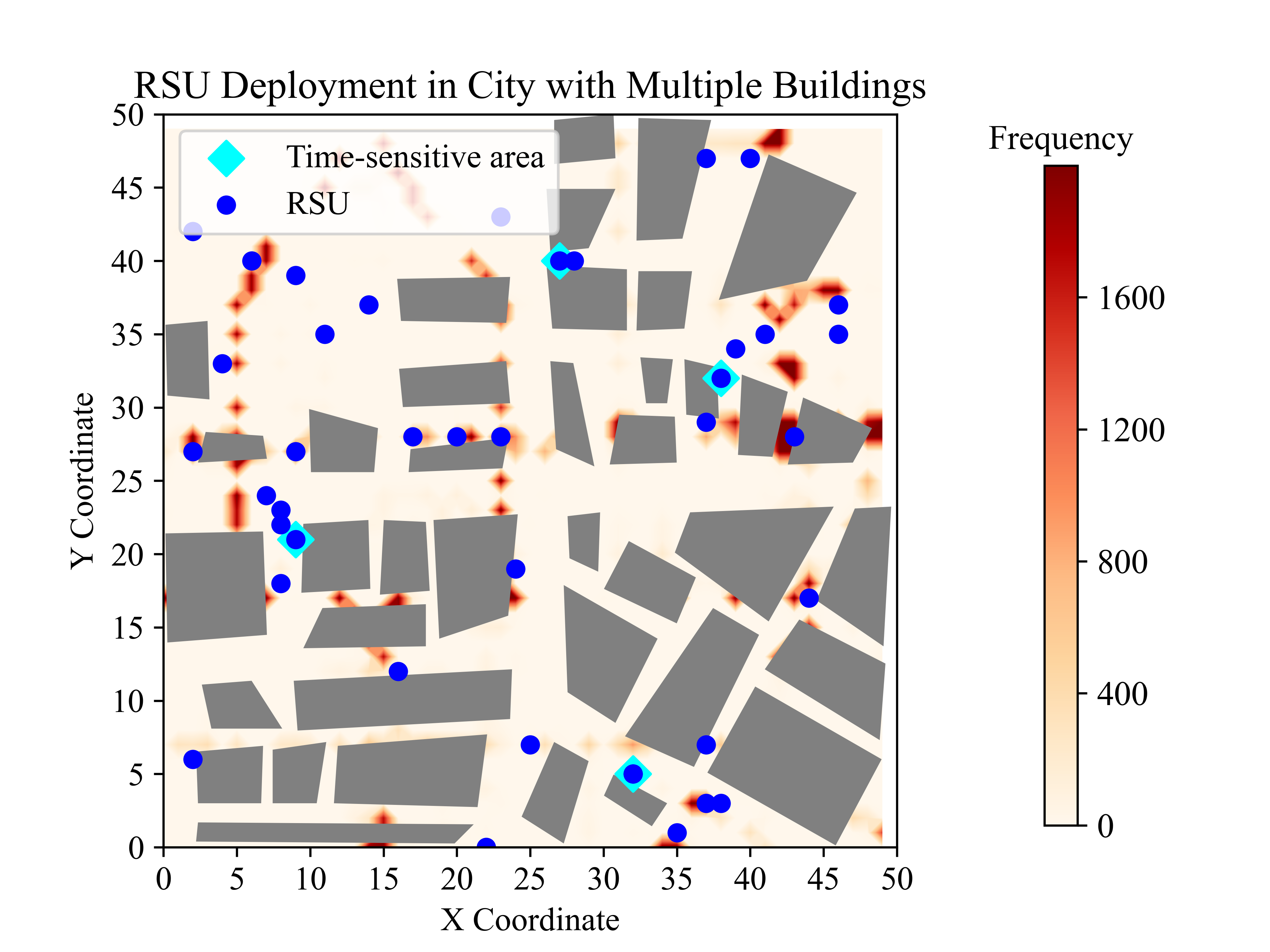}%
    		\label{fig_sim_eebnsgaiii_1_fig_third_case}}
    	\hfil
    	\subfloat[]{\includegraphics[width=1.7in]{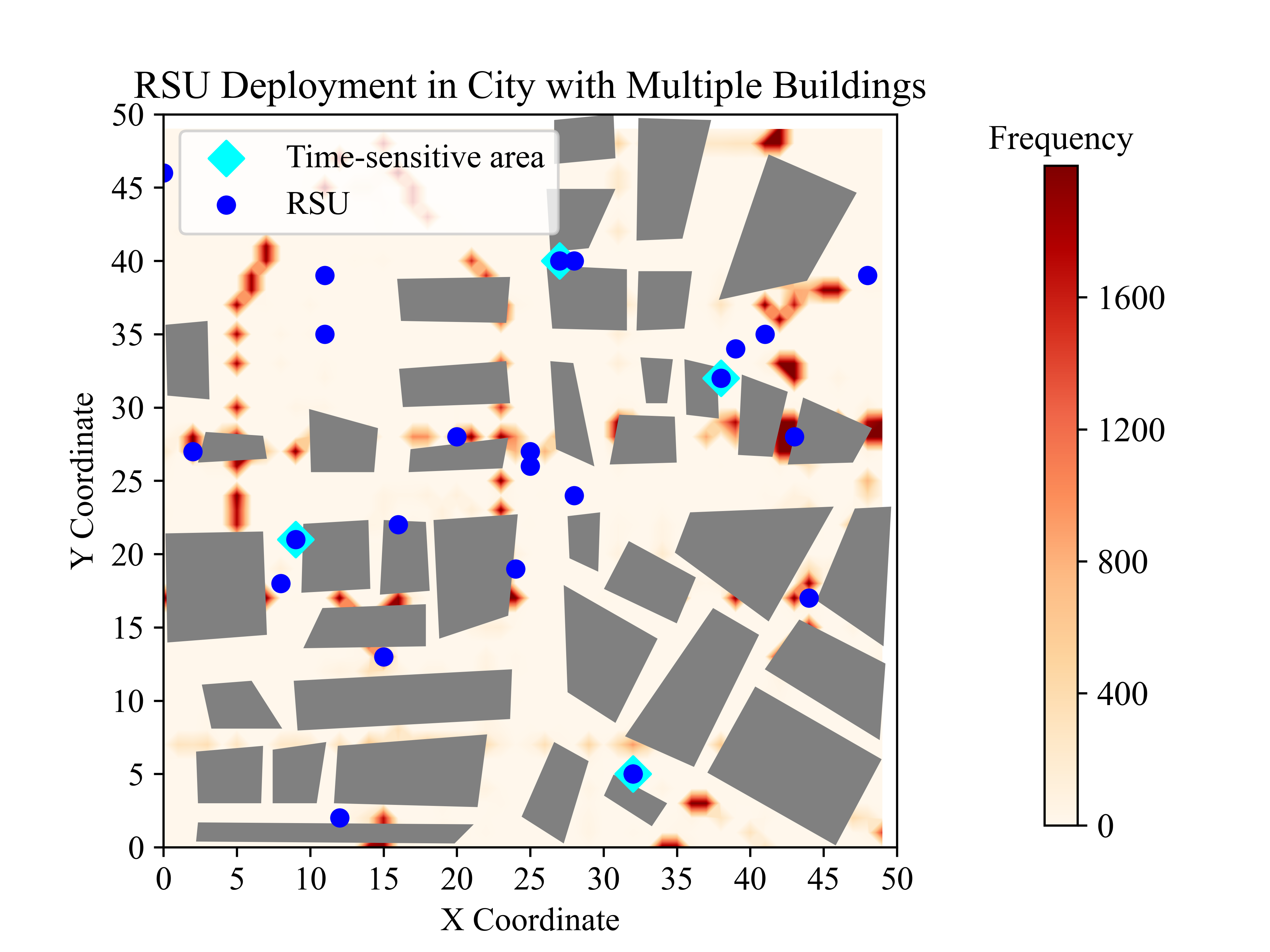}%
    		\label{fig_sim_eebnsgaiii_1_fig_fourth_case}}
    	\caption{RSUs deployment by AM-NSGA-III in high-density scenario.}
    	\label{fig_sim_eebnsgaiii_1}
    \end{figure*}

\begin{figure*}[!htbp]
    	\centering
    	\subfloat[]{\includegraphics[width=1.7in]{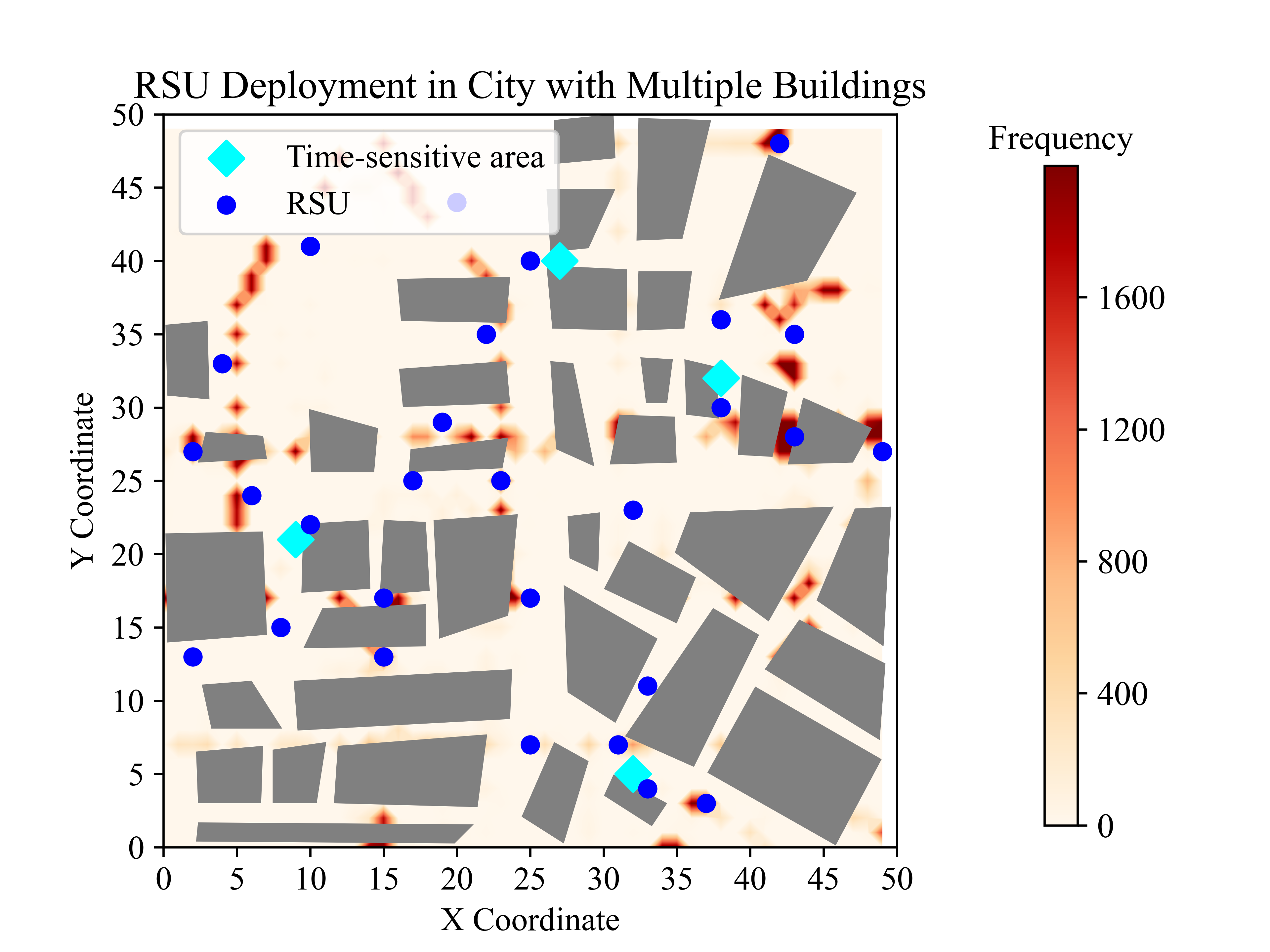}%
    		\label{fig_sim_ieebnsgaiii_1_fig_first_case}}
    	\hfil
    	\subfloat[]{\includegraphics[width=1.7in]{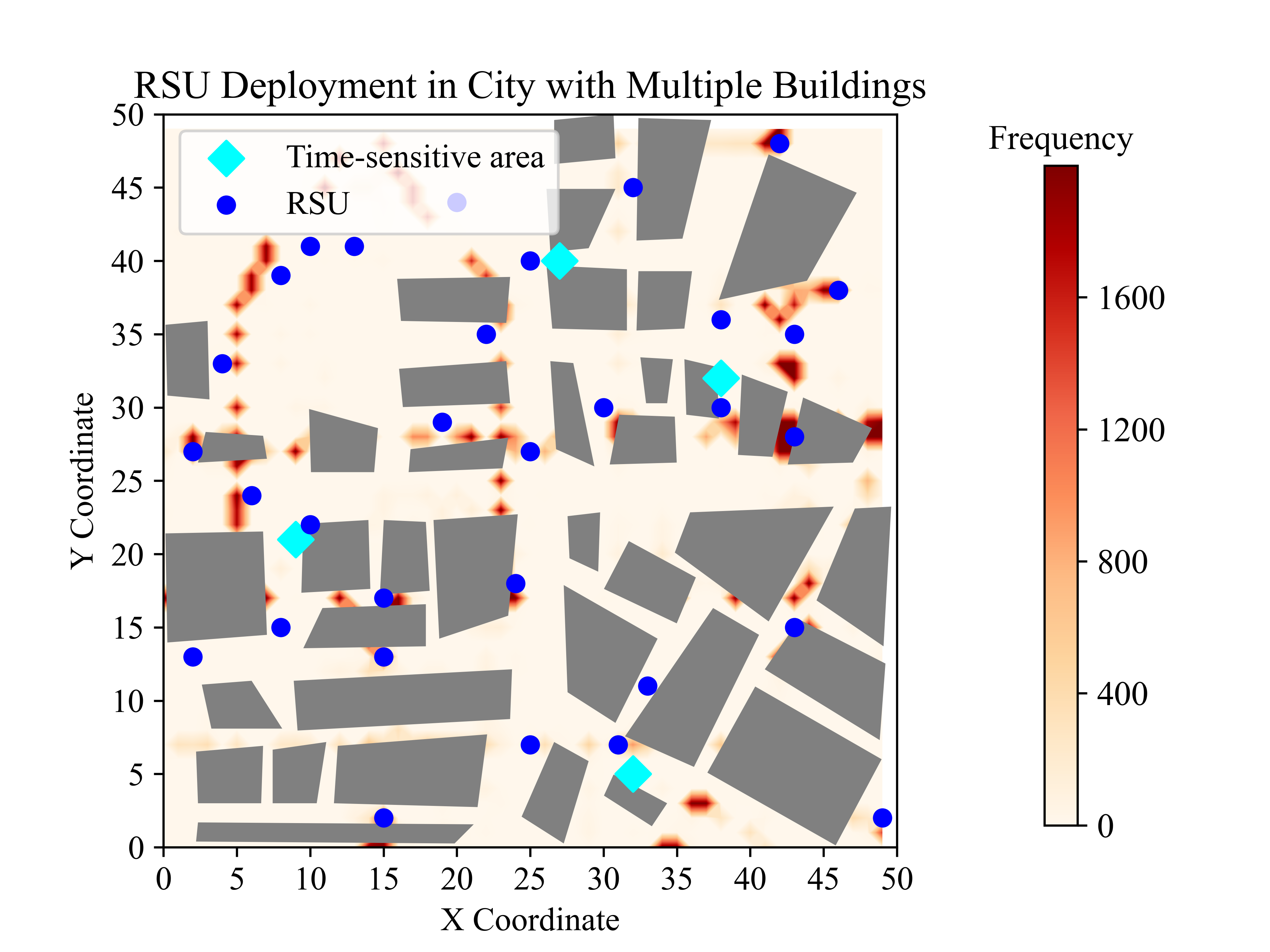}%
    		\label{fig_sim_ieebnsgaiii_1_fig_second_case}}
    	\hfil
    	\subfloat[]{\includegraphics[width=1.7in]{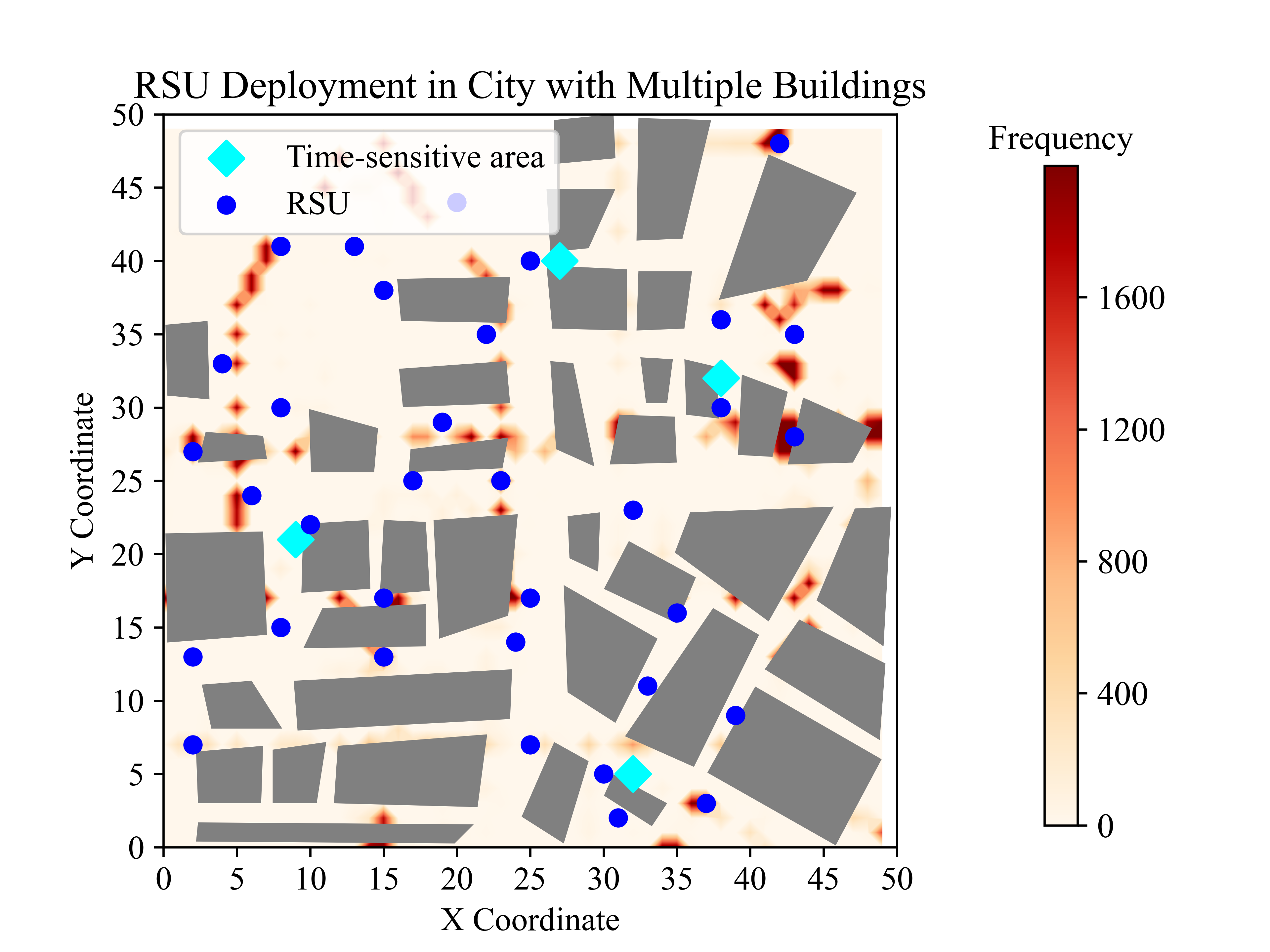}%
    		\label{fig_sim_ieebnsgaiii_1_fig_third_case}}
    	\hfil
    	\subfloat[]{\includegraphics[width=1.7in]{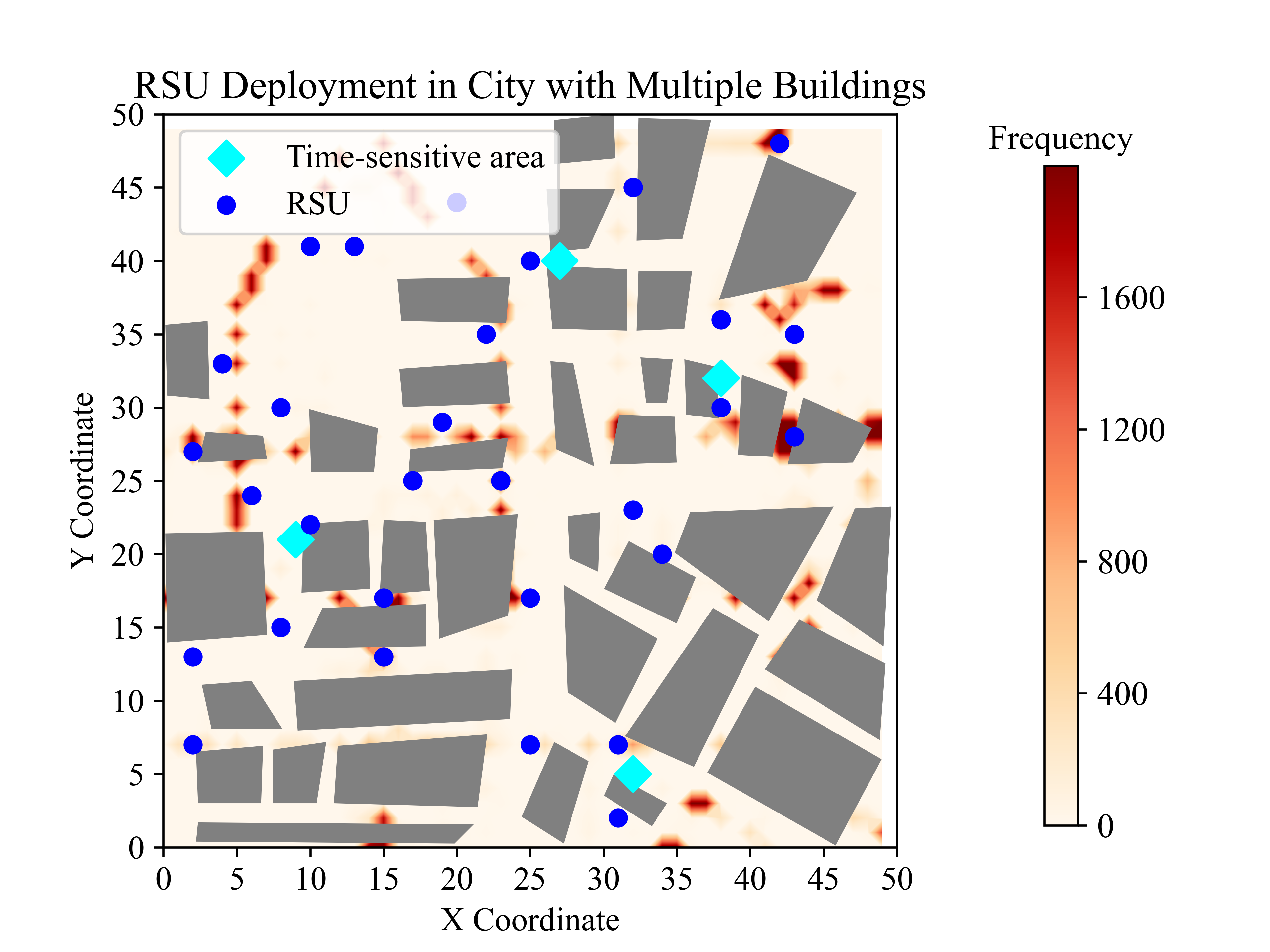}%
    		\label{fig_sim_ieebnsgaiii_1_fig_fourth_case}}
    	\caption{RSUs deployment by AM-NSGA-III-c in high-density scenario.}
    	\label{fig_sim_ieebnsgaiii_1}
    \end{figure*}

	In high-density scenario, the feasible solutions obtained by NSGA-III, AM-NSGA-III, and AM-NSGA-III-c are presented in Table \ref{tab:performance_objectives}. Based on the data in such table, we compare all the feasible solutions to compose a pareto front shown in Table \ref{tab:pareto_all_high}. For each algorithm, we present four diagrams to show the RSU deployment in Fig. \ref{fig_sim_nsgaiii_1}, Fig. \ref{fig_sim_eebnsgaiii_1} and Fig. \ref{fig_sim_ieebnsgaiii_1}. 
	\begin{table}[!htbp]
		\centering
		\caption{The feasible solutions obtained by NSGA-III, AM-NSGA-III and AM-NSGA-III-c for the high-density scenario}
		\label{tab:performance_objectives}
		\begin{tabular}{cccc}
			\toprule
			\textbf{Algorithms} & \textbf{Objective 1} & \textbf{Objective 2} & \textbf{Objective 3} \\
			\midrule
			\multirow{4}{*}{NSGA-III}
			&268975.4  & 0.74705 & 52 \\
			&270625.3  & 0.78494 & 47 \\
			&292602.1  & 0.82984 & 43 \\
			&347545.7  & 0.84349 & 24 \\
			\hline
			\multirow{15}{*}{AM-NSGA-III}    
			&205928.2  & 0.76515 & 44\\
			&223522.9  & 0.74558 & 41 \\
			&224745.4  & 0.76874 & 39 \\
			&235735.0  & 0.74113 & 39\\
			&241672.7  & 0.78226 & 36\\
			&262040.6  & 0.82293 & 33 \\
			&267599.3  & 0.79763 & 31 \\
			&269848.4 & 0.81164  & 30\\
			&284045.1 & 0.80584 & 30 \\
			&288118.1 & 0.80405 & 30 \\
			&288685.4  & 0.82349 & 29\\
			&289546.3  & 0.81212 & 28 \\
			&290385.5  & 0.82069 & 27\\
			&293218.1  & 0.82359 & 24\\
			&318575.6  & 0.85887 & 22\\
			\hline
			\multirow{22}{*}{AM-NSGA-III-c}        
			&202880.4  & 0.80161 & 34 \\
			&204962.1  & 0.78154 & 34 \\
			&207677.2  & 0.79127 & 31 \\
			&209613.3 & 0.81132  & 30 \\
			&212552.1  & 0.80095& 30 \\
			&215207.2  & 0.82127 & 29 \\
			&222515.8  & 0.84418 & 28\\
			&225287.5  & 0.81077 & 28\\
			&230424.5  & 0.84525 & 27\\
			&236999.8 & 0.83394  & 27\\
			&237892.9  & 0.84922 & 26\\
			&243671.7  & 0.82374 & 26\\
			&245787.9  & 0.81008 & 26\\
			&269620.9  & 0.80388 & 25\\
			&275152.2  & 0.79682 & 25\\
			&278087.6  & 0.82612 & 24\\
			&278601.3  & 0.82032 & 24\\
			&283443.7 & 0.82831 & 22 \\
			&285710.3  & 0.81678 & 20\\
			&286750.5 & 0.81344 & 20\\
			&288313.5 & 0.85788 & 18 \\
			&289497.0 & 0.84867 & 18 \\
			% \multicolumn{2}{l}{Long Text Here} & Data 10 & Data 11 \\
			\bottomrule
		\end{tabular}
	\end{table}
	
	\begin{table}[!htbp]
	\centering
	\caption{Pareto front by comparing all feasible solutions obtained by NSGA-III, AM-NSGA-III and AM-NSGA-III-c for the high-density scenario.}
	\label{tab:pareto_all_high}
	\begin{tabular}{cccc}
		\toprule
		\textbf{Algorithms} &\textbf{Objective 1} & \textbf{Objective 2} & \textbf{Objective 3} \\
		\midrule
		\multirow{6}{*}{AM-NSGA-III}    
		&205928.2      &  0.76515  &  44\\
		&223522.9      &  0.74558  &  41 \\
		&224745.4      &  0.76874  &  39\\
		&235735.0      &  0.74113  &  39 \\
		&241672.7      &  0.78226  &  36 \\
		&267599.3      &  0.79763  &  31\\    
		\hline
		\multirow{22}{*}{AM-NSGA-III-c}    
		&202880.4      &  0.80161  &34 \\
		&204962.1      &  0.78154  &34\\
		&207677.2      &  0.79127  &31\\
		&209613.3      &  0.80132  &30\\
		&212552.1      &  0.81095  &30\\
		&215207.2      &  0.82127  &29\\
		&222515.8      &  0.84418  &28\\
		&225287.5      &  0.81077  &28\\
		&230424.5      &  0.83525  &27\\
		&236999.8      &  0.83394  &27\\
		&237892.9      &  0.80922  &26\\
		&243671.7      &  0.84374  &26\\
		&245787.9      &  0.81008  &26\\
		&269620.9      &  0.80388  &25\\
		&275152.2      &  0.79682  &25\\
		&278087.6      &  0.82612  &24\\
		&278601.3      &  0.82032  &24\\
		&283443.7      &  0.82831  &22\\
		&285710.3      &  0.81678  &20\\
		&286750.5      &  0.81344  &20\\
		&288313.5      &  0.84788  &18\\
		&289497.0      &  0.84867  &18\\
		\bottomrule
	\end{tabular}
    \end{table}
    
    \subsection{Experimental Results for the Low-density Scenario}
	\label{sec:app:experiments_low}
\begin{figure*}[!htbp]
		\centering
		\subfloat[]{\includegraphics[width=1.7in]{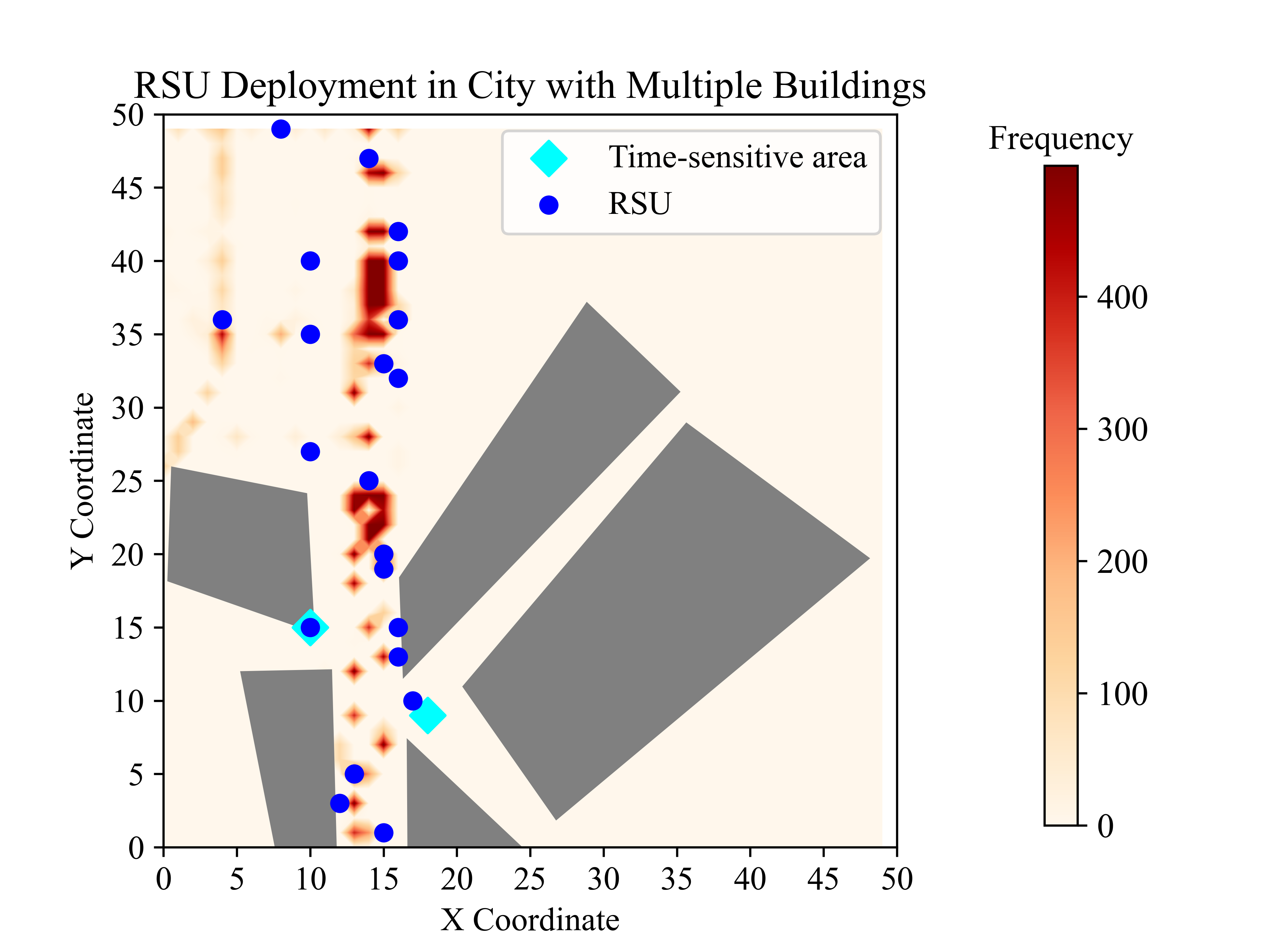}%
			\label{fig_sim_nsgaiii_2_fig_first_case}}
		\hfil
		\subfloat[]{\includegraphics[width=1.7in]{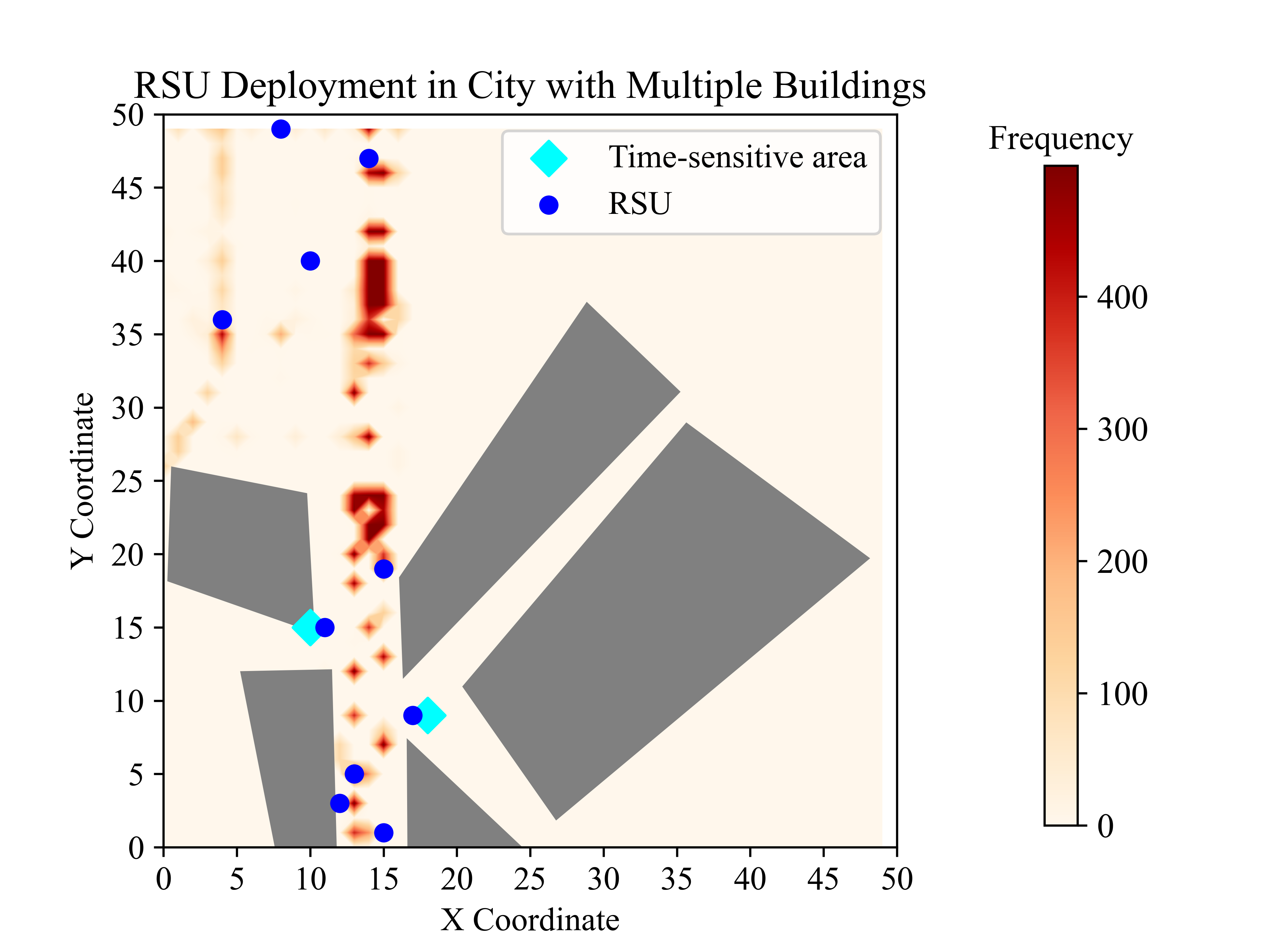}%
			\label{fig_sim_nsgaiii_2_fig_second_case}}
		\hfil
		\subfloat[]{\includegraphics[width=1.7in]{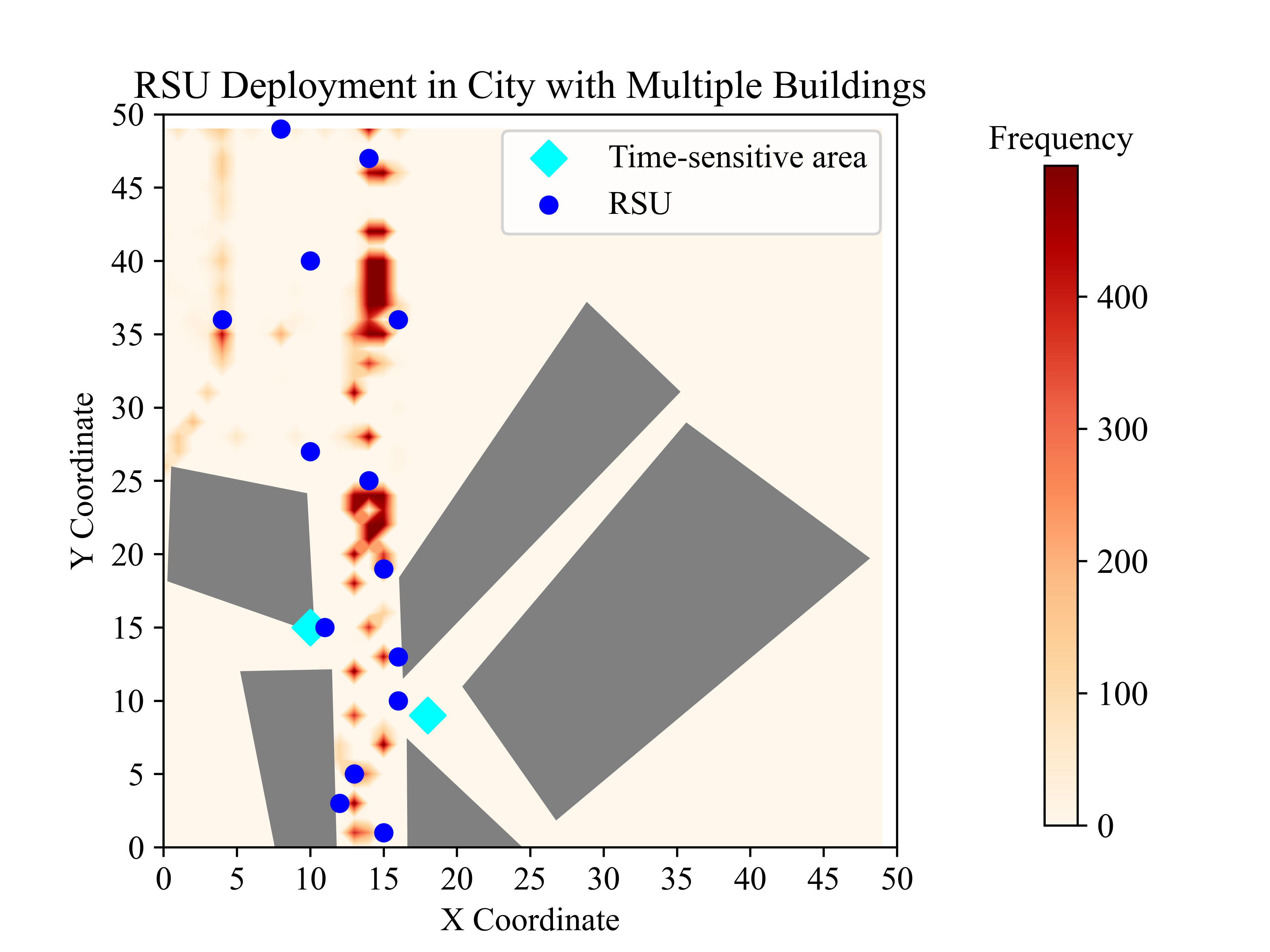}%
			\label{fig_sim_nsgaiii_2_fig_third_case}}
		\hfil
		\subfloat[]{\includegraphics[width=1.7in]{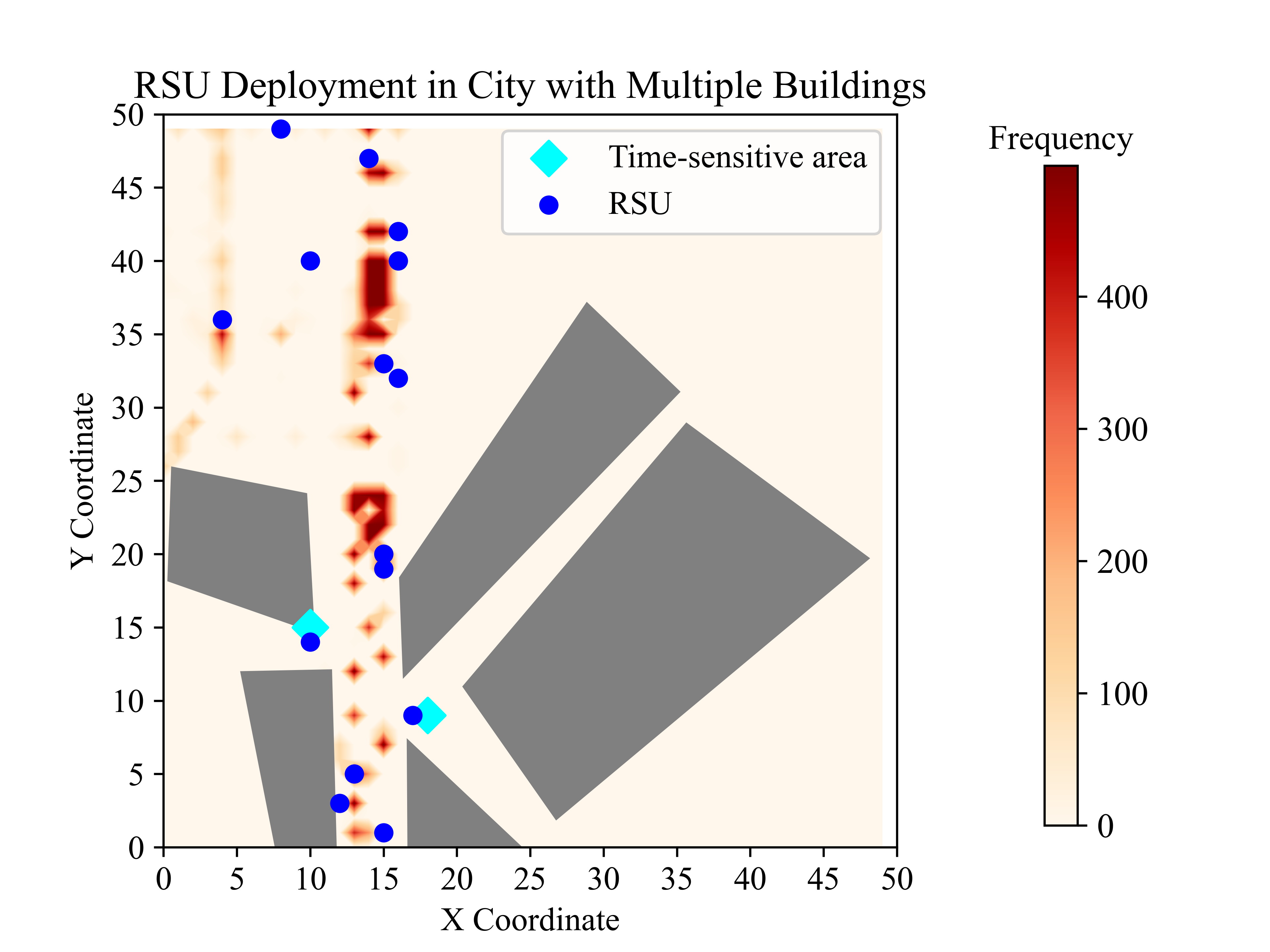}%
			\label{fig_sim_nsgaiii_2_fig_fourth_case}}
		\caption{NSGA-III deployment in low-density scenario.}
		\label{fig_sim_nsgaiii_2}
	\end{figure*}

\begin{figure*}[!htb]
		\centering
		\subfloat[]{\includegraphics[width=1.7in]{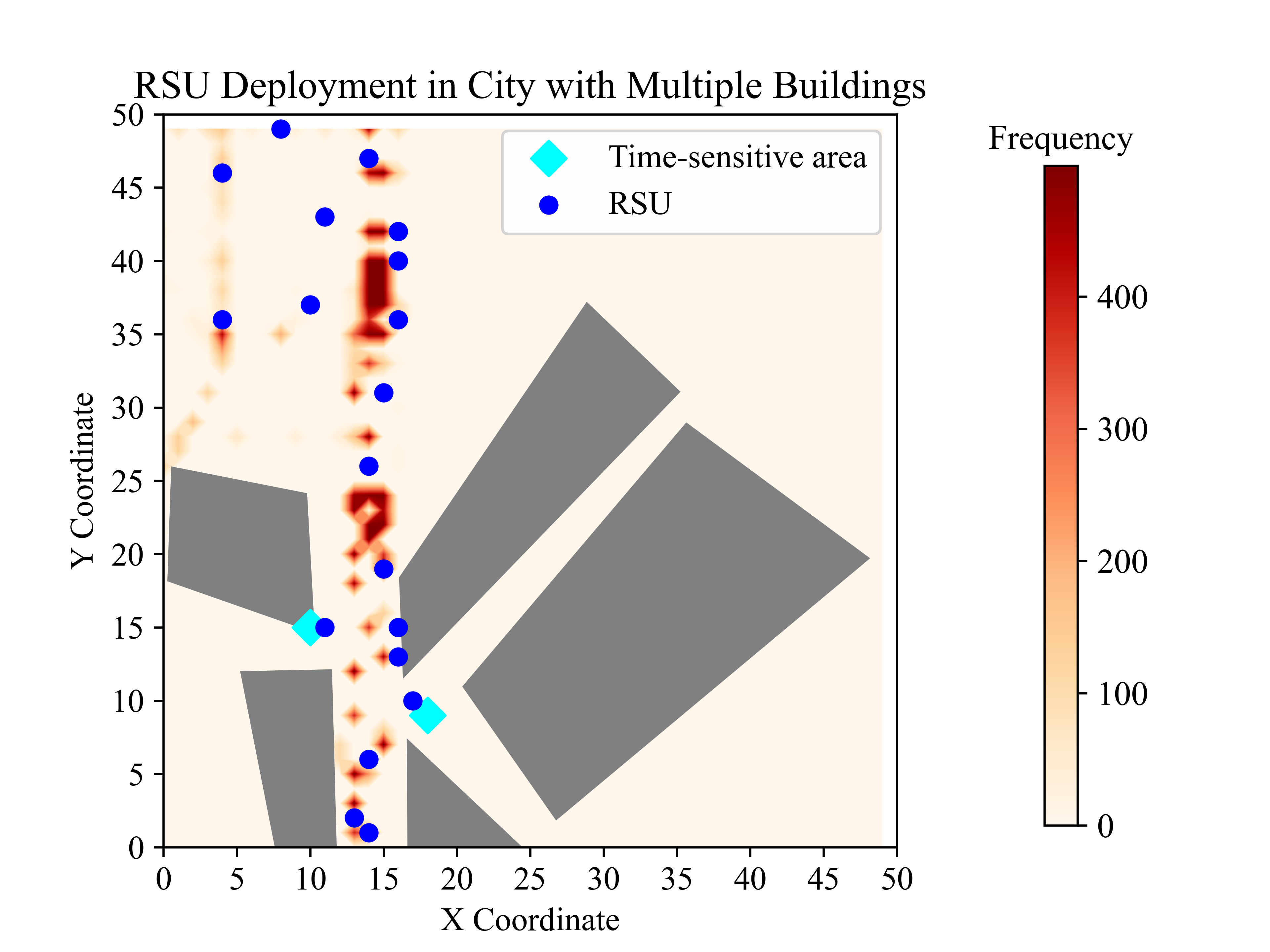}%
			\label{fig_sim_eebnsgaiii_2_fig_first_case}}
		\hfil
		\subfloat[]{\includegraphics[width=1.7in]{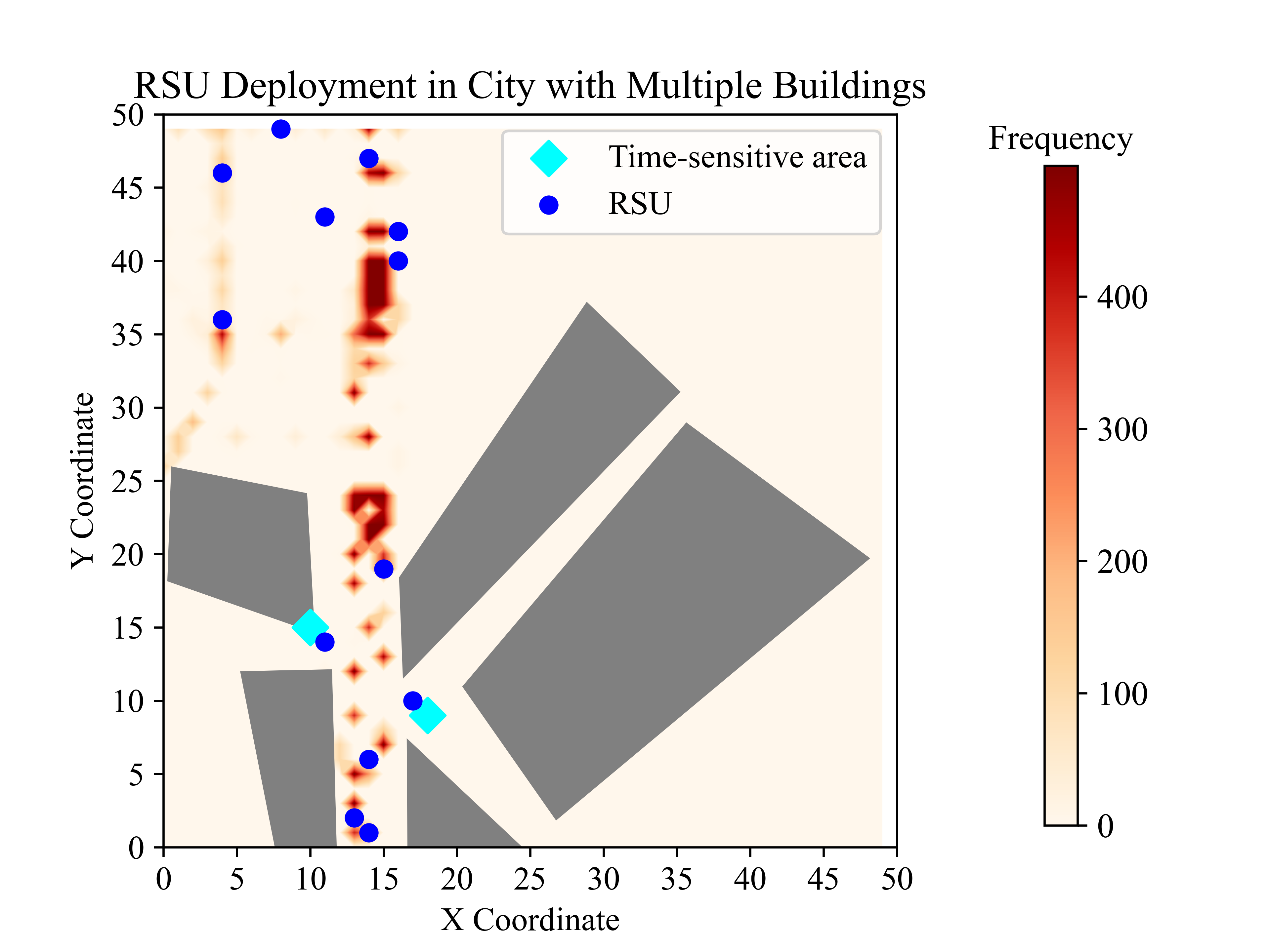}%
			\label{fig_sim_eebnsgaiii_2_fig_second_case}}
		\hfil
		\subfloat[]{\includegraphics[width=1.7in]{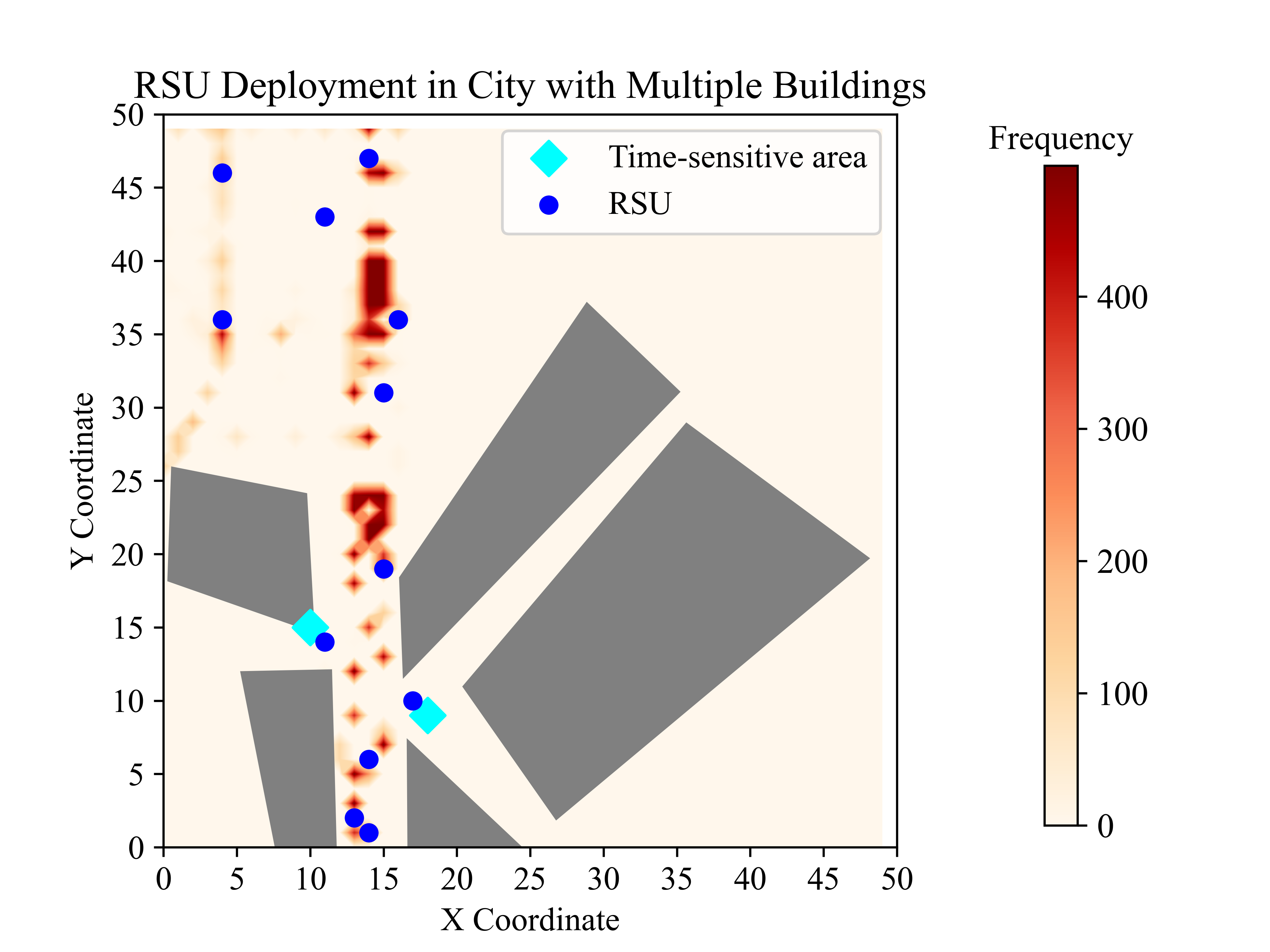}%
			\label{fig_sim_eebnsgaiii_2_fig_third_case}}
		\hfil
		\subfloat[]{\includegraphics[width=1.7in]{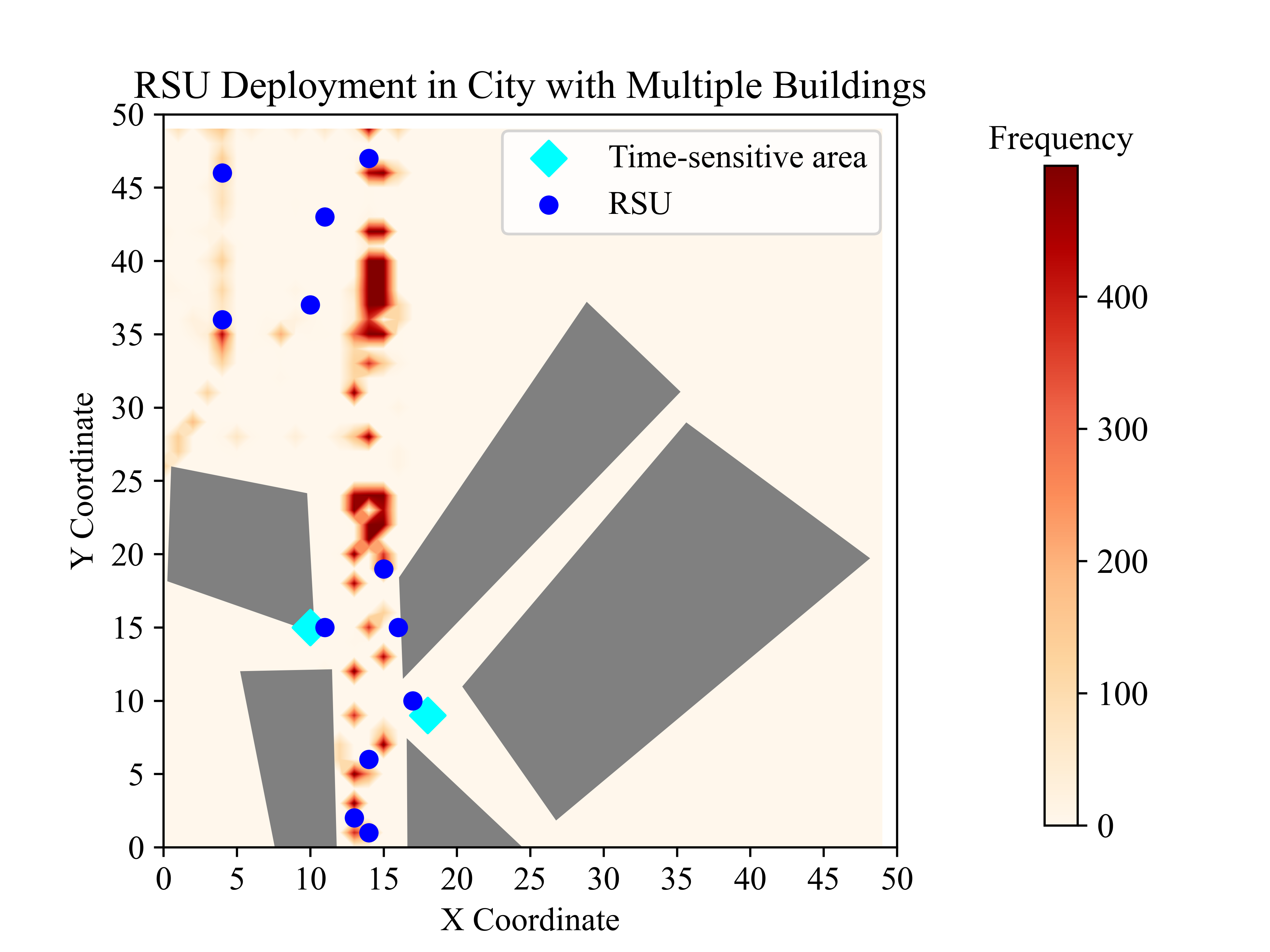}%
			\label{fig_sim_eebnsgaiii_2_fig_fourth_case}}
		\caption{AM-NSGA-III deployment in low-density scenario.}
		\label{fig_sim_eebnsgaiii_2}
	\end{figure*}

\begin{figure*}[!htb]
		\centering
		\subfloat[]{\includegraphics[width=1.7in]{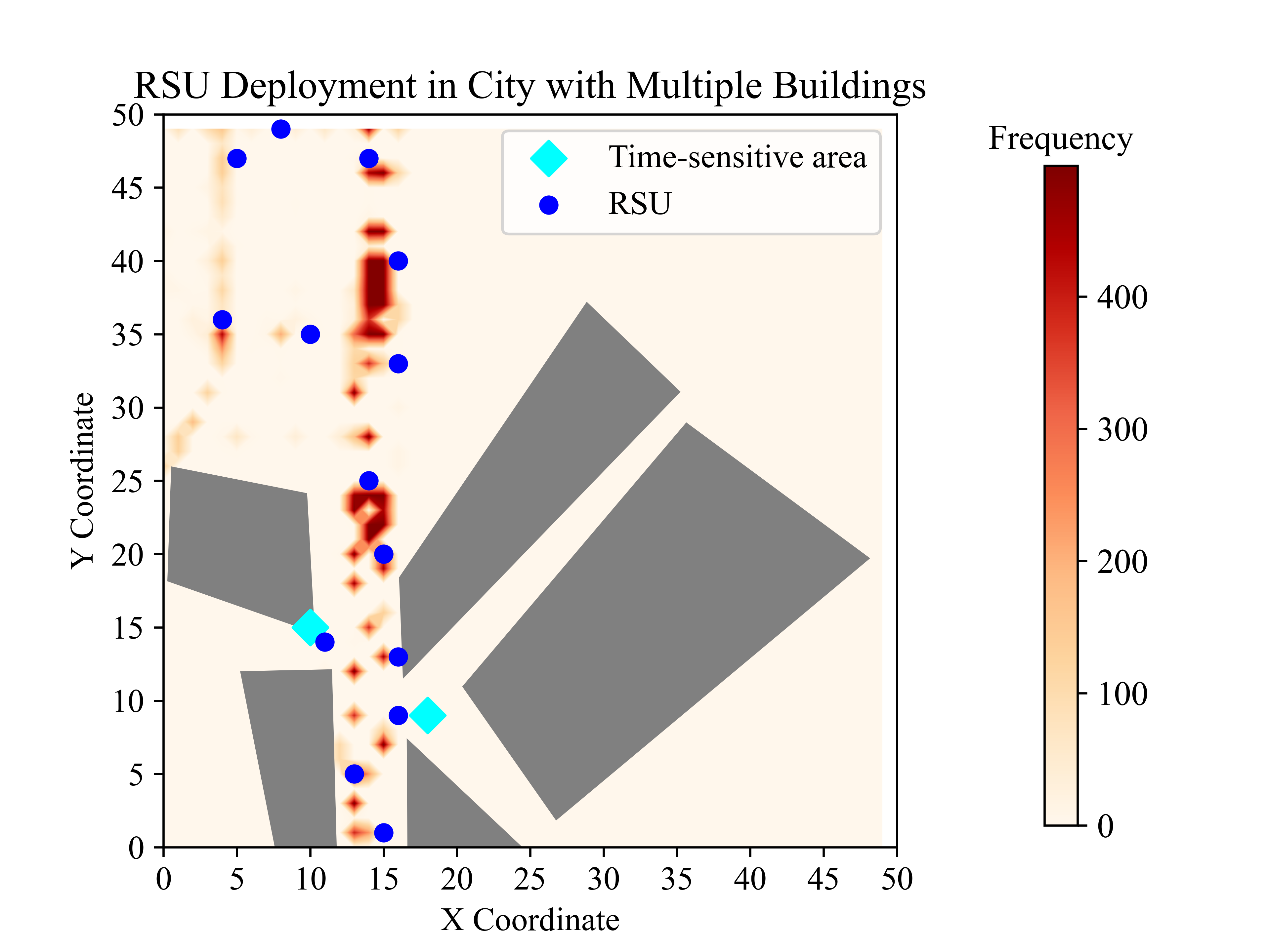}%
			\label{fig_sim_ieebnsgaiii_2_fig_first_case}}
		\hfil
		\subfloat[]{\includegraphics[width=1.7in]{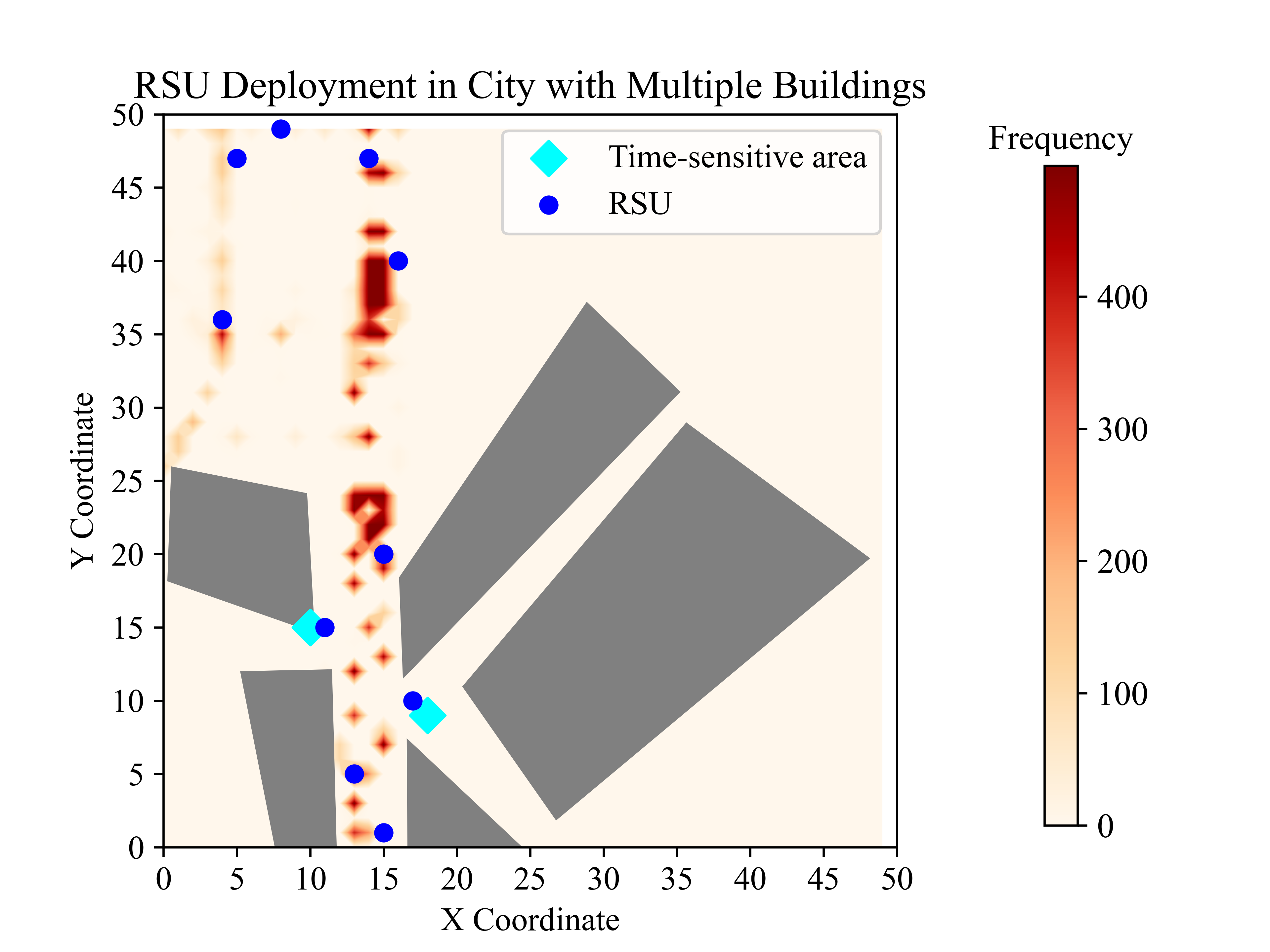}%
			\label{fig_sim_ieebnsgaiii_2_fig_second_case}}
		\hfil
		\subfloat[]{\includegraphics[width=1.7in]{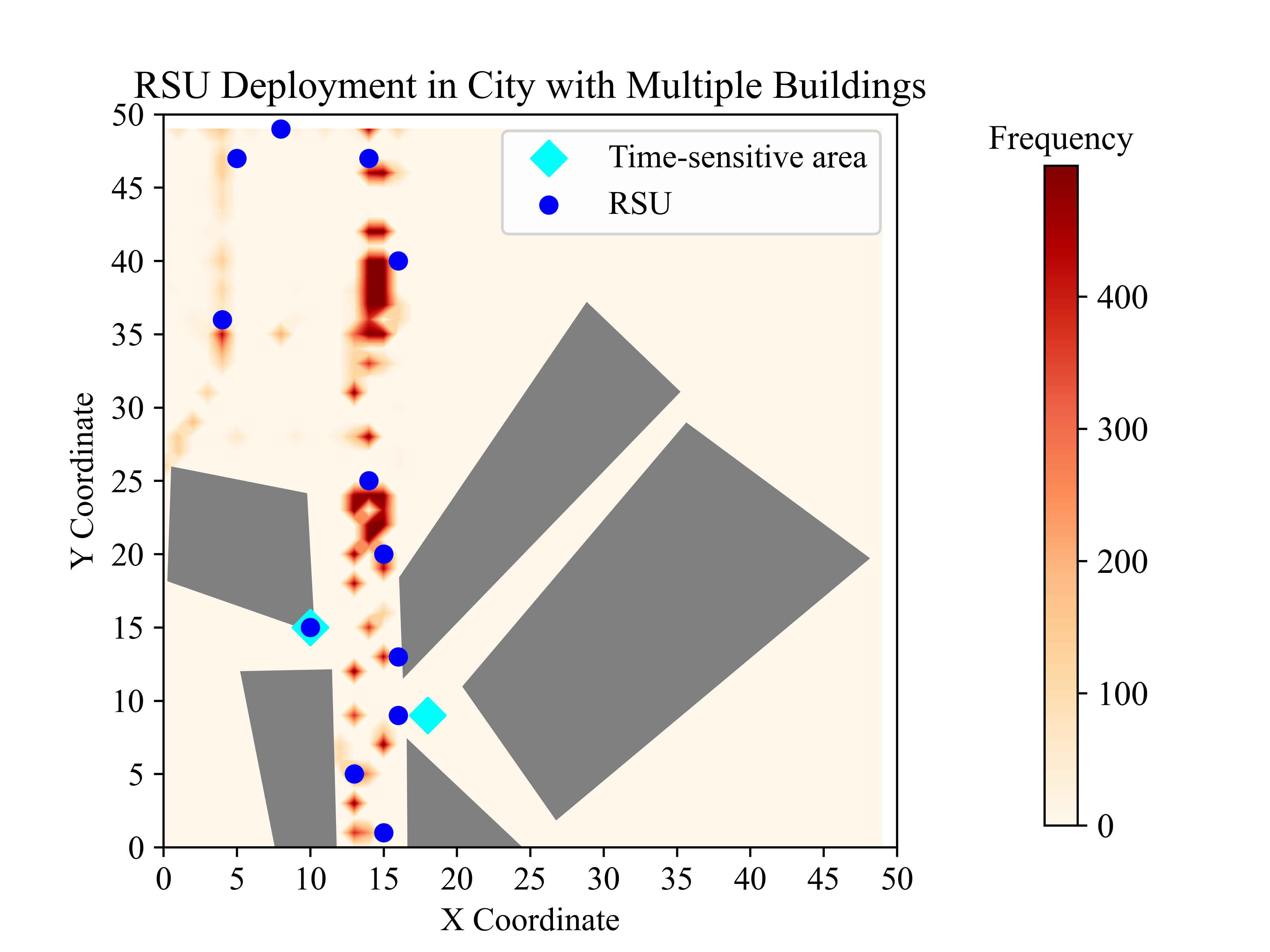}%
			\label{fig_sim_ieebnsgaiii_2_fig_third_case}}
		\hfil
		\subfloat[]{\includegraphics[width=1.7in]{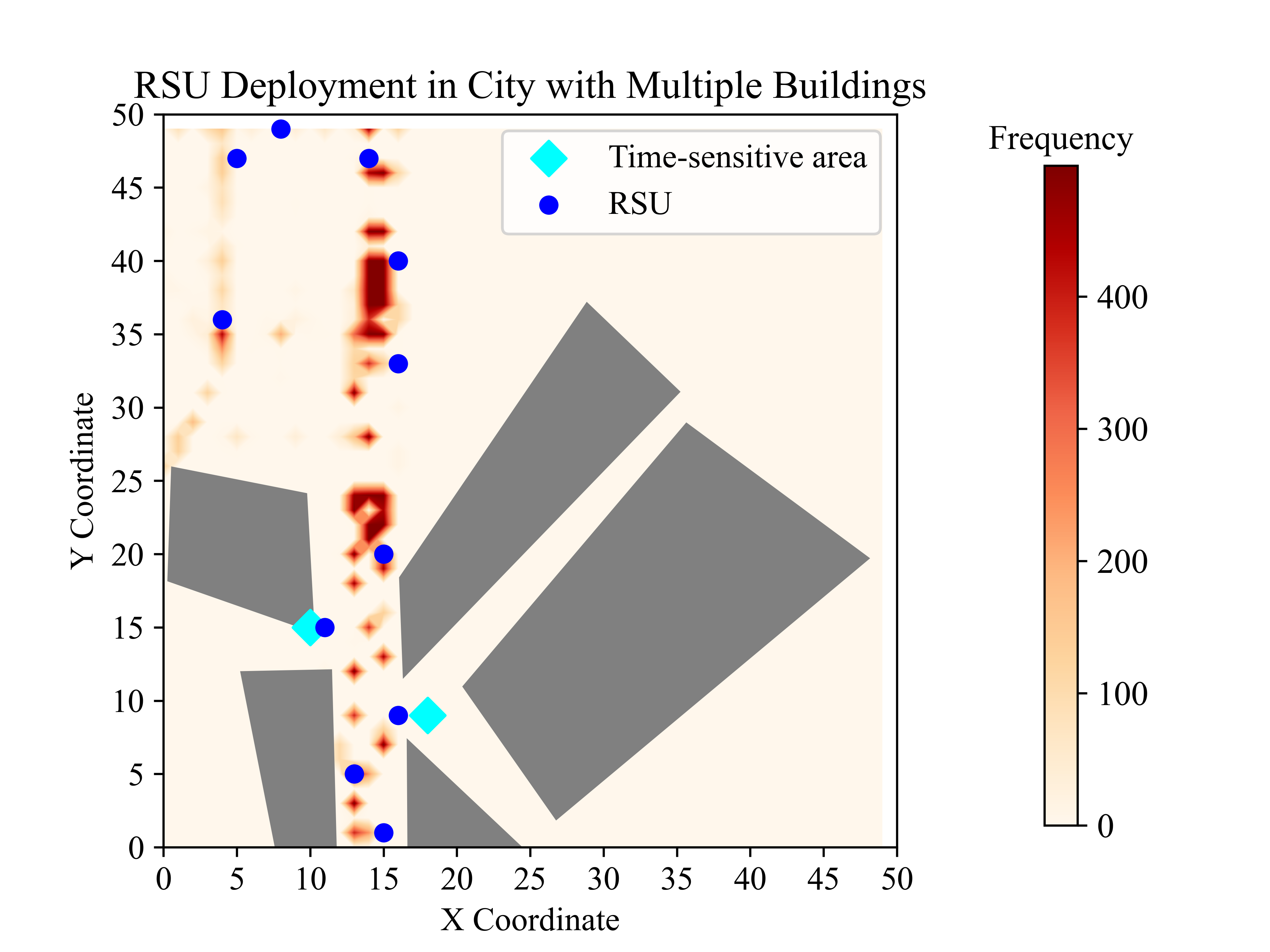}%
			\label{fig_sim_ieebnsgaiii_2_fig_fourth_case}}
		\caption{AM-NSGA-III-c deployment in low-density scenario.}
		\label{fig_sim_ieebnsgaiii_2}
	\end{figure*}

	In low-density scenario, the feasible solutions obtained by NSGA-III, AM-NSGA-III, and AM-NSGA-III-c are presented in Table \ref{tab:performance_low_density}. Based on the data in such table, we compare all the feasible solutions to compose a pareto front shown in Table \ref{tab:new_pareto2}. For each algorithm, we present four diagrams to show the RSU deployment in Fig. \ref{fig_sim_nsgaiii_2}, Fig. \ref{fig_sim_eebnsgaiii_2} and Fig. \ref{fig_sim_ieebnsgaiii_2}.

	\begin{table}[!htbp]
		\centering
		\caption{The feasible solutions obtained by NSGA-III, AM-NSGA-III and AM-NSGA-III-c for the low-density scenario.}
		\label{tab:performance_low_density}
		\begin{tabular}{cccc}
			\toprule
			\textbf{Algorithms} &\textbf{Objective 1} & \textbf{Objective 2} & \textbf{Objective 3} \\
			\midrule
			\multirow{6}{*}{MOEA/D}    
			&12854.36                   & 0.829952         & 17  \\
			&13772.94                    & 0.80525          & 15 \\
			&13869.27                    & 0.79663          & 13 \\
			&14804.98                   & 0.81433           & 11 \\
			&14693.35                  & 0.854056           & 11 \\
			&15560.63                    & 0.815891        & 10  \\
			\hline
			\multirow{16}{*}{NSGA-III}    
			&10616.77    &0.749952    &33    \\
			&10772.94    &0.73525    &29    \\
			&10911.69    &0.768727    &25    \\
			&11369.27    &0.77663    &24    \\
			&11474.81    &0.702563    &24    \\
			&11604.98    &0.74433    &23    \\
			&11704.99    &0.701207    &22    \\
			&11893.35    &0.754056    &21    \\
			&12560.63    &0.795891    &19    \\
			&12665.17    &0.717512    &18    \\
			&12788.85    &0.730365    &18    \\
			&12965.76    &0.726708    &17    \\
			&13128.92    &0.72856    &17    \\
			&13881.26    &0.71809    &14    \\
			&13990.79    &0.796187    &14    \\
			&14136.02    &0.720669    &13    \\
			\hline
			\multirow{18}{*}{AM-NSGA-III}    
			&10243.95    &0.781843    &29    \\
			&10661.78    &0.724627    &28    \\
			&10880.98    &0.776299    &27    \\
			&10985.55    &0.786457    &25    \\
			&11129.62    &0.758313    &24    \\
			&11246.17    &0.712753    &22    \\
			&11367.18    &0.782226    &21    \\
			&11899.37    &0.74847    &19    \\
			&12133.94    &0.719733    &19    \\
			&116428.2    &0.712334    &18    \\
			&12735.53    &0.768231    &18    \\
			&12941.74    &0.798218    &18    \\
			&13054.87    &0.748149    &16    \\
			&13222.71    &0.764303    &15    \\
			&13340.65    &0.736728    &14    \\
			&13551.23    &0.730376    &13    \\
			&13721.36    &0.725468    &12    \\
			&14580.23    &0.785957    &10    \\
			\hline
			\multirow{17}{*}{AM-NSGA-III-c}    
			&11042.11    &0.749952    &19    \\
			&11300.03    &0.73525    &18    \\
			&11434.38    &0.768727    &18    \\
			&11575.73    &0.77663    &16    \\
			&11706.61    &0.702563    &14    \\
			&11830.42    &0.701207    &14    \\
			&11944.71    &0.754056    &14    \\
			&12077.8    &0.717512    &14    \\
			&12178.16    &0.730365    &12    \\
			&12310.18    &0.726708    &12    \\
			&12441.02    &0.72856    &11    \\
			&12645.51    &0.71809    &11    \\
			&12765.64    &0.796187    &11    \\
			&12959.33    &0.78192    &11    \\
			&13120.02    &0.782896    &10    \\
			&13306.02    &0.784019    &10    \\
			&13475.85    &0.759641    &9    \\
			\bottomrule
		\end{tabular}
	\end{table}
	
	\begin{table}[!htbp]
		\centering
		\caption{Pareto front by comparing all feasible solutions obtained by NSGA-III, AM-NSGA-III and AM-NSGA-III-c for the low-density scenario.}
		\label{tab:new_pareto2}
		\begin{tabular}{cccc}
			\toprule
			\textbf{Algorithms} &\textbf{Objective 1} & \textbf{Objective 2} & \textbf{Objective 3} \\
			\midrule
			\multirow{3}{*}{NSGA-III}
			     &10911.69    &0.768727    &25    \\
			  &11474.81    &0.702563    &24    \\
			    &11704.99    &0.701207    &22    \\
			\hline
			\multirow{8}{*}{AM-NSGA-III}
			  &10243.95    &0.781843    &29    \\
			   &10661.78    &0.724627    &28    \\
			  &10880.98    &0.776299    &27    \\
			   &10985.55    &0.786457    &25    \\
			   &11129.62    &0.758313    &24    \\
			   &11246.17    &0.712753    &22    \\
			    &11367.18    &0.782226    &21    \\
			   &11899.37    &0.74847    &19    \\
			\hline
			\multirow{10}{*}{AM-NSGA-III-c}
			   &11042.11    &0.749952    &19    \\
			   &11300.03    &0.73525    &18    \\
			   &11575.73    &0.77663    &16    \\
			   &11830.42    &0.701207    &14    \\
			   &12178.16    &0.730365    &12    \\
			    &12310.18    &0.726708    &12    \\
		   &12441.02    &0.72856    &11    \\
			 &12645.51    &0.71809    &11    \\
			    &13120.02    &0.782896    &10    \\
			   &13475.85    &0.759641    &9    \\
			\bottomrule
		\end{tabular}
	\end{table}	
		
	\subsection{Experimental Results for Increased Number of Latency-sensitive Areas} 
	\label{sec:app:increase_latency}	
\begin{figure*}[!htbp]
		\centering
		\subfloat[]{\includegraphics[width=1.7in]{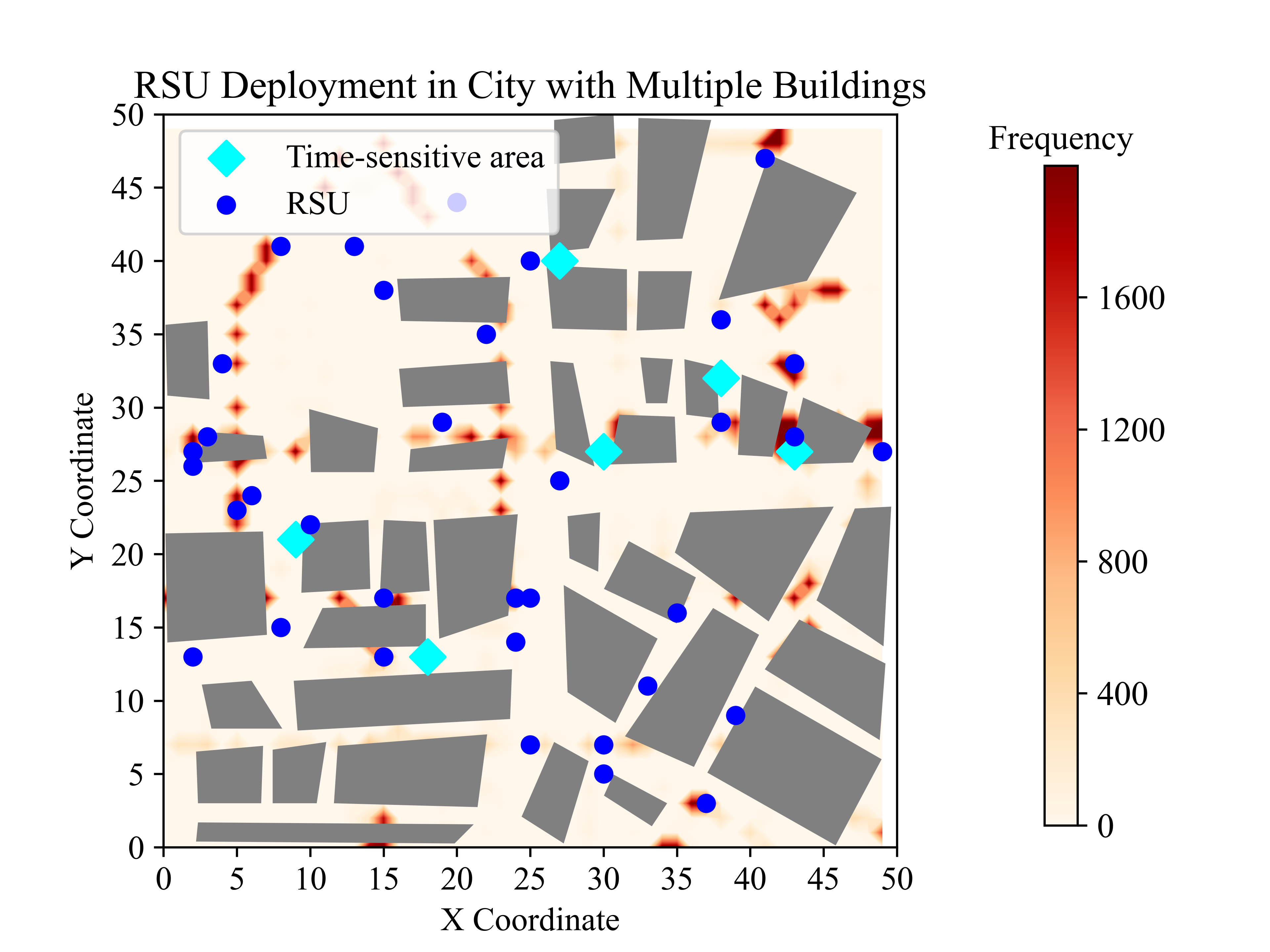}%
			\label{fig_nsga_6_first_case}}
		\hfil
		\subfloat[]{\includegraphics[width=1.7in]{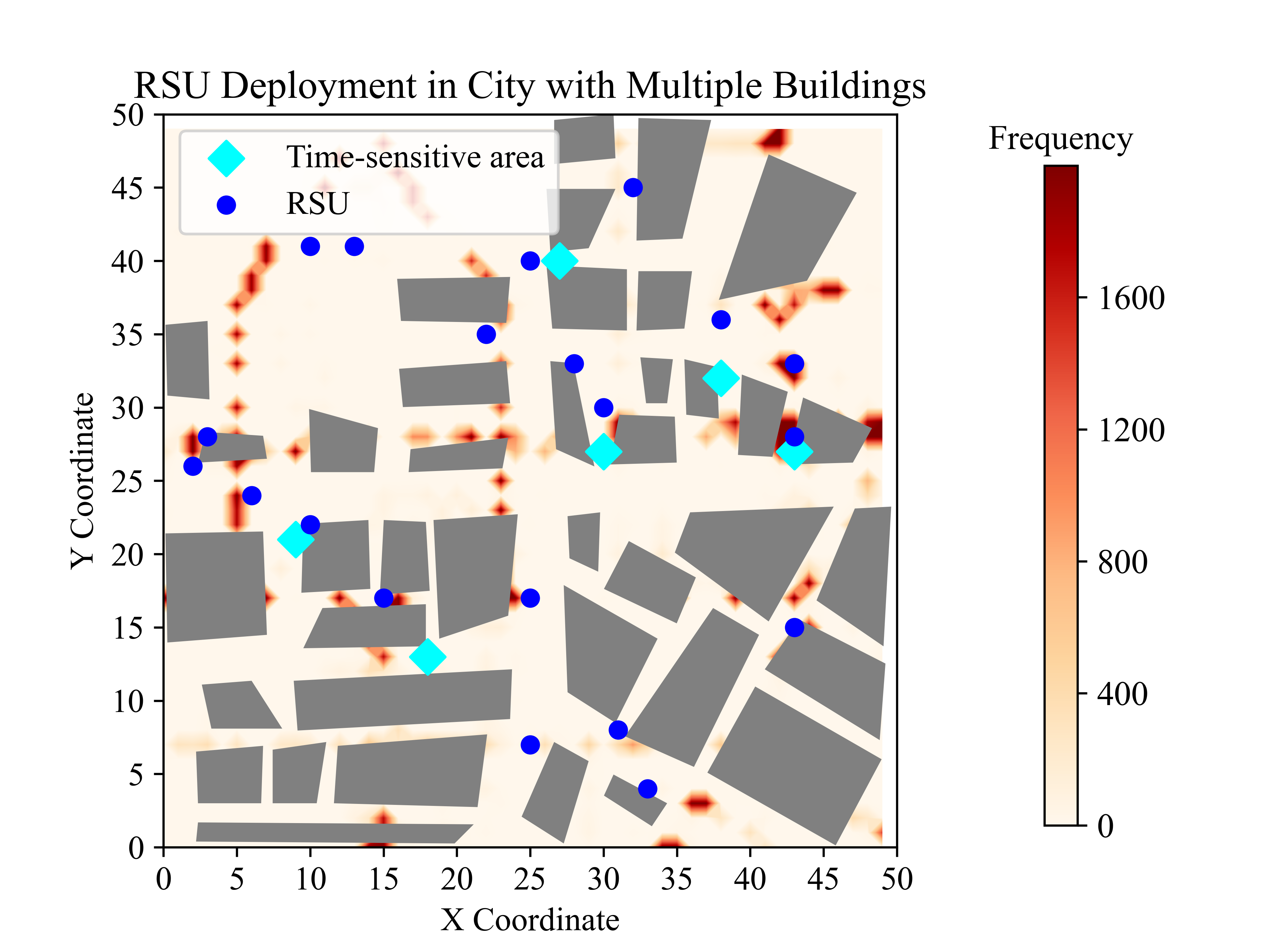}%
			\label{fig_nsga_6_second_case}}
		\hfil
		\subfloat[]{\includegraphics[width=1.7in]{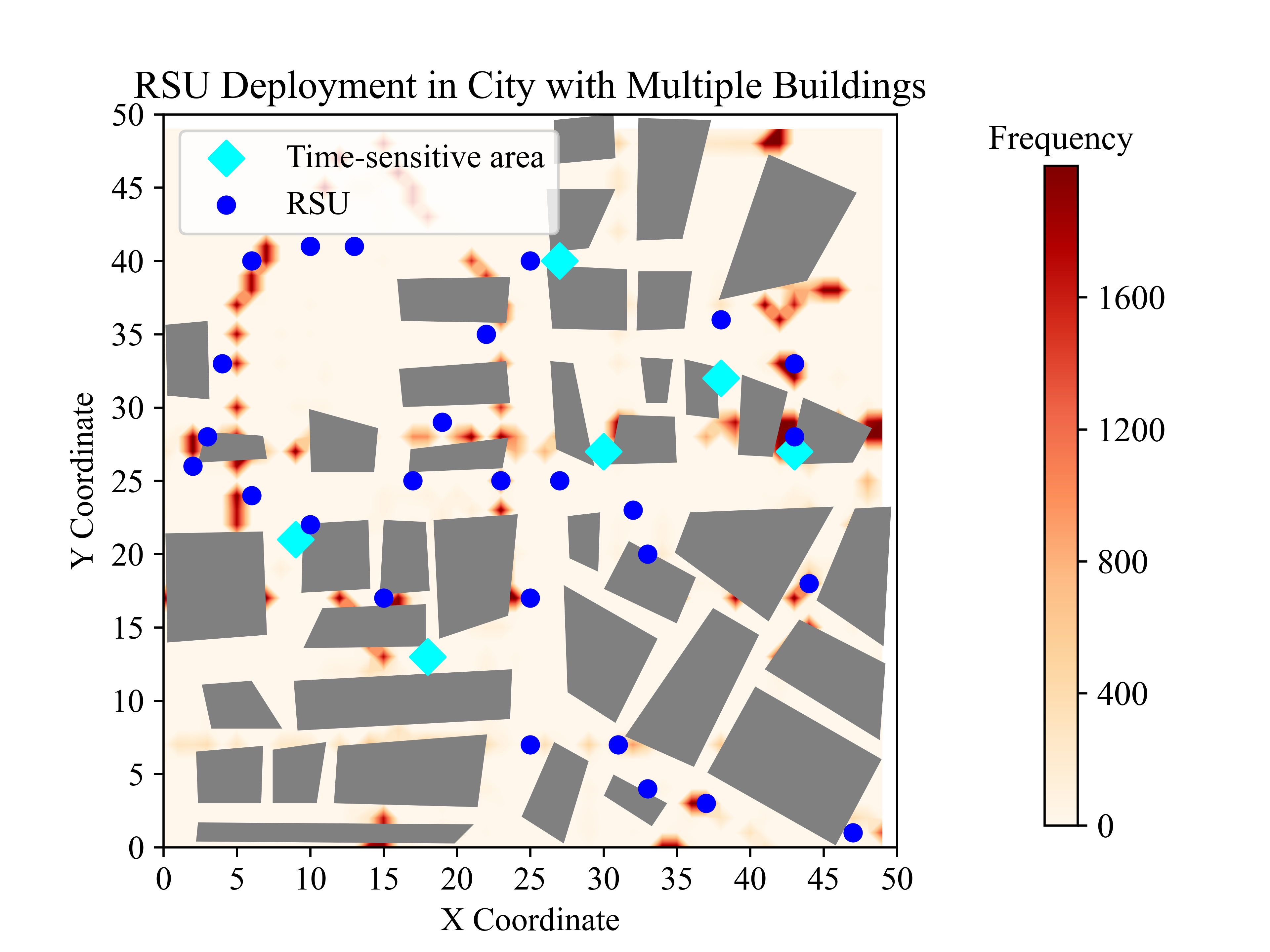}%
			\label{fig_nsga_6_third_case}}
		\hfil
		\subfloat[]{\includegraphics[width=1.7in]{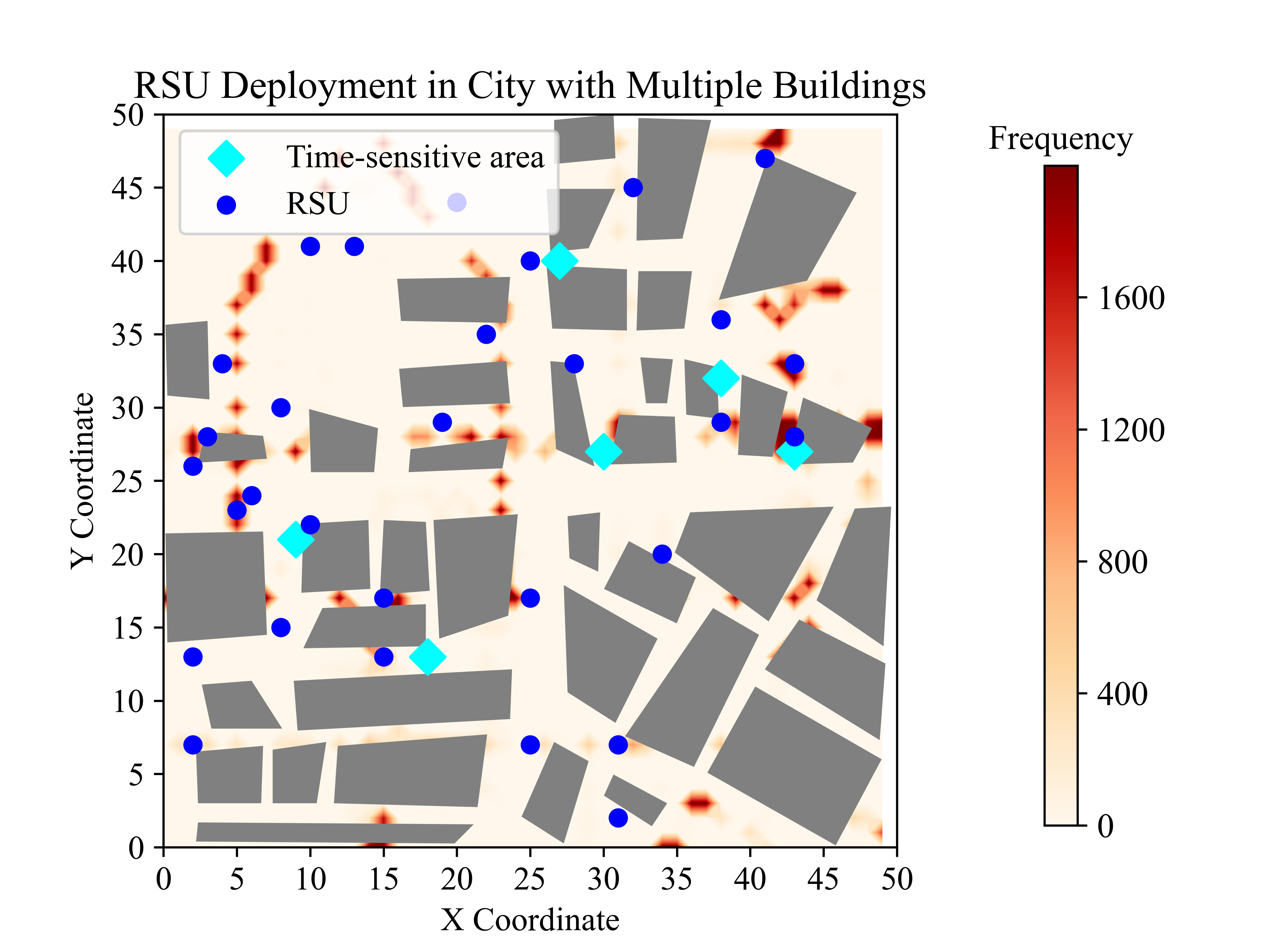}%
			\label{fig_nsga_6_fourth_case}}
		\hfil
		\caption{AM-NSGA-III deployment of with 6 latency-sensitive areas.}
		\label{fig_nsga_6}
	\end{figure*}

\begin{figure*}[!htbp]
		\centering
		\subfloat[]{\includegraphics[width=1.7in]{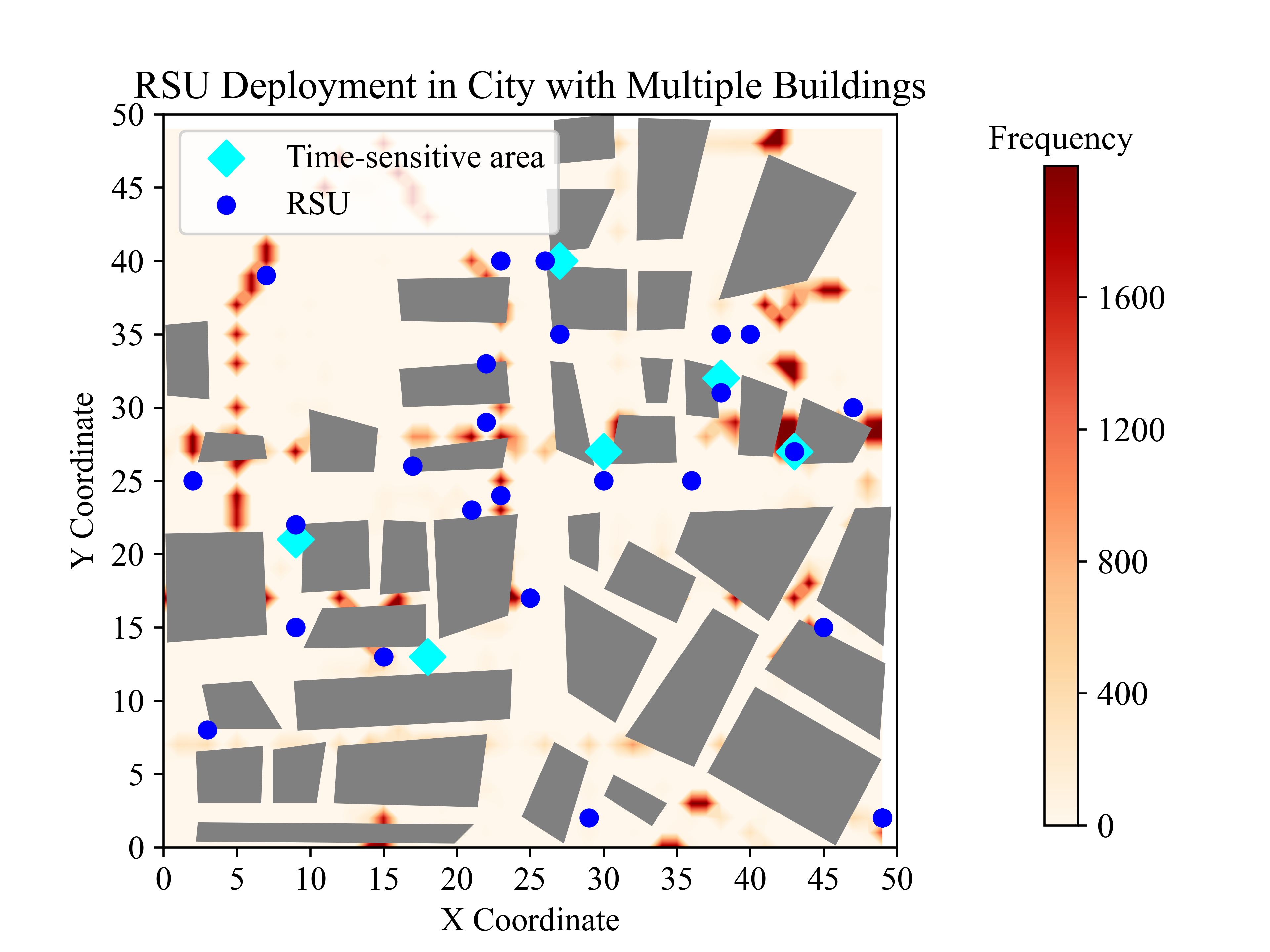}%
			\label{fig_am_nsga_6_first_case}}
		\hfil
		\subfloat[]{\includegraphics[width=1.7in]{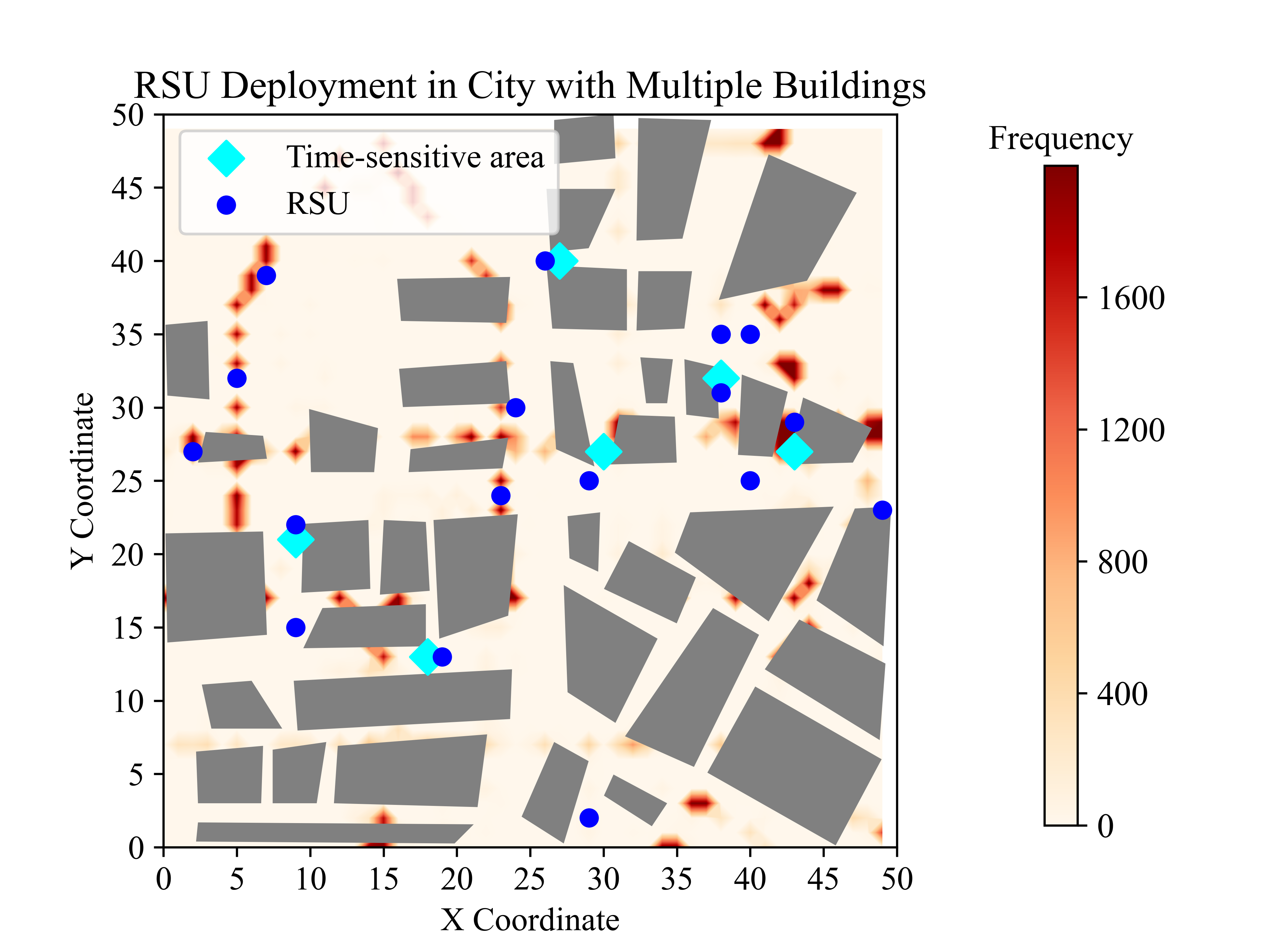}%
			\label{fig_am_nsga_6_second_case}}
		\hfil
		\subfloat[]{\includegraphics[width=1.7in]{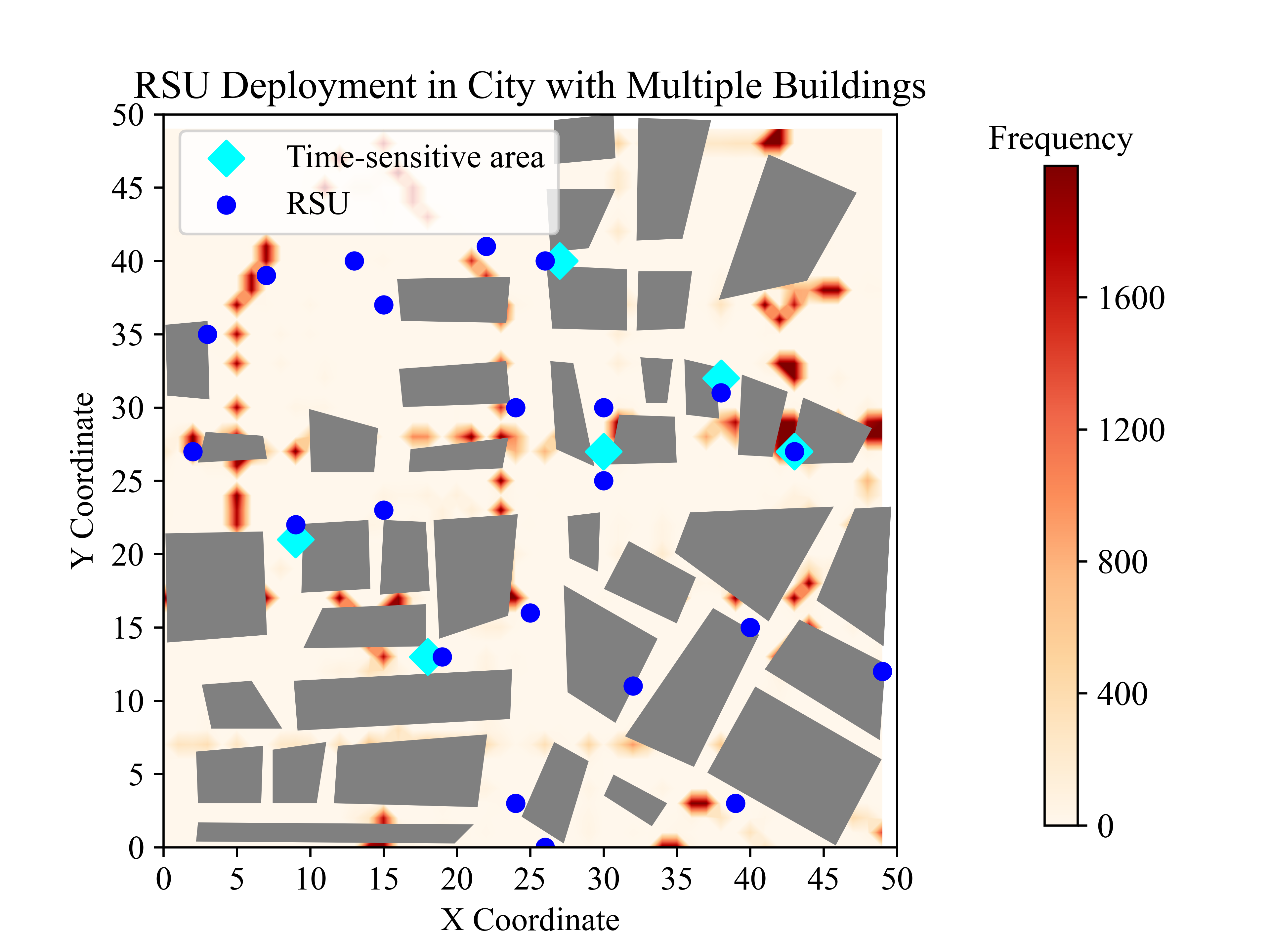}%
			\label{fig_am_nsga_6_third_case}}
		\hfil
		\subfloat[]{\includegraphics[width=1.7in]{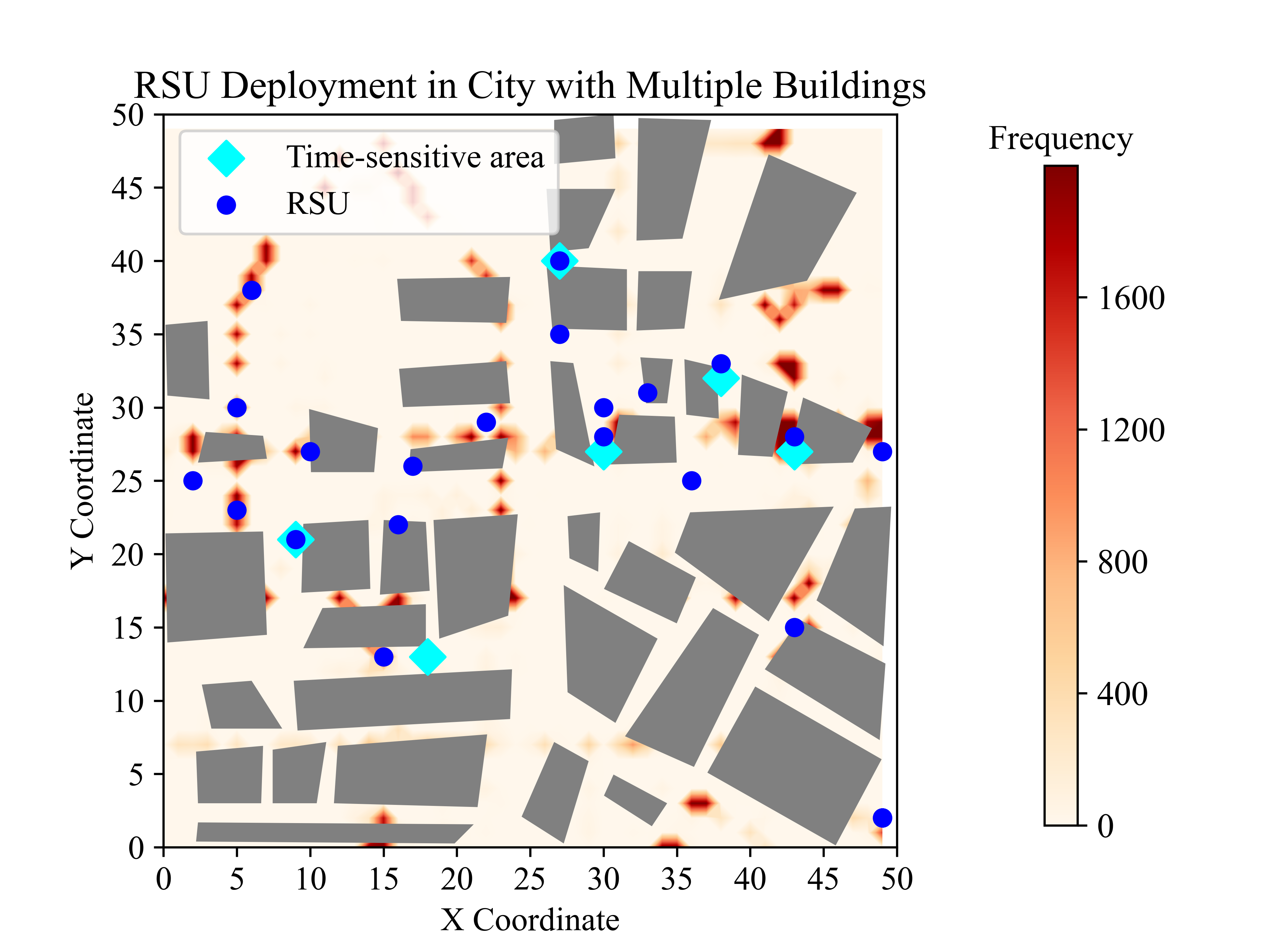}%
			\label{fig_am_nsga_6_fourth_case}}
		\hfil
		\caption{AM-NSGA-III-c deployment of with 6 latency-sensitive areas.}
		\label{fig_am_nsga_6}
	\end{figure*}

\begin{figure*}[!htbp]
		\centering
		\subfloat[]{\includegraphics[width=1.3in]{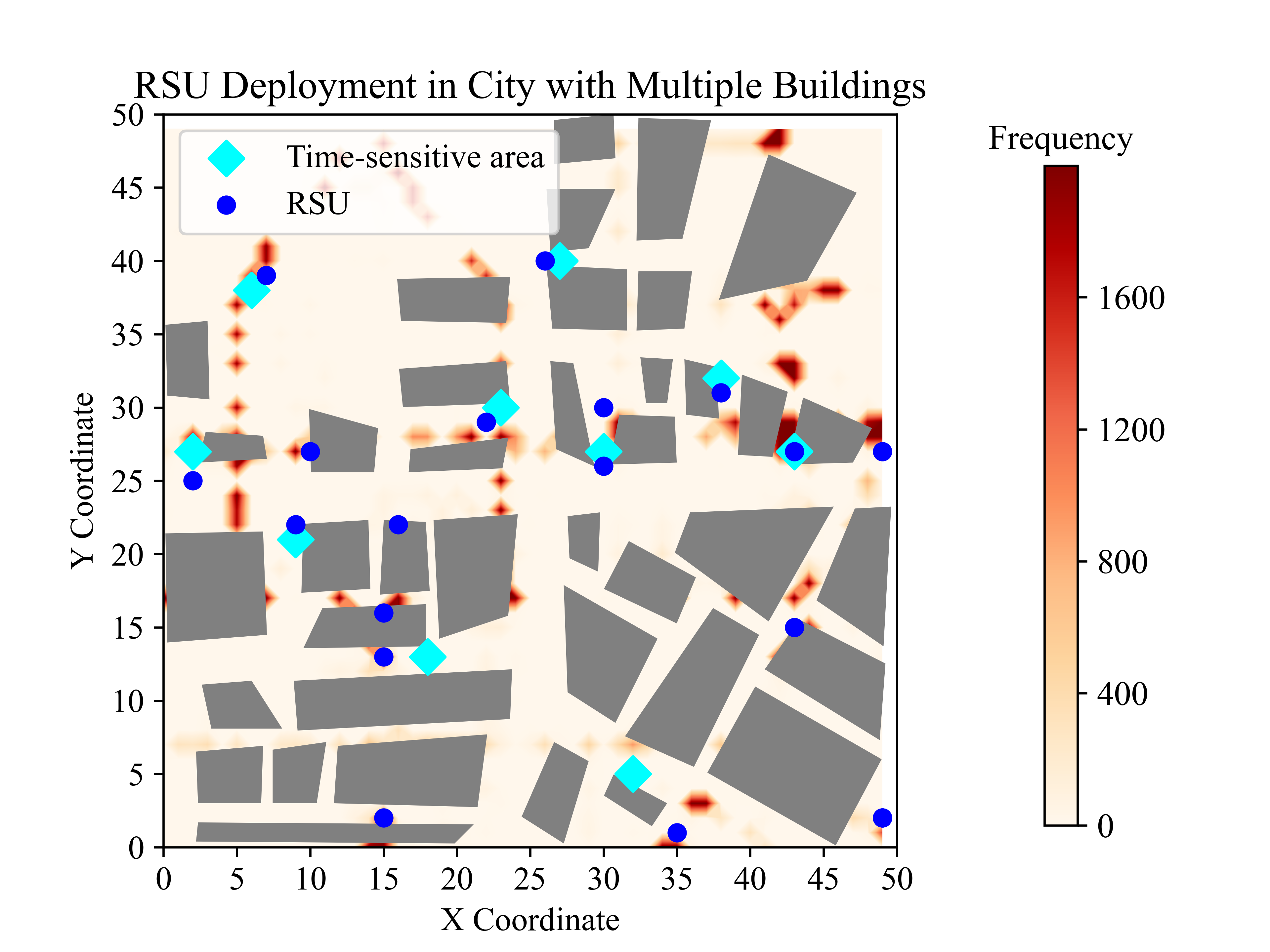}%
			\label{fig_am_nsga_10_first_case}}
		\hfil
		\subfloat[]{\includegraphics[width=1.3in]{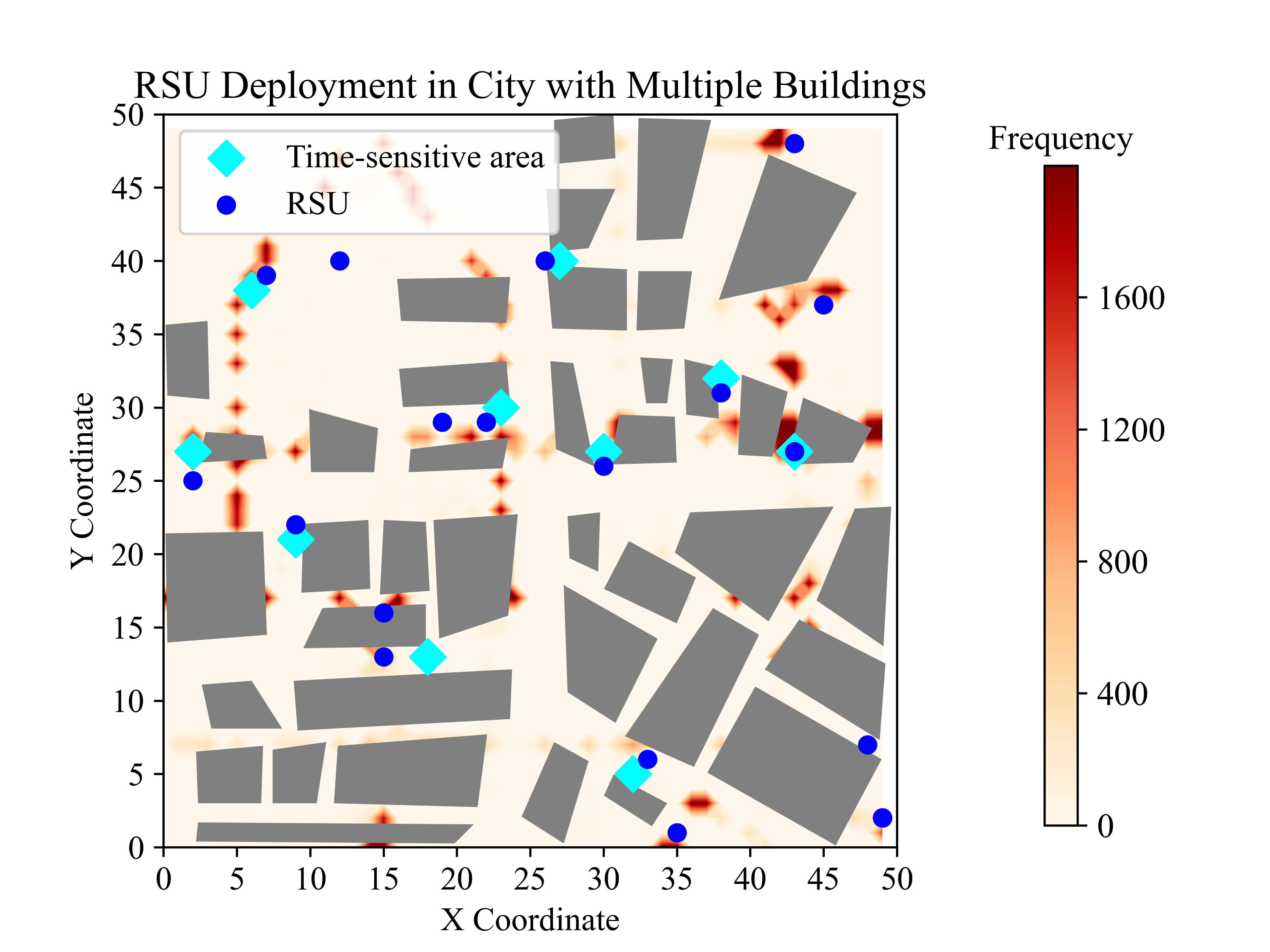}%
			\label{fig_am_nsga_10_second_case}}
		\hfil
		\subfloat[]{\includegraphics[width=1.3in]{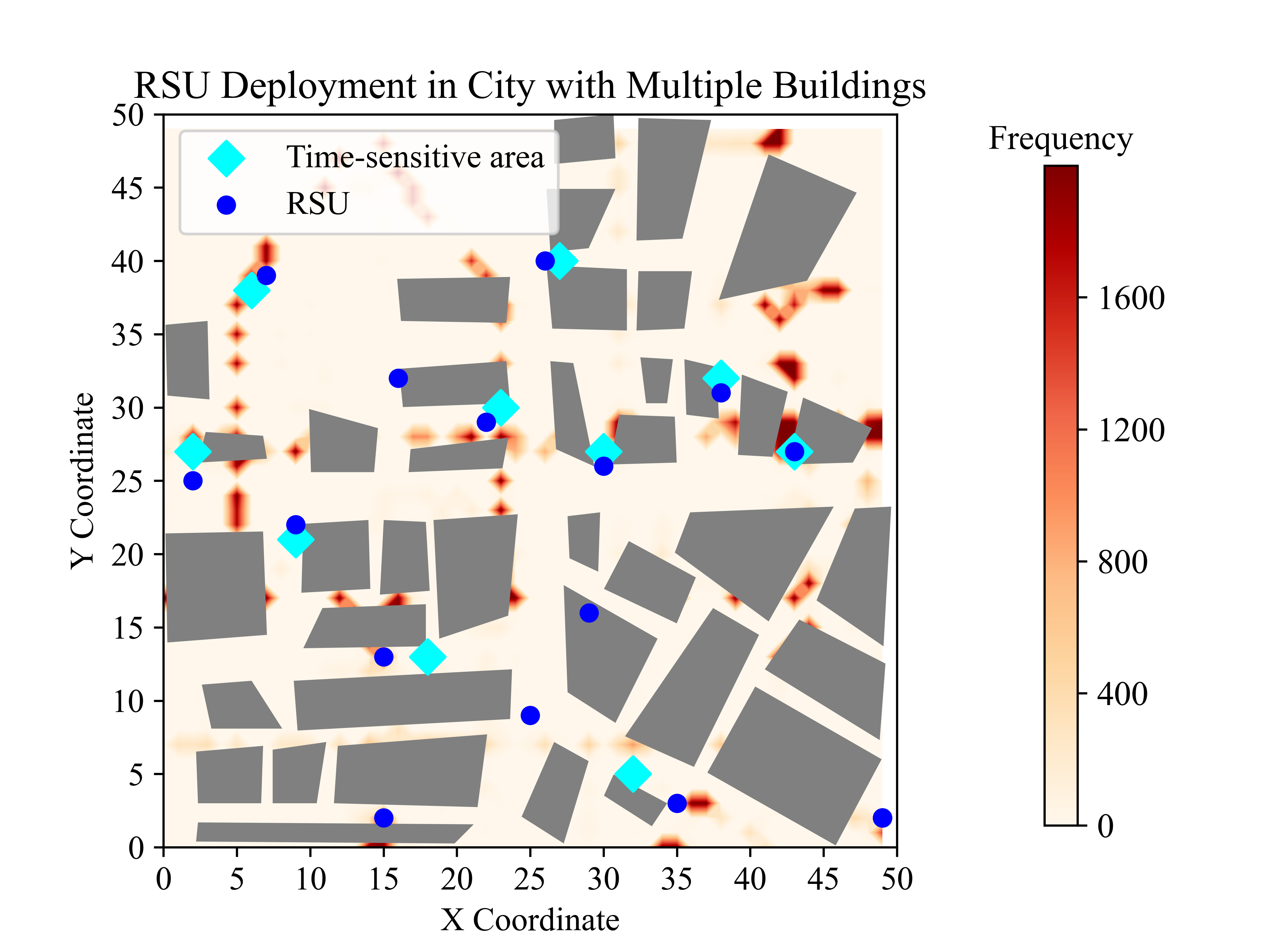}%
			\label{fig_am_nsga_10_third_case}}
		\hfil
		\subfloat[]{\includegraphics[width=1.3in]{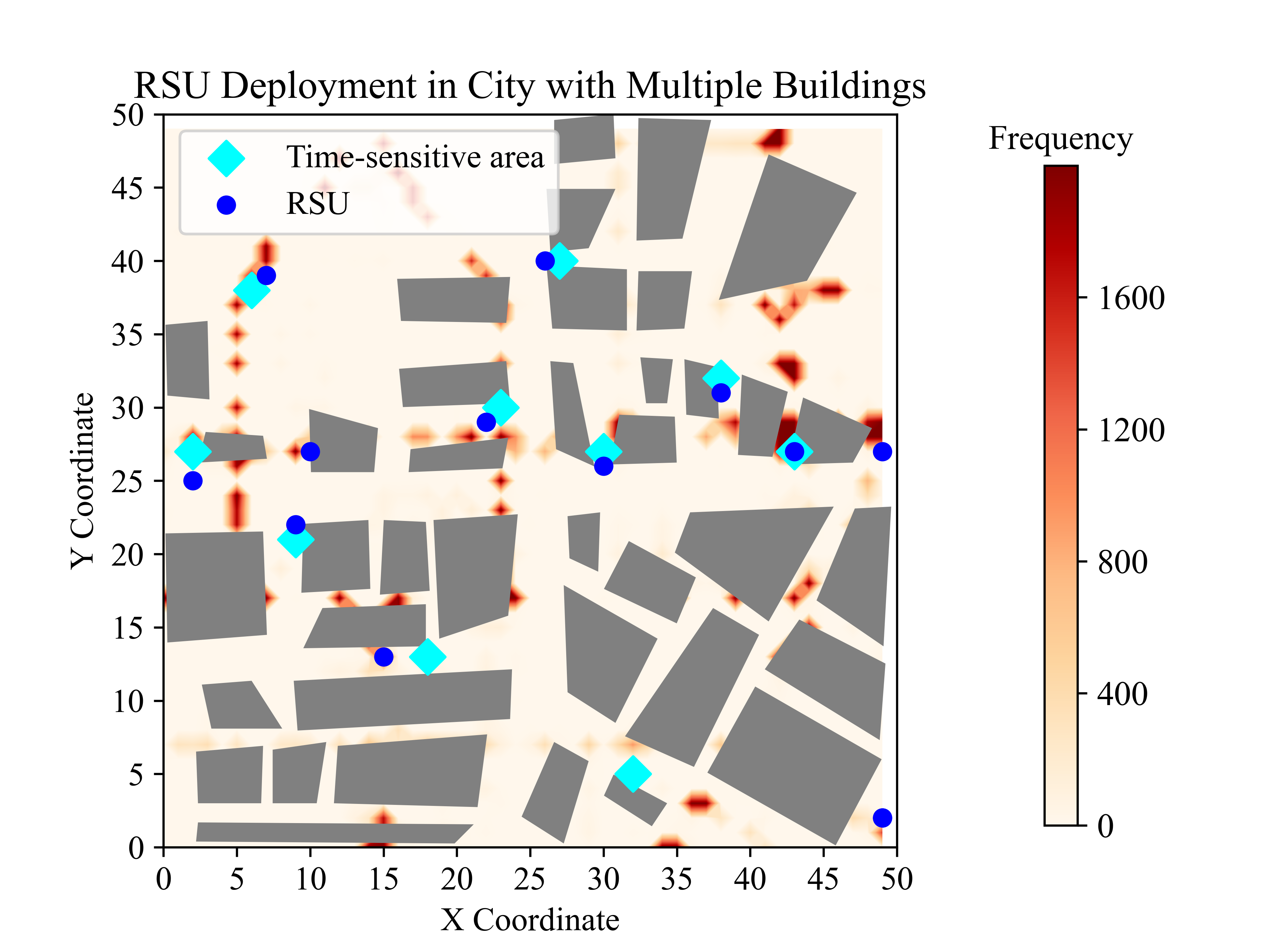}%
			\label{fig_am_nsga_10_fourth_case}}
		\subfloat[]{\includegraphics[width=1.3in]{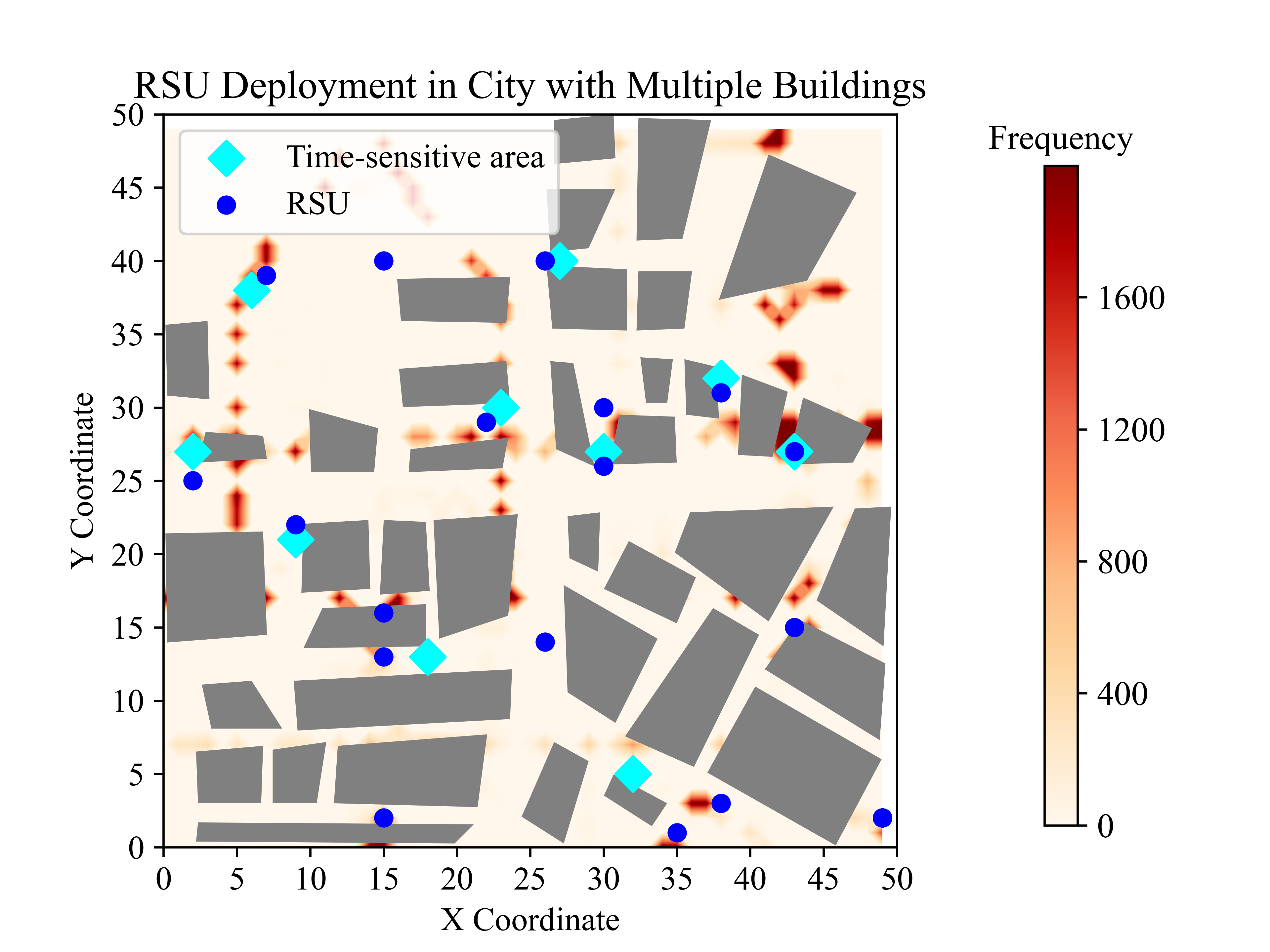}%
			\label{fig_am_nsga_10_fifth_case}}
		\caption{AM-NSGA-III-c deployment of with 10 latency-sensitive areas.}
		\label{fig_am_nsga_10}
	\end{figure*}

	The number of latency-sensitive areas may greatly influence the practical implementation of RSU deployment. To investigate the influences, it will be helpful to explore algorithms' capabilities in dealing with complex urban environments. In this experiments, the principle to select the areas is based on the intersections of roads. Two situations are conducted in this subsection, which are with 6 and 10 latency-sensitive areas. For the 6 latency-sensitive areas, results are given in Fig. \ref{fig_nsga_6} and Fig. \ref{fig_am_nsga_6} for AM-NSGA-III and AM-NSGA-III-c, respectively. For the 10 latency-sensitive areas, results are presented in Fig. \ref{fig_am_nsga_10}. 

	\subsection{Comparison results for Different Data Offloading Strategies}
	\label{sec:app:Data_offloading}	
	To evaluate the performance of different offloading strategies,  the metrics are presented as follows. 	
	\begin{enumerate}
		\item Total delay: The sum of the delays experienced by all vehicles across all time periods based on \eqref{eqn:total_delay}.
		\item The load balance: The load balance ensures that the vehicles' offloading requests are distributed evenly among the available RSUs. This helps prevent overloading a certain number of RSUs and promotes RSU utilization, which is defined in \eqref{eqn:balancing}.
		\begin{equation}
			\label{eqn:balancing}
			\text{Balancing} = \sqrt{\frac{\sum_{i=1}^{R} (\text{Load}_i - \text{Load}_\text{mean})^2}{R}}
		\end{equation}		
		where $\text{Load}_\text{mean}$ is the average load across all RSUs and can be calculated by $\text{Load}_\text{mean} = \frac{\sum_{i=1}^{M} \text{Load}_i}{M}$, $\text{Load}_i$ represents the load (number of connected vehicles) on the $i$-th RSU within a time period, and $R$ denotes the total number of RSUs in the network.		
		\item Time consumption: It is evaluated by the time consumption for a strategy to converge. This metric indicates the efficiency of different data offloading strategies.
	\end{enumerate}

\end{document}